\documentclass[journal]{IEEEtran}
\ifCLASSINFOpdf
\else
\fi
\usepackage[stretch=10]{microtype}
\usepackage[utf8]{inputenc}
\usepackage[T1]{fontenc}
\PassOptionsToPackage{hyphens}{url}
\usepackage{times,multirow,float}
\usepackage{graphicx}
\usepackage{epsfig,xspace,layout}
\usepackage{color}
\usepackage{amsfonts}
\usepackage{times}
\usepackage{amssymb}
\usepackage{amsthm}
\usepackage{hyperref}
\usepackage{amsmath,bm}
\usepackage{rotating}
\usepackage{mathrsfs}
\usepackage{mathtools}
\usepackage{makeidx}
\usepackage[]{threeparttable}
\usepackage{dsfont}
\usepackage{cite}
\usepackage{cuted}
\usepackage{flushend}
\AfterEndEnvironment{strip}{\leavevmode}
\usepackage{xcolor}
\usepackage{cleveref}
\usepackage[ruled, lined, linesnumbered, commentsnumbered, longend]{algorithm2e}
\usepackage{tikz}
\usetikzlibrary{shapes,arrows,positioning,calc}
\usepackage{siunitx}
\usepackage{multicol,lipsum}
\usepackage{subcaption}
\usepackage[font=footnotesize]{caption}
\usepackage{booktabs}
\usepackage{lineno}
\usepackage{hhline}
\usepackage[Symbol]{upgreek}
\usepackage{float}
\theoremstyle{remark}
\newtheorem{remark}{Remark}

\SetKwComment{Comment}{$\triangleright$\ }{}
\SetKwInput{Return}{Return}
\makeatletter
\renewcommand{\Indentp}[1]{%
  \advance\leftskip by #1
  \advance\skiptext by -#1
  \advance\skiprule by #1}%
\renewcommand{\Indp}{\algocf@adjustskipindent\Indentp{\algoskipindent}}
\renewcommand{\Indm}{\algocf@adjustskipindent\Indentp{-\algoskipindent}}
\makeatother

\begin{document}
\captionsetup[table]{name=TABLE,labelsep=newline,textfont=sc}
%
\title{Auto-Multilift: Distributed Learning and Control for Cooperative Load Transportation With Quadrotors}
%
%

\author{Bingheng Wang, Rui Huang, and Lin Zhao

\thanks{The authors are with the Department of Electrical and Computer Engineering,
        National University of Singapore, Singapore 117583, Singapore (email: 
        {wangbingheng@u.nus.edu},
        {elezhli@nus.edu.sg, ruihuang@nus.edu.sg}).}
        }


%
%
\newcommand{\Lin}[1]{\textcolor{blue}{[#1]}}

\markboth{IEEE XXX}%
{B. Wang \MakeLowercase{\textit{et al.}}: Auto-Multilift}
%



\maketitle

\begin{abstract}
Designing motion control and planning algorithms for multilift systems remains challenging due to the complexities of dynamics, collision avoidance, actuator limits, and scalability. Existing methods that use optimization and distributed techniques effectively address these constraints and scalability issues. However, they often require substantial manual tuning, leading to suboptimal performance. This paper proposes Auto-Multilift, a novel framework that automates the tuning of model predictive controllers (MPCs) for multilift systems. We model the MPC cost functions with deep neural networks (DNNs), enabling fast online adaptation to various scenarios. We develop a distributed policy gradient algorithm to train these DNNs efficiently in a closed-loop manner. Central to our algorithm is distributed sensitivity propagation, which is built on fully exploiting the unique dynamic couplings within the multilift system. It parallelizes gradient computation across quadrotors and focuses on actual system state sensitivities relative to key MPC parameters. Extensive simulations demonstrate favorable scalability to a large number of quadrotors. Our method outperforms a state-of-the-art open-loop MPC tuning approach by effectively learning adaptive MPCs from trajectory tracking errors. It also excels in learning an adaptive reference for reconfiguring the system when traversing multiple narrow slots.

\end{abstract}

\begin{IEEEkeywords}
Model predictive control, Bilevel optimization, Distributed learning, Neural network, Unmanned aerial vehicle.
\end{IEEEkeywords}

%
\IEEEpeerreviewmaketitle

\section*{Supplementary Material}
The source code of this work is available at \url{https://github.com/RCL-NUS/Auto-Multilift}.

\section{Introduction}
\label{sec:intro}
%
%
%
%

\begin{figure}[t!]
	\centering
	{\includegraphics[width=1\columnwidth]{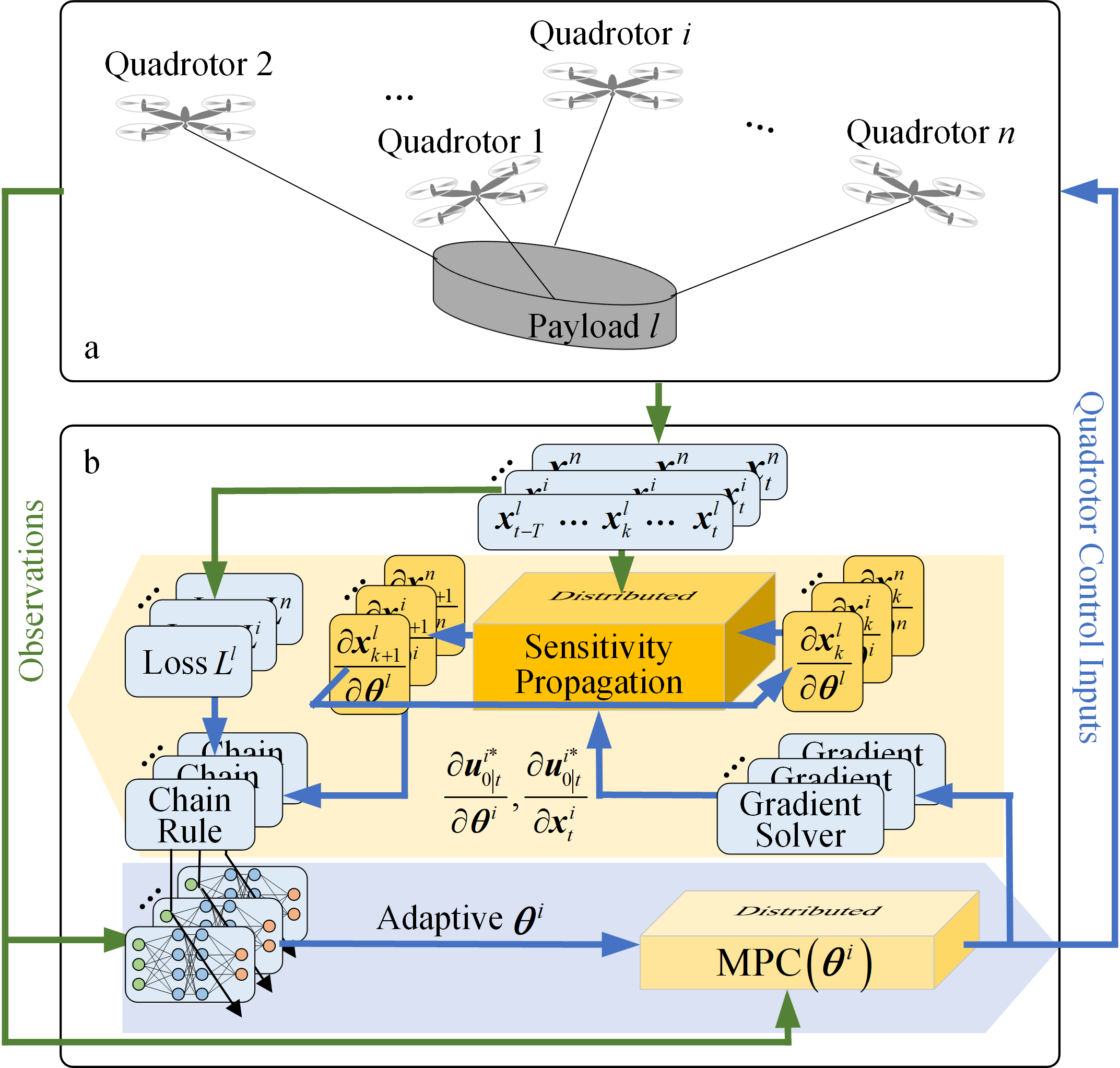}}
	\caption{\footnotesize Illustration of a multilift system and Auto-Multilift learning pipelines. (a) Components and structure of a typical multilift system; (b) A block diagram of the proposed framework. We fuse DNNs with distributed MPC controllers to obtain adaptive cost function parameters online. These DNNs can be trained efficiently in a distributed and close-loop manner. This is achieved through distributed sensitivity propagation, a key module in our method that computes actual state sensitivities in parallel across the quadrotors.}
\label{fig:automultilift learning pipeline}	
\end{figure}

\IEEEPARstart{A}{erial} transportation with multiple quadrotors typically offers greater load capacity and enhanced system robustness than employing a single quadrotor. Cables represent a promising mechanism for attaching the load to the quadrotors because they are lightweight, easy to design, and particularly suited for large-scale load deployment~\cite{masone2021shared}. The multilift system, in which a group of quadrotors cooperatively transports a cable-suspended load (see Fig.~\ref{fig:automultilift learning pipeline}-a), has therefore received increasing attention for potential applications such as rescue operations and logistics~\cite{michael2011cooperative,masone2016cooperative,villa2020survey,li2023rotortm}.

Despite its advantages, the multilift system faces unique challenges due to the cables. The quadrotors' motions are constrained by the cable length when taut and dynamically coupled with the load through the cable tensions. Moreover, the cables impart a hybrid nature to the system during transitions between slack and taut conditions, which significantly complicates the dynamics. These dynamic couplings and constraints require careful coordination among the quadrotors to maintain safe inter-robot separation and minimize slack-taut transitions. Other challenges include managing the quadrotor's control constraints, avoiding obstacles, and scaling to a large number of quadrotors.

Research on motion planning and control of the multilift system has progressed over the years. Early works assumed the load as a point mass and treated the tensions as external disturbances on the quadrotors~\cite{klausen2018cooperative,de2019flexible}. Disturbance observers were used to estimate and counteract the tensions for maintaining the quadrotors' trajectory tracking performance. This control method, which does not require a load model for simplicity, has recently regained attention~\cite{liu2021analysis,liu2023configuration,zhang2023formation}. However, ignoring the load dynamics in control design can compromise the system's maneuverability, restricting this method to quasi-static or slow flights. To improve control performance in agile flights, Lee et al.~\cite{lee2013geometric,lee2015collision} developed a geometric control method that explicitly incorporates the dynamic couplings between the quadrotors and a point-mass load. Additionally, Jackson et al.~\cite{jackson2020scalable} tackled the collision-free motion planning problem for a point mass load and proposed a distributed optimization approach that parallelizes trajectory planning across quadrotors. Nevertheless, methods based on the point mass assumption are limited to special cases where the load's shape is negligible compared to the cable length, a condition often overly conservative for practical use. 

On the other hand, treating the load as a rigid body with 6 degree-of-freedom (DoF) is more realistic but complicates the design of motion planning and control strategies. Achieving arbitrary control of a rigid-body load in 6 DoF requires at least three quadrotors~\cite{fink2011planning}. A multilift system with more than two quadrotors introduces redundancy in the cable tensions~\cite{sreenath2013dynamics}, leading to their non-unique distribution across the quadrotors. In~\cite{lee2017geometric,li2021cooperative,zhao2023composite}, the authors proposed the minimum-norm solution for distributing the cable tensions. In contrast, Geng et al.~\cite{geng2019implementation,geng2020cooperative} exploited this redundancy for a multilift system with four quadrotors to achieve optimal tension distribution while maintaining safe inter-rotor separation and respecting the quadrotor's thrust limits. Furthermore,  recent works~\cite{li2023nonlinear,sun2023nonlinear} included this redundancy in a centralized MPC framework, which not only ensures these safety constraints but also improves the system's maneuverability by fully accounting for the nonlinear system dynamics. In addition, Geng et al.~\cite{geng2022load} recently developed an optimal motion planning method, aiming for even tension distribution and system dynamics compliance. However, its assumption of small load attitude angles prevents applicability to large load attitude maneuvers.

As demonstrated in the above works~\cite{jackson2020scalable,geng2019implementation,geng2020cooperative,li2023nonlinear,sun2023nonlinear,geng2022load}, optimization methods are effective for multilift systems due to their ability to manage constraints and nonlinear dynamics. However, their performance significantly depends on the tuning of hyperparameters, including the weightings and references in cost functions. This tuning typically requires extensive manual effort, complicated by the large number of parameters and their dynamic interconnections from the cables. To simplify this, it often involves conservative assumptions such as treating the load as a point mass, assuming quasi-static flight, and uniform tension distribution, which can lead to suboptimal results. The challenge of manually selecting these parameters increases with more quadrotors and when their optimal values are dynamic and interconnected, such as when the load’s mass distribution is non-uniform and the system must dynamically adjust its configuration to avoid obstacles.

In this paper, we propose Auto-Multilift, a novel framework designed to automate the tuning of MPCs for multilift systems. We employ DNNs to dynamically adjust the weightings and references online within the MPC cost functions and present a highly efficient approach for training these DNNs using advanced machine learning techniques. Our method integrates the benefits of model-based control with end-to-end learning. Fig.~\ref{fig:automultilift learning pipeline}-b illustrates the block diagram of Auto-Multilift and its learning pipelines. In the forward pass, distributed MPCs with these adaptive hyperparameters solve open-loop optimal control problems in parallel to generate control trajectories for a future horizon, but only the first control commands are applied to the system. In the backward pass, we develop a distributed policy gradient algorithm to efficiently train these DNNs. Central to our algorithm is distributed sensitivity propagation (DSP), which calculates the system state sensitivities relative to these hyperparameters using the closed-loop states and the gradients of the first control commands. We employ and tailor Safe-PDP~\cite{jin2021safe} to obtain the gradients. A critical insight is that we can fully exploit the unique dynamic couplings within the multilift system to solve these sensitivities in parallel, enabling the efficient training of the DNNs in a \textit{distributed and closed-loop} manner. Our method also allows for the formulation of a loss function that evaluates performance over a time interval longer than the MPC's horizon.   

Interest in learning for MPC has burgeoned, with advances like  differentiable MPC~\cite{amos2018differentiable}, PDP~\cite{jin2020pontryagin}, Safe-PDP~\cite{jin2021safe}. Compared to differentiable MPC, PDP offers greater computational efficiency by recursively computing the MPC gradient via implicit differentiation of Pontryagin’s maximum principle. This technique, extended to constrained optimal control problems by Safe-PDP, has also been applied to learning cooperative control policies for multi-agent systems without dynamic couplings among agents~\cite{lu2022cooperative,cao2022game}. However, all these methods use open-loop state trajectories obtained by solving the MPC problem once per training episode, typically confining them to imitation learning that heavily relies on expert demonstrations. Conversely, using closed-loop state trajectories allows for broader training settings, such as reinforcement learning. Various closed-loop MPC learning strategies have been developed, including sampling-based methods~\cite{williams2018information,loquercio2022autotune}, black-box optimization methods~\cite{song2022policy,frohlich2022contextual,romero2023weighted}, and Actor-Critic MPC~\cite{romero2023actor}. The most recent works~\cite{tao2023difftune,zhang2024inverse} sought to enhance training efficiency by integrating differentiable MPC and PDP with Diff-Tune~\cite{cheng2022difftune}, respectively, which computes the closed-loop state gradient relative to the MPC hyperparameters via sensitivity propagation. Our work contributes to this collection by providing a distributed, closed-loop MPC learning framework for multi-agent systems with strong dynamic couplings and constraints among agents.

We evaluate the effectiveness of Auto-Multilift through extensive simulations that account for realistic effects such as cable elasticity and motor dynamics. In a complex simulation with up to six quadrotors transporting a load with non-uniform mass distribution, we demonstrate that our framework effectively learns adaptive MPC weightings from trajectory tracking errors. These weightings optimally distribute the cable tensions among the quadrotors to maintain the load attitude. This simulation also underscores the importance of using DSP in learning MPCs for multilift systems. Compared to training with Safe-PDP~\cite{jin2021safe} alone, our method offers more stable performance and reduces the tracking errors by up to $69\%$. Finally, we perform a challenging obstacle avoidance simulation, showing that Auto-Multilift can learn an adaptive tension reference for MPCs. This capability enables the system to reconfigure itself while passing through multiple narrow slots with heights shorter than the cable length. 

In summary, our main contributions are as follows:
\begin{enumerate}
\item We propose Auto-Multilift, a distributed and closed-loop learning framework that automatically tunes various MPC hyperparameters, modeled by DNNs, for multilift systems and adapts to different flight scenarios.
\item We develop a distributed sensitivity propagation (DSP) algorithm that efficiently computes the close-loop state sensitivities relative to the MPC hyperparameters in parallel. 
\item We design a distributed policy gradient algorithm to train these DNNs directly from the system tracking errors in a distributed and closed-loop manner. It consists of DSP and MPC gradient solvers, the latter derived by tailoring Safe-PDP~\cite{jin2021safe}. 
\item We conduct extensive simulations to showcase the scalability, the effective closed-loop learning, and the improved learning stability and trajectory tracking performance over the state-of-the-art open-loop MPC tuning method.
\end{enumerate}

The rest of this paper is organized as follows. Section~\ref{section:preliminaries} establishes the multilift system model and designs the MPC controllers. Section~\ref{section:automultilift} formulates the Auto-Multilift problem. In Section~\ref{section:distributed sensitivity propagatiuon}, we detail the development of the DSP algorithm. Section~\ref{section:policy gradient} designs the distributed policy gradient algorithm. Simulation results are reported in Section~\ref{section: simulation}. We discuss the advantages and disadvantages of our method in Section~\ref{section: discussion} and conclude this paper in Section~\ref{section: conclusion}.

\section{Preliminaries}\label{section:preliminaries}
\subsection{Multilift Model}\label{subsec:multilift model}
\begin{figure}[h]
	\centering
	{\includegraphics[width=0.75\columnwidth]{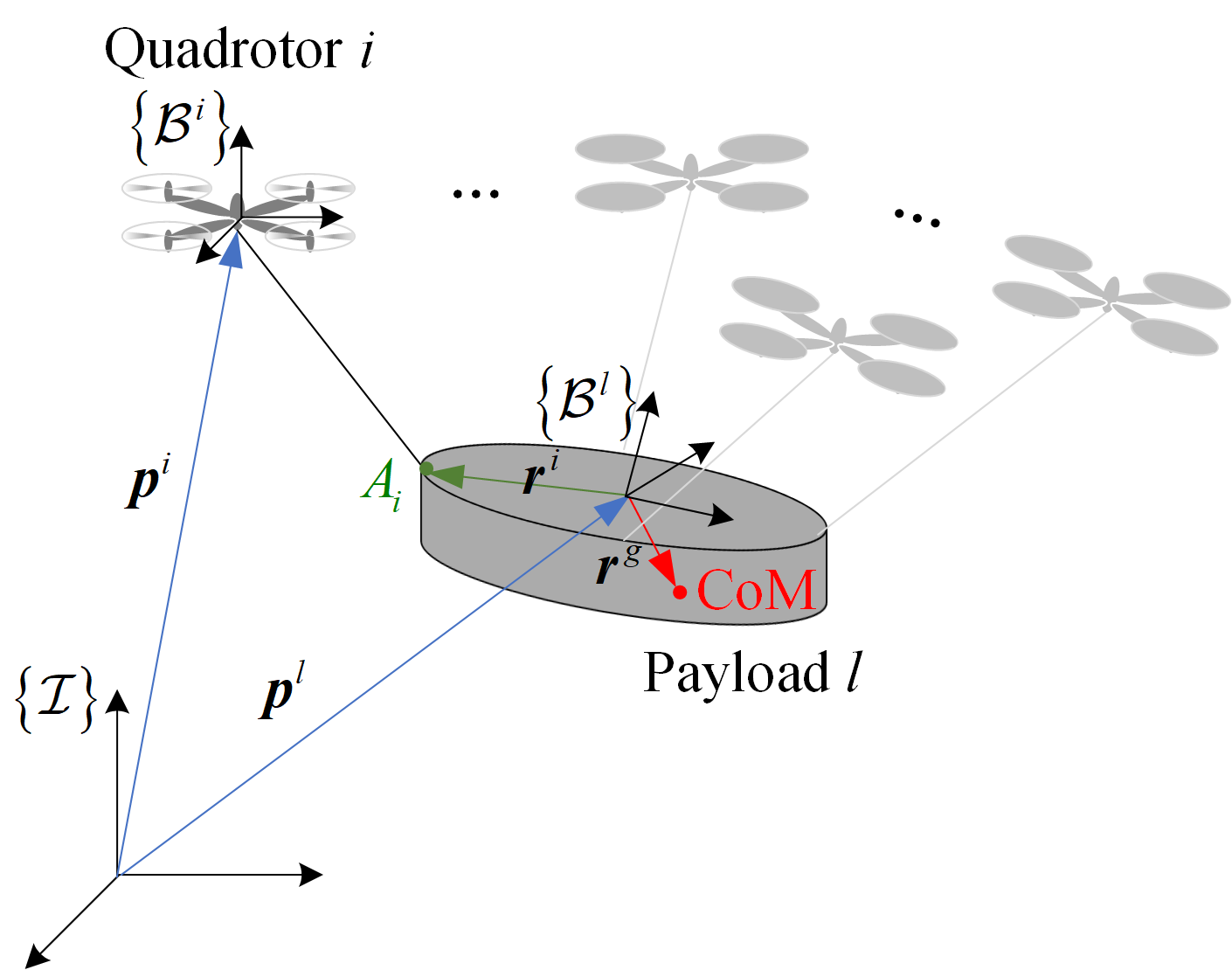}}
	\caption{\footnotesize Illustration of the multilift system. Let $\mathcal{I}$, $\mathcal{B}^l$, and $\mathcal{B}^i$ denote the world frame, the body frame attached to the load, and the body frame attached to the $i$-th quadrotor, respectively. }
\label{fig:multilift model}	
\end{figure}
We consider a multilift system with $n$ quadrotors and a rigid-body load. Each quadrotor is connected to an attachment point on the load via an elastic cable. For example, as shown in Fig.~\ref{fig:multilift model}, the $i$-th quadrotor is linked to the attachment point $A_i, \forall i\in {\cal I}_q$ with ${\cal I}_q= \left \{ 1,\cdots,n \right \}$ denoting the indices of the quadrotors. Each quadrotor is modeled as a 6 DoF rigid-body with the mass $m^i\in \mathbb{R}_{+}$ and the moment of inertia $\mathbf{J}^i\in \mathbb{R}^{3\times 3}$. We assume that each cable is attached to the center-of-mass (CoM) of the quadrotor, producing only a tension force acting on it. For the $i$-th quadrotor, $\forall i\in {\cal I}_q$, let $\bm{p}^i\in \mathbb{R}^3$ denote the CoM in $\mathcal{I}$, $\bm{v}^{i}\in \mathbb{R}^{3}$ the velocity of the CoM in $\mathcal{I}$, $\bm{q}^{i}\in \mathbb{R}^{4}$ the quaternion, and ${\boldsymbol {\omega}} ^{i}\in \mathbb{R}^{3}$ the angular velocity in $\mathcal{B}^i$. The quadrotor's dynamics model is given by
\begin{subequations}
\begin{align}
\dot {\bm p}^i & = {\bm v}^i, \\
\dot {\bm v}^i & = \frac{1}{m^{i}}\left( -{m^i g {\bm e}_3  + {\mathbf{R}_i} f^i {\bm e}_3  + {{\bm T}^{i,l}}} \right),\\
\dot{\bm {q}}^{i} & =\frac{1}{2}{\boldsymbol{\Omega}}\left ( {\bm\omega}^i \right ) {\bm q}^{i},\\
\dot {\boldsymbol \omega}^i  & = {{\left({\mathbf{J}}^i\right)}^{ - 1}}\left( { - {\left(\boldsymbol {\omega}^i \right)^ \times } {\mathbf{J}}^i\left(\boldsymbol {\omega}^i \right)  + \boldsymbol{\tau}^i} \right)
\end{align}
\label{eq:quadrotor model}%
\end{subequations}
where $g$ is the gravitational acceleration, ${\bm e}_3  = {\left[ {0,0,1} \right]}^T$, $\mathbf{R}_i:={\mathbf{R}\left(\bm {q}^i\right)}\in SO\left(3\right)$ is the rotation matrix with $\bm {q}^i$ from $\mathcal{B}^i$ to $\mathcal{I}$, ${\bm T}^{i,l}\in \mathbb{R}^3$ is the $i$-th cable tension in $\mathcal{I}$, $\left(\boldsymbol {\omega}^i \right)^ \times $ denotes the skew-symmetric matrix form of $\boldsymbol {\omega}^i$ as an element of the Lie algebra $\mathfrak{so}(3)$, $f^i \in \mathbb{R}_{+}$ and $\boldsymbol{\tau}^i\in \mathbb{R}^3$ are the total thrust and control torque produced by the quadrotor's four motors in $\mathcal{B}^i$, respectively, and ${\boldsymbol{\Omega}}\left ( {\bm\omega}^i \right )$ is given by
\begin{equation}
    \boldsymbol{\Omega }\left ( {\bm \omega} ^{i} \right )=\begin{bmatrix}
0 & -\left (  {\bm\omega} ^{i}\right )^{T}\\ 
{\bm\omega} ^{i} & -\left (  {\bm\omega} ^{i}\right )^{\times }
\end{bmatrix}.
\nonumber
\end{equation}

We model the load as a 6 DoF rigid-body with the mass $m^l\in \mathbb{R}_{+}$ and the moment of inertia $\mathbf{J}^l\in\mathbb{R}^{3\times3}$. The load's body frame $\mathcal{B}^l$ is built at its geometric center (GC), which can be easily determined when using a regular shaped container for load transportation, commonly seen in scenarios such as logistics. In pursuit of more realistic modelling, we discard the conservative assumption of uniform mass distribution of the load, resulting in a bias vector ${\bm r}^g\in\mathbb{R}^3$ for the load's CoM in $\mathcal{B}^l$. Let $\bm {p}^l\in \mathbb{R}^3$ denote the position of the GC in $\mathcal{I}$, $\bm {v}^l\in \mathbb{R}^3$ the velocity of the GC in $\mathcal{B}^l$, $\bm{q}^l\in\mathbb{R}^4$ the quaternion of the load's attitude, $\boldsymbol{\omega}^l\in\mathbb{R}^3$ the angular velocity of the load in $\mathcal{B}^l$. The load's dynamics model is given by
\begin{subequations}
\begin{align}
\dot{\bm p}^{l} &=\mathbf{R}_l{\bm v}^{l},\\
\begin{split}
    \dot{\bm v}^{l} &= -\left (\dot{\boldsymbol{\omega }}^{l}  \right )^{\times }{\bm r}^{g}-\left ( \boldsymbol{\omega }^{l} \right )^{\times }{\bm v}^{l}-\left ( \boldsymbol{\omega }^{l} \right )^{\times }\left ( \left ( \boldsymbol{\omega }^{l} \right )^{\times }{\bm r}^{g} \right )\\
    &\quad +\frac{1}{m^{l}}\mathbf{R}^{T}_l\left ( \sum_{i=1}^{n}{\bm T}^{l,i} -m^{l}g{\bm e}_{3}\right ),
\end{split}\\
\dot{\bm {q}}^{l} & =\frac{1}{2}{\boldsymbol{\Omega}}\left ( {\bm\omega}^l \right ) {\bm q}^{l},\\
\begin{split}
{\dot {\boldsymbol\omega} ^l} & = {\left( {{{\mathbf {J}}^l}} \right)^{ - 1}}\left[ {\sum\limits_{i = 1}^n {\left( {{{\left( {{{\bm r}^i}} \right)}^ \times }{{\mathbf {R}}^T_l}{{\bm T}^{l,i}}} \right)} } \right.\\
&\quad\quad\quad\quad\quad -\left ( {\bm r}^{g} \right )^{\times }\mathbf{R}^{T}_lm^{l}g{\bm e}_{3}-\left ( \boldsymbol{\omega }^{l} \right )^{\times }\left ( \mathbf{J}^{l} \boldsymbol{\omega }^{l}\right )\\
&\quad\quad\quad\quad\quad \left. {-m^{l}\left ( {\bm r}^{g} \right )^{\times }\left ( \dot{\bm v}^{l}+\left ( \boldsymbol{\omega }^{l} \right )^{\times } {\bm v}^{l}\right )} \right]
\end{split}
\end{align}
\label{eq:load model}%
\end{subequations}
where $\mathbf{R}_{l}:= \mathbf{R}\left ( {\bm q}^{l} \right )\in RO\left( 3\right)$ is the rotation matrix with ${\bm q}^l$ from $\mathcal{B}^l$ to $\mathcal{I}$, $\bm T^{l,i}=-\bm T^{i,l}$, and ${\bm r}^{i}$ denotes the coordinate of $A_i$ in $\mathcal{B}^l$.

Given that the cable's mass is typically negligible compared to the quadrotor and the load, we model the cables as massless spring-damper links. Despite this, our model retains the cable's hybrid nature, meaning it can only 'pull' but not 'push' the attached object. Therefore, the magnitude of the tension force is a piecewise function given by
\begin{equation}
    T^{i} = \left\{\begin{matrix}
K\left ( l^{i}-l^{0} \right ) + c_{t}K\dot{l}^{i} & \mathrm{if}\ l^{i}> l^{0}\\ 
0 & \mathrm{otherwise}
\end{matrix}\right.
\label{eq:tension magnitude}
\end{equation}
where $K\in\mathbb{R}_+$ is the cable stiffness, $c_t\in\mathbb{R}_+$ is the cable damping ratio, $l^i$, $\forall i\in {\cal I}_q$, is the $i$-th cable's stretched length, and $l^0$ is the cable's natural length. The stretched length $l^i$ is defined as the $2$-norm of the relative position 
\begin{equation}
    {\bm p}^{l,i}={\bm p}^{i}-{\bm p}^{l}-{\mathbf{R}}_{l}{\bm r}^{i}
\label{eq:relative position}
\end{equation}
between the $i$-th quadrotor and its attachment point $A_i$ on the load. Using \eqref{eq:tension magnitude} and \eqref{eq:relative position}, the $i$-th tension vector is computed by
\begin{equation}
    {\bm T}^{l,i}=T^{i}\frac{{\bm p}^{l,i}}{\left \| {\bm p}^{l,i} \right \|_2}.
\label{eq:tension force}
\end{equation}

\subsection{Centralized MPC}\label{subsec:centralized mpc}

We employ MPC to design motion control and planning strategies for a multilift system. Our aim is to track the references of all the agents (including the load) while maintaining a stable attitude of the load during transportation. In this subsection, we present a single, large MPC formulation that generates the state and control trajectories for the system in a centralized manner. This formulation is inspired by the method in~\cite{jackson2020scalable} but extended to a rigid-body load. Let ${\bm x}^{i}=\left [ {\bm p}^{i},{\bm v}^{i},{\bm q}^{i},\boldsymbol{\omega }^{i} \right ]\in \mathbb{R}^{13}$ denote the $i$-th quadrotor's state, ${\bm u}^{i}=\left [ f^{i},\boldsymbol{\tau }^{i} \right ]\in \mathbb{R}^{4}$ the $i$-th quadrotor's control, ${\bm x}^{l}=\left [ {\bm p}^{l},{\bm v}^{l},{\bm q}^{l},\boldsymbol{\omega }^{l} \right ]\in \mathbb{R}^{13}$ the load's state, and ${\bm u}^{l}=\left [ \bar{T}^{1},\cdots ,\bar{T}^{n} \right ]\in \mathbb{R}^{n}$ the load's virtual control. Note that the tension magnitude~\eqref{eq:tension magnitude} is not used in the MPC as it introduces a hybrid nature into the system model, making the resulting optimization problem difficult to solve. Instead, we denote $\bar{T}^i$ as a tension magnitude optimized in the MPC for open-loop prediction but not applied to the system. This makes it a virtual control for the load.

The cost function of each quadrotor is designed in a quadratic form to penalize the deviations of the quadrotor's state and control from their references. It is written by
\begin{equation}
    J^{i}=\frac{1}{2}\sum_{k=0}^{N-1}\left ({\bm e}_{x_{k}^{i}}^{T}\mathbf{Q}_{x^{i}}{\bm e}_{x_{k}^{i}} + {\bm e}_{u_{k}^{i}}^{T}\mathbf{Q}_{u^{i}}{\bm e}_{u_{k}^{i}} \right ) + \frac{1}{2}{\bm e}_{x_{N}^{i}}^{T}\mathbf{Q}_{x_{N}^{i}}{\bm e}_{x_{N}^{i}}
\label{eq:quadrotor cost}
\end{equation}
where $N\in\mathbb{R}_+$ is the MPC's prediction horizon, $\mathbf{Q}_{x^{i}}\succ 0\in \mathbb{R}^{12\times 12}$, $\mathbf{Q}_{x_{N}^{i}}\succ 0\in \mathbb{R}^{12\times 12}$, and $\mathbf{Q}_{u^{i}}\succ 0\in \mathbb{R}^{4\times 4}$ are the positive definite weighting matrices. In \eqref{eq:quadrotor cost}, ${\bm e}_{x_{k}^{i}}$ and ${\bm e}_{u_{k}^{i}}$ denote the quadrotor's state and control tracking errors at the time step $k$, respectively, which are defined as 
\begin{equation}
    {\bm e}_{x_{k}^{i}}=\begin{bmatrix}
{\bm p}_{k}^{i}-{\bm p}_{k}^{i,\mathrm{ref}}\\ 
{\bm v}_{k}^{i}-{\bm v}_{k}^{i,\mathrm{ref}}\\ 
\frac{1}{2}\left ({\mathbf{R}_{i,k}^{\mathrm{ref}}}^T\mathbf{R}_{i,k}-\mathbf{R}_{i,k}^{T}\mathbf{R}_{i,k}^{\mathrm{ref}}  \right )^{\vee }\\ 
\boldsymbol{\omega }_{k}^{i}-\boldsymbol{\omega }_{k}^{i,\mathrm{ref}}
\end{bmatrix},\ {\bm e}_{u_{k}^{i}}=\begin{bmatrix}
f_{k}^{i}-f_{k}^{i,\mathrm{ref}}\\ 
\boldsymbol{\tau }_{k}^{i}-\boldsymbol{\tau }_{k}^{i,\mathrm{ref}}
\end{bmatrix}
\label{eq:tracking errors}
\end{equation}
where $\left ( \cdot  \right )^{\mathrm{ref}}$ denotes a reference signal and $\left ( \cdot  \right )^{\vee }$ represents the \textit{vee} operator: $\mathfrak{so}(3)\rightarrow \mathbb{R}^{3}$.

The cost function of the load has the same form as $J^i$ and is defined by
\begin{equation}
    J^{l}=\frac{1}{2}\sum_{k=0}^{N-1}\left ({\bm e}_{x_{k}^{l}}^{T}\mathbf{Q}_{x^{l}}{\bm e}_{x_{k}^{l}} + {\bm e}_{u_{k}^{l}}^{T}\mathbf{Q}_{u^{l}}{\bm e}_{u_{k}^{l}} \right ) + \frac{1}{2}{\bm e}_{x_{N}^{l}}^{T}\mathbf{Q}_{x_{N}^{l}}{\bm e}_{x_{N}^{l}}
\label{eq:load cost}
\end{equation}
where $\mathbf{Q}_{x^{l}}\succ 0\in \mathbb{R}^{12\times 12}$, $\mathbf{Q}_{x_{N}^{l}}\succ 0\in \mathbb{R}^{12\times 12}$, and $\mathbf{Q}_{u^{l}}\succ 0\in \mathbb{R}^{n\times n}$ are the positive definite weighting matrices, ${\bm e}_{x_{k}^{l}}$ and ${\bm e}_{u_{k}^{l}}$ denote the load's state and control tracking errors, respectively, having the same form as defined in~\eqref{eq:tracking errors}.

With the above definitions, we can formulate the centralized MPC problem as follows:
\begin{subequations}
\begin{align}
&\mathop {\min }\limits_{{\bf{X}},{\bf{U}}} \ {J^{l}\left ( X^{l},U^{l} \right )+\sum_{i=1}^{n}J^{i}\left ( X^{i},U^{i} \right )}\\
{\rm s.t.}\ & {\bm x}_{k+1}^{i} =\bar{\bm f}_{k}^{i}\left ( {\bm x}_{k}^{i},{\bm u}_{k}^{i},\Delta t;{\bm x}_{k}^{l},{\bm u}_{k}^{l} \right ), \forall i\in \mathcal{I}_{q}\label{eq:quadrotor model constraint}\\
& {\bm x}_{k+1}^{l} =\bar{\bm f}_{k}^{l}\left ( {\bm x}_{k}^{l},{\bm u}_{k}^{l},\Delta t;{\bm x}_{k}^{1},\cdots ,{\bm x}_{k}^{n} \right )\label{eq:load model constraint}\\
& {\bm x}_{0}^{i}={\bm x}_t^{i},\forall i\in \mathcal{I}_{q}\label{eq:initial quadrotor state constraint}\\
& {\bm x}_{0}^{l}={\bm x}_t^{l}\label{eq:initial load state constraint}\\
& {\bm u}_{\min}^{i}\leq {\bm u}_{k}^{i}\leq {\bm u}_{\max}^{i},\forall i\in \mathcal{I}_{q}\label{eq:quadrotor control constraint}\\
& 0< \left ( {\bm u}_{k}^{l} \right )_{i}\leq \bar{T}_{\max},\forall i\in \mathcal{I}_{q}\label{eq:load tension constraint}\\
& \left \| {\bm p}_{k}^{i}-{\bm p}_{k}^{l}-\mathbf{R}_{l,k}{\bm r}^{i} \right \|_{2}= l^{0},\forall i\in \mathcal{I}_{q}\label{eq:cable length constraint}\\
& 2d_{\mathrm{quad}}-\left \| {\bm p}_{k}^{i}-{\bm p}_{k}^{j}\right \|_{2}< 0,\forall i,j\in \mathcal{I}_{q} ,i\neq j \label{eq:interrobot separation}\\
& d_{\mathrm{quad}}+d_{\mathrm{obs}}-\left \| {\bm p}_{k}^{i}-{\bm p}_{\mathrm{obs}} \right \|_{2}< 0,\forall i\in \mathcal{I}_q\label{eq:obstacle avoidance for quadrotor}\\
& d_{\mathrm{load}}+d_{\mathrm{obs}}-\left \| {\bm p}_{k}^{l}-{\bm p}_{\mathrm{obs}} \right \|_{2}< 0\label{eq:obstacle avoidance for load}
\end{align}
\label{eq:centralized mpc}%
\end{subequations}
where $X^{i}=\left [ {\bm x}_{0}^{i},\cdots ,{\bm x}_{N}^{i} \right ]$ and $X^{l}=\left [ {\bm x}_{0}^{l},\cdots ,{\bm x}_{N}^{l} \right ]$ denote the state trajectories of length $N+1$ for the $i$-th quadrotor and the load, respectively, $U^{i}=\left [ {\bm u}_{0}^{i},\cdots ,{\bm u}_{N-1}^{i} \right ]$ and $U^{l}=\left [ {\bm u}_{0}^{l},\cdots ,{\bm u}_{N-1}^{l} \right ]$ are the corresponding control trajectories of length $N$, $\mathbf{X}=\left [ X^{1},\cdots ,X^{n},X^{l} \right ]$ and $\mathbf{U}=\left [ U^{1},\cdots ,U^{n},U^{l} \right ]$ are sets of the system's trajectories, ${\bm x}_t^{i}$ and ${\bm x}_t^{l}$ are the feedback states from the closed-loop system sampled at the current time $t$, $d_{\cdot } \in \mathbb{R}_{+}$ is a scalar dimension (e.g., $d_{\mathrm{quad} }$ denotes the quadrotor radius), ${\bm p}_{\mathrm{obs}}\in \mathbb{R}^{3}$ is the obstacle's location in $\mathcal{I}$, and $\Delta t\in \mathbb{R}_{+}$ is the discretization step in MPC. From top to bottom, the constraints in~\eqref{eq:centralized mpc} are introduced as follows: \eqref{eq:quadrotor model constraint} and \eqref{eq:load model constraint} are the discrete dynamics from \eqref{eq:quadrotor model} and \eqref{eq:load model} for the quadrotor and the load, respectively, using the virtual control $\bm u^l$ as the tension magnitudes; \eqref{eq:initial quadrotor state constraint} and \eqref{eq:initial load state constraint} are the initial conditions; \eqref{eq:quadrotor control constraint} is the quadrotor control constraint; \eqref{eq:load tension constraint} is the tension constraint that keeps the cable taut; \eqref{eq:cable length constraint} is the cable length constraint that prevents the potential collision between the quadrotor and the load, since the tension magnitude $\bar{T}^i$ is not generated by the cable deformation~\eqref{eq:tension magnitude} but is an optimization variable, the natural length $l^0$ is used in this constraint; \eqref{eq:interrobot separation} ensures safe inter-robot separation to avoid collisions among the quadrotors; \eqref{eq:obstacle avoidance for quadrotor} and \eqref{eq:obstacle avoidance for load} denote obstacle avoidance constraints for the quadrotors and the load, respectively.

\subsection{Distributed MPC}\label{subsec:distributed mpc}
Solving the centralized MPC problem~\eqref{eq:centralized mpc} is computationally intensive and scales poorly to large multilift systems. Here, we present a distributed formulation that decomposes the problem across the quadrotors. To achieve this, several features in~\eqref{eq:centralized mpc} are worth discussing. The objective function is separable by agent, as no coupling exists between the states and controls of the quadrotors and the load. Furthermore, the quadrotors' dynamics are indirectly coupled through the cable tensions (i.e., functions of $\bm x^l$ and $\bm {u}^l$) and the length constraint~\eqref{eq:cable length constraint}. Additionally, the states of the quadrotors are only coupled via the constraint of safe inter-robot separation~\eqref{eq:interrobot separation}.

From these observations, the decomposition can be achieved by treating all other agents' MPC trajectories as external signals during an agent's MPC open-loop prediction. Based on this, the distributed formulation is realized by independently updating all the MPC trajectories until convergence. 

\begin{remark}
\label{rm:soft constraints}
We shift the constraints \eqref{eq:cable length constraint}, \eqref{eq:interrobot separation}, \eqref{eq:obstacle avoidance for quadrotor}, and \eqref{eq:obstacle avoidance for load} to soft constraints, incorporating them into the cost function. Doing so facilitates the convergence, as demonstrated in~\cite{son2018model}. In the spirit of interior-point methods~\cite{forsgren2002interior}, we present the soft constraints using barrier functions, which can approximate the original constrained problem given a sufficiently small barrier parameter.
\end{remark}

Specifically, the trajectories $X^j, \forall j\neq i$, $X^l$, and $U^l$ are considered as the external trajectories for the $i$-th quadrotor in its decomposed MPC problem. These external trajectories remain unchanged during the open-loop prediction.  Thus, the problem can be defined as
\begin{subequations}
\begin{align}
&\mathop {\min }\limits_{X^i,U^i}\ J^{i}+\sum_{k=0}^{N}\left (\frac{1}{2\gamma } \left ( h_{k}^{i} \right )^{2}-\gamma \sum_{j\neq i}^{n}\ln\left ( -g_{k}^{j} \right ) +J_k^{\mathrm{o},i}\right )\\
{\rm s.t.}\ & {\bm x}_{k+1}^{i} =\bar{\bm f}_{k}^{i}\left ( {\bm x}_{k}^{i},{\bm u}_{k}^{i},\Delta t;{\bm x}_{k}^{l},{\bm u}_{k}^{l} \right ),\\
& {\bm x}_{0}^{i}={\bm x}_t^{i},\\
& {\bm u}_{\min}^{i}\leq {\bm u}_{k}^{i}\leq {\bm u}_{\max}^{i}
\end{align}
\label{eq:distributed mpc for quadrotor}%
\end{subequations}
where $\gamma \in \mathbb{R}_+$ is the barrier parameter, $h_k^{i} = \left \| \bm p^{l,i} \right \|_{2} - l^0$ with $\bm p^{l,i}$ defined in~\eqref{eq:relative position}, $g_k^j=2d_{\mathrm{quad}}-\left \| {\bm p}_{k}^{i}-{\bm p}_{k}^{j}\right \|_{2}$, $J_k^{\mathrm{o},i}=-\gamma \ln\left ( -g_{k}^{\mathrm{o},i} \right ) $, and $g_{k}^{\mathrm{o},i}=d_{\mathrm{quad}}+d_{\mathrm{obs}}-\left \| {\bm p}_{k}^{i}-{\bm p}_{\mathrm{obs}} \right \|_{2}$.

Accordingly, the trajectories $X^{i}, \forall i\in \mathcal{I}_{q}$, are considered as the external trajectories for the load's decomposed MPC problem, which is defined as
\begin{subequations}
\begin{align}
&\mathop {\min }\limits_{X^l,U^l}\ J^{l}+\sum_{k=0}^{N}\left (\frac{1}{2\gamma } \sum_{i=1}^{n} \left ( h_{k}^{i} \right )^{2} + J_k^{\mathrm{o},l}\right )\label{eq:cost of load distributed mpc}\\
{\rm s.t.}\ & {\bm x}_{k+1}^{l} =\bar{\bm f}_{k}^{l}\left ( {\bm x}_{k}^{l},{\bm u}_{k}^{l},\Delta t;{\bm x}_{k}^{1},\cdots,{\bm x}_{k}^{n} \right ),\\
& {\bm x}_{0}^{l}={\bm x}_t^{l},\\
& 0< \left ( {\bm u}_{k}^{l} \right )_{i}\leq \bar{T}_{\max},\forall i\in \mathcal{I}_{q}\label{eq:tension magnitude constraint}
\end{align}
\label{eq:distributed mpc for load}%
\end{subequations}
where $J_k^{\mathrm{o},l}=-\gamma \ln\left ( -g_{k}^{\mathrm{o},l} \right )$ and $g_{k}^{\mathrm{o},l}=d_{\mathrm{load}}+d_{\mathrm{obs}}-\left \| {\bm p}_{k}^{l}-{\bm p}_{\mathrm{obs}} \right \|_{2}$.

\begin{figure}[h]
	\centering
	{\includegraphics[width=0.7\columnwidth]{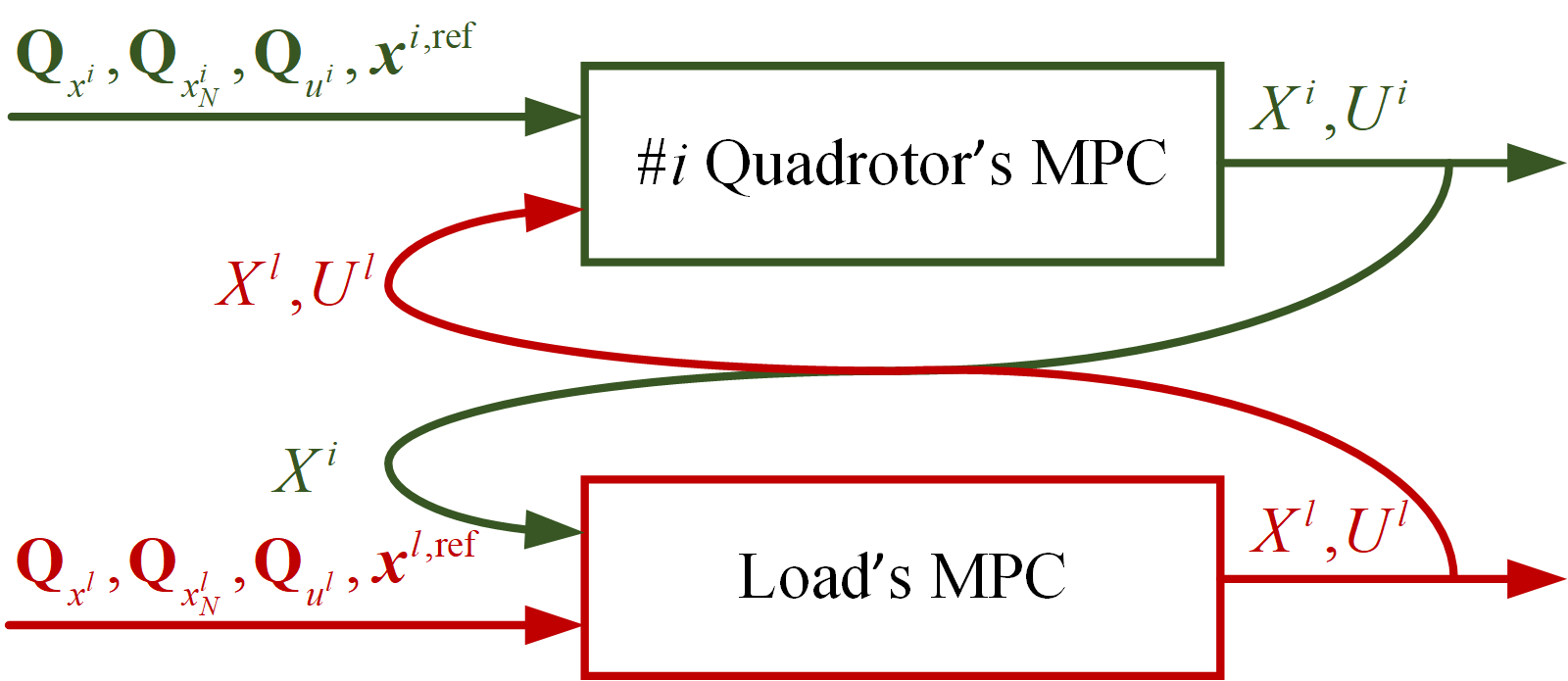}}
	\caption{Illustration of the data exchange used in the distributed MPC. }
\label{fig:mpc data flow}	
\end{figure}

The distributed MPC formulation consists of Problem~\eqref{eq:distributed mpc for quadrotor} and Problem~\eqref{eq:distributed mpc for load}, both solved iteratively. Define $X_{k}^{i}$, $U_{k}^{i}$, $X_{k}^{l}$, and $U_{k}^{l}$ as the trajectories during the iteration $k$. At each sampling time $t$, given the initial conditions $X_{0}^{i},\forall i\in \mathcal{I}_{q}$, $X_{0}^{l}$, and $U_{0}^{l}$, Problem~\eqref{eq:distributed mpc for quadrotor} is solved in parallel for each quadrotor. Subsequently, Problem~\eqref{eq:distributed mpc for load} is solved for the load using the updated trajectories $X_{1}^{i},\forall i\in \mathcal{I}_{q}$. The process is repeated until the errors $\left \| X_{k}^{i}-X_{k-1}^{i} \right \|_{2}$ and $\left \| U_{k}^{i}-U_{k-1}^{i} \right \|_{2}, \forall i\in \mathcal{I}_{A}$ for all the agents, indexed by $\mathcal{I}_{A} = \left \{ 1,\cdots ,n,l \right \}$, fall within a predefined threshold $\delta \in \mathbb{R}_{+}$. Fig.~\ref{fig:mpc data flow} provides a clearer illustration of the data exchange (viewed as the external trajectories for one agent's MPC) in solving the distributed MPC problem.
Since the load lacks computational capability, a 'central' agent is randomly selected from the quadrotors and assigned to solve both Problem~\eqref{eq:distributed mpc for quadrotor} and Problem~\eqref{eq:distributed mpc for load} sequentially. The distributed MPC is summarized in Algorithm~\ref{alg: distributed mpc}.

\begin{algorithm}[h]
\caption{Distributed MPC for Multilift Systems}
\label{alg: distributed mpc}
\SetKwInput{Input}{Input}
\SetKwInput{Output}{Output}
\Input{The threshold $\delta$, the maximum iteration $k_{\max}$, and the initial conditions $X_{0}^{i}$, $U_{0}^{i}$, $X_{0}^{l}$, and $U_{0}^{l}$.}
$k=1$\\
\While{$e\geq \delta $ and $k\leq k_{\max}$}{
\For {$i \leftarrow 1$ \KwTo $n$ (in parallel)}{
Compute $X^i$ and $U^i$ by solving Problem~\eqref{eq:distributed mpc for quadrotor} based on the external trajectories $X_{k-1}^j, \forall j\in \mathcal{I}_{q},j\neq i$, $X_{k-1}^l$, and $U_{k-1}^l$; \Comment{run in each agent and sent to the central agent}
}
Update the quadrotor trajectories: $X_{k}^{i}\leftarrow X^{i}$ and $ U_{k}^{i}\leftarrow U^{i},\  \forall i\in \mathcal{I}_{q}$;\\
Compute $X^l$ and $U^l$ by solving Problem~\eqref{eq:distributed mpc for load} based on $X_{k}^{i},\  \forall i\in \mathcal{I}_{q}$;\\
Update the load trajectories: $X_{k}^{l}\leftarrow X^{l},U_{k}^{l}\leftarrow U^{l}$; \Comment{run in the central agent}
Compute the error: $e\leftarrow \max\left \{ e_{X}^{i}, e_{U}^{i}\right \},\forall i\in \mathcal{I}_{A}$ with 
$e_{X}^{i}=\frac{1}{N}\left \| X_{k}^{i}-X_{k-1}^{i} \right \|_{2}$ and $e_{U}^{i}=\frac{1}{N}\left \| U_{k}^{i}-U_{k-1}^{i} \right \|_{2}$;\\
Update the iteration: $k\leftarrow k+1$;
}
\Output{$X^{i}$ and $U^{i},\ \forall i\in \mathcal{I}_{A}$}
\end{algorithm}

\section{Formulation of Auto-Multilift} \label{section:automultilift}
\subsection{Problem Statement}\label{subsec:problem statement}
The distributed MPC, outlined in Algorithm~\ref{alg: distributed mpc}, approximates the centralized MPC \eqref{eq:centralized mpc} to enhance computational efficiency. Typically, the accuracy of this approximation improves as the threshold $\delta$ decreases and the maximum iteration $k_{\max}$ increases. However, the performance of Algorithm~\ref{alg: distributed mpc} in terms of the feasibility of the predicted trajectories and the tracking accuracy is contingent upon careful tuning of various hyperparameters. This includes the selection of the weighting matrices and the design of the references in~\eqref{eq:quadrotor cost} and~\eqref{eq:load cost}. For clarity, let $\boldsymbol{\theta }^{i}\in \mathbb{R}^{m^i}$, $ \forall i\in \mathcal{I}_{q}$, denote the hyperparameters for the $i$-th quadrotor and $\boldsymbol{\theta }^{l}\in \mathbb{R}^{m^l}$ the hyperparameters for the load. We can, therefore, parameterize the decomposed MPC problems for the quadrotor and the load as $\mathrm{QMPC}\left ( \boldsymbol{\theta } ^{i}\right )$, $\forall i\in \mathcal{I}_{q}$, and $\mathrm{LMPC}\left ( \boldsymbol{\theta } ^{l}\right )$, respectively, and denote the corresponding MPC's predicted trajectories as $\xi ^{\ast,i}\left ( \boldsymbol{\theta }^{i} \right )=\left \{ X^{\ast,i}\left ( \boldsymbol{\theta }^{i} \right ),  U^{\ast,i}\left ( \boldsymbol{\theta }^{i} \right )\right \}$ and $\xi ^{\ast,l}\left ( \boldsymbol{\theta }^{l} \right )=\left \{ X^{\ast,l}\left ( \boldsymbol{\theta }^{l} \right ),  U^{\ast,l}\left ( \boldsymbol{\theta }^{l} \right )\right \}$.

In practice, manually engineering these hyperparameters is generally difficult and inefficient due to their large number and dynamic interconnections. The first reason is intuitive: the number of hyperparameters increases with the use of more quadrotors in the multilift system, complicating the tuning process. The second reason arises from the dynamic couplings between the agents' motions, which are induced by the cables. This coupling effect becomes more pronounced in scenarios such as balancing a load with non-uniform mass distribution during agile flights. In such cases, the tension allocation among the quadrotors tends to be uneven and dynamic, necessitating dynamically interconnected MPC weightings. Furthermore, the MPC references should adapt to environments and the system states as the system dynamically adjusts its configuration to avoid collision with obstacles. It is less likely to achieve these goals by tuning the MPC hyperparameters of all the agents independently, and one set of fixed hyperparameters cannot be universally applied to all the agents. Manual engineering typically relies on conservative assumptions, such as even tension distribution and quasi-static flight, and is achieved by trial and error, leading to suboptimal performance.   

In this paper, \textbf{our interests} are to design adaptive MPC hyperparameters for multilift systems and to develop a systematic method for automatically tuning these hyperparameters. We use DNNs to generate the adaptive MPC hyperparameters, whose dynamic behaviors are typically challenging to model using first principles. Mathematically, the adaptive MPC hyperparameters for the $i$-th agent are given by
\begin{equation}
    \boldsymbol{\theta }^{i}=f_{\boldsymbol{\varpi}^{i} }\left ( \boldsymbol{\chi}^{i}  \right ),\ \forall i\in \mathcal{I}_{A}
    \label{eq:adaptive mpc hyperparameters}
\end{equation}
where $\boldsymbol{\varpi}^{i}$ denotes the learnable parameters of the $i$-th DNN and the input $\boldsymbol{\chi}^{i}$ includes observations from the environment and the multilift system, such as the system states and obstacle information, depending on the application. Thus, the tuning problem is to find optimal DNN parameters $\boldsymbol{\varpi }^{\ast,i}$, $\forall i\in \mathcal{I}_{A}$, for all the agents (including the load), such that the total loss, composed of individual losses that evaluate the performance of each agent, is minimized. It can be formulated as the following optimization problem:
\begin{subequations}
    \begin{align}
        &\mathop {\min }\limits_{\boldsymbol{\varpi}}\ L=L^{l}+\sum_{i=1}^{n}L^{i}\label{eq:closed loop total loss}\\
        {\rm s.t.}\ & {\bm x}_{t+1}^{i} =\bar{\bm f}_{t}^{i}\left ( {\bm x}_{t}^{i},{\bm u}_{0|t}^{\ast,i},\Delta t;{\bm x}_{t}^{l}, {\bm u}_{0|t}^{\ast,l}\right ),\ \forall i\in \mathcal{I}_{q} \label{eq:nominal quadrotor dynamics}\\
        & {\bm u}_{0|t}^{\ast,i}\ \mathrm{generated}\ \mathrm{by}\ \mathrm{QMPC}\left ( \boldsymbol{\theta }^{i}\left ( \boldsymbol{\varpi }^{i} \right ) \right )\label{eq:first qmpc control command}\\
        & {\bm x}_{t+1}^{l} =\bar{\bm f}_{t}^{l}\left ( {\bm x}_{t}^{l},{\bm u}_{0|t}^{\ast,l},\Delta t;{\bm x}_{t}^{1},\cdots ,{\bm x}_{t}^{n} \right )\label{eq:nominal load dynamics}\\
        & {\bm u}_{0|t}^{\ast,l}\ \mathrm{generated}\ \mathrm{by}\ \mathrm{LMPC}\left ( \boldsymbol{\theta }^{l}\left ( \boldsymbol{\varpi }^{l} \right ) \right )\label{eq:first lmpc control command}
    \end{align}
    \label{eq:closed-loop tuning problem}%
\end{subequations}
where $L^l$ and $L^i$ are the individual losses for the load and the $i$-th quadrotor based on their respective states $\bm x_t^l$ and $\bm x_t^i$ over a horizon with length of $N_{\rm cl}\in \mathbb{R}_{+}$, and $\boldsymbol{\varpi}$ includes all the DNN parameters $\boldsymbol{\varpi}^i$, $\forall i\in \mathcal{I}_{A}$. 

We refer to the dynamics~\eqref{eq:nominal quadrotor dynamics} and~\eqref{eq:nominal load dynamics} as the control system models, since they employ the virtual control ${\bm u}^l$ (see the definition in subsection~\ref{subsec:centralized mpc}). In contrast, the actual system models~\eqref{eq:quadrotor model} and~\eqref{eq:load model} utilize the hybrid tension magnitudes ${\bm T}=\left [ T^{1},\cdots,T^{n} \right ]$, with each $T^i$ defined in~\eqref{eq:tension magnitude}. We assume that, during practical implementation, the discrepancies between ${\bm u}^l$ and $\bm T$ can be effectively addressed using robust control methods, such as NeuroMHE~\cite{10313083} and $\mathcal{L}_1$ adaptive control~\cite{wu20221}. In this case, the closed-loop states satisfy the control system models. However, note that these actual system models are unsuitable for use in the formulation of Problem~\eqref{eq:closed-loop tuning problem} because they are not differentiable, given the hybrid nature of the tension magnitude $T^i$.

\subsection{Closed-loop Training of Multilift via Gradient Descent}\label{subsec:training via gradient descent}

Problem~\eqref{eq:closed-loop tuning problem} presents a closed-loop training formulation for the multilift system using a bi-level structure. At the lower level, the decomposed MPC problems~\eqref{eq:first qmpc control command} and~\eqref{eq:first lmpc control command} are solved using Algorithm~\ref{alg: distributed mpc} for open-loop prediction. Only their first optimal control commands, ${\bm u}_{0|t}^{\ast,i}$ and ${\bm u}_{0|t}^{\ast,l}$, are then applied to the control system. Here, ${\bm u}_{\cdot |t}^{\ast,i}$ denotes the optimal control trajectory based on the feedback state ${\bm x}_t^i$. At the higher level, we optimize the DNN parameters $\boldsymbol{\varpi}$ to minimize the total loss, thereby evaluating the closed-loop performance of the system. 

To better present the advantages of the closed-loop training, let us compare it with the existing open-loop training approaches~\cite{amos2018differentiable,jin2020pontryagin,jin2021safe}.  Consistent with the MPC settings in the closed-loop training, we continue to employ the adaptive MPC hyperparameters defined in~\eqref{eq:adaptive mpc hyperparameters}. Then, the open-loop training of the multilift system can be formulated as follows:
\begin{subequations}
    \begin{align}
        &\mathop {\min }\limits_{\boldsymbol{\varpi}}\ \bar{L}^{l}+\sum_{i=1}^{n}\bar{L}^{i}\\
        {\rm s.t.}\ & {\xi}^{\ast,i}\ \mathrm{generated}\ \mathrm{by}\ \mathrm{QMPC}\left ( \boldsymbol{\theta }^{i}\left ( \boldsymbol{\varpi }^{i} \right ) \right ),\ \forall i\in \mathcal{I}_{q}\label{eq:qmpc open loop traj}\\
        & {\xi}^{\ast,l}\ \mathrm{generated}\ \mathrm{by}\ \mathrm{LMPC}\left ( \boldsymbol{\theta }^{l}\left ( \boldsymbol{\varpi }^{l} \right ) \right )\label{eq:lmpc open loop traj}
    \end{align}
    \label{eq:open loop tuning problem}%
\end{subequations}
where $\bar{L}^l$ and $\bar{L}^i$ are the individual losses for the load and the $i$-th quadrotor, based on their respective open-loop prediction trajectories $\xi^l$ and $\xi^i$, over a horizon with length of $N_{\rm ol}\in \mathbb{R}_+$. 

\begin{remark}
\label{rm:time indices}
We use different time indices to distinguish between the open-loop prediction within MPC (where $k$ is used) and the closed-loop training (where $t$ is used).    
\end{remark}

By comparison, there are two main advantages of the closed-loop training~\eqref{eq:closed-loop tuning problem} over the open-loop training~\eqref{eq:open loop tuning problem}. Firstly, $N_{\rm ol}$ cannot exceed the MPC horizon $N$, whereas $N_{\rm cl}$ can be longer than $N$, offering greater flexibility in training. Secondly, $\bar{L}^l$ and $\bar{L}^i$ require expert demonstrations $\xi^{{\rm d},l}$ and $\xi^{{\rm d},i}$, which are typically challenging to obtain, limiting the open-loop training to imitation learning. In contrast, $L^l$ and $L^i$ directly evaluate the closed-loop system states using broader criteria, such as tracking performance and collision avoidance, enabling the use of reinforcement learning (RL). 

We aim to solve Problem~\eqref{eq:closed-loop tuning problem} via gradient descent. Note that in the multilift system, the strong cable-induced dynamic couplings among the agents cause their closed-loop states to be influenced not only by their respective controllers but also by the controllers of other agents.  

The influence on one agent’s closed-loop state from the controllers of other agents mainly occurs between each quadrotor and the load. This is because the dynamics of all quadrotors are directly coupled to the load’s dynamics via the cable tensions, whereas they are only indirectly coupled to each other through the load. In other words, the load's controller exerts a more dominant influence on the dynamic behaviors of each quadrotor than do the controllers of the remaining quadrotors. This effect is quantified by the gradient of the quadrotor's closed-loop state with respect to (w.r.t.) the load's MPC hyperparameters: $\frac{\partial {\bm x}{t}^{i}}{\partial \boldsymbol{\theta }^{l}}$. In turn, the gradient $\frac{\partial {\bm x}_{t}^{l}}{\partial \boldsymbol{\theta }^{i}}$ captures the influence of the quadrotor's MPC hyperparameters on the load's closed-loop state.

Therefore, the gradient of the total loss $L$ (defined in \eqref{eq:closed loop total loss}) w.r.t. the quadrotor DNN parameters can be obtained via chain rule as follows:
\begin{equation}
\frac{dL}{d\boldsymbol{\varpi }^{i}} =\left (\frac{\partial L^{i}}{\partial X_{\mathrm cl}^{i}}\frac{\partial X_{\mathrm cl}^{i}}{\partial \boldsymbol{\theta }^{i}} + \frac{\partial L^{l}}{\partial X_{\mathrm cl}^{l}}\frac{\partial X_{\mathrm cl}^{l}}{\partial \boldsymbol{\theta }^{i}}\right )\frac{\boldsymbol{\theta }^{i}}{\partial \boldsymbol{\varpi }^{i}}, \ \forall i\in \mathcal{I}_{q} 
\label{eq:chain rule}
\end{equation}
where $X_{\mathrm {cl}}^{i}=\left \{ {\bm x}_{t}^{i} \right \}_{t=T}^{T+N_{\mathrm {cl}}}$ and $X_{\mathrm {cl}}^{l}=\left \{ {\bm x}_{t}^{l} \right \}_{t=T}^{T+N_{\mathrm {cl}}}$ represent the closed-loop state trajectories for each quadrotor and the load, respectively, starting from a specific time step $T \in \mathbb{R}_+$. Slightly different from the chain rule for the quadrotor, which only needs to account for the coupling effect from the load, the chain rule for computing the gradient of the total loss w.r.t. the load DNN parameters must consider the coupling effects from all the quadrotors, thus leading to the following form
\begin{equation}
\frac{dL}{d\boldsymbol{\varpi }^{l}} =\left (\frac{\partial L^{l}}{\partial X_{\mathrm cl}^{l}}\frac{\partial X_{\mathrm cl}^{l}}{\partial \boldsymbol{\theta }^{l}} + \sum_{i=1}^{n}{\frac{\partial L^{i}}{\partial X_{\mathrm cl}^{i}}\frac{\partial X_{\mathrm cl}^{i}}{\partial \boldsymbol{\theta }^{l}}}\right )\frac{\boldsymbol{\theta }^{l}}{\partial \boldsymbol{\varpi }^{l}}.
\label{eq:chain rule for the load}
\end{equation}

In~\eqref{eq:chain rule} and \eqref{eq:chain rule for the load}, the gradients $\frac{\partial L^{i}}{\partial X_{\mathrm cl}^{i}}$, $\frac{\partial L^{l}}{\partial X_{\mathrm cl}^{l}}$, $\frac{\boldsymbol{\theta }^{i}}{\partial \boldsymbol{\varpi }^{i}}$, and $\frac{\boldsymbol{\theta }^{l}}{\partial \boldsymbol{\varpi }^{l}}$ are straightforward to compute, as both the total loss and the MPC hyperparameters are explicit functions of the closed-loop states and the DNN parameters, respectively. The main challenge lies in solving for $\frac{\partial X_{\mathrm cl}^{i}}{\partial \boldsymbol{\theta }^{i}}$, $\frac{\partial X_{\mathrm cl}^{l}}{\partial \boldsymbol{\theta }^{i}}$, $\frac{\partial X_{\mathrm cl}^{l}}{\partial \boldsymbol{\theta }^{l}}$, and $\frac{\partial X_{\mathrm cl}^{i}}{\partial \boldsymbol{\theta }^{l}}$, the gradients of the closed-loop states w.r.t the MPC hyperparameters. This involves not only differentiating through the nonlinear MPC problems (\ref{eq:first qmpc control command}) and (\ref{eq:first lmpc control command}), but also through the nonlinear system models (\ref{eq:nominal quadrotor dynamics}) and (\ref{eq:nominal load dynamics}). Next, we will demonstrate that these gradients $\frac{\partial X_{\mathrm cl}^{i}}{\partial \boldsymbol{\theta }^{i}}$, $\frac{\partial X_{\mathrm cl}^{l}}{\partial \boldsymbol{\theta }^{i}}$, $\frac{\partial X_{\mathrm cl}^{l}}{\partial \boldsymbol{\theta }^{l}}$, and $\frac{\partial X_{\mathrm cl}^{i}}{\partial \boldsymbol{\theta }^{l}}$ can be efficiently computed in parallel among the quadrotors using the distributed sensitivity propagation method proposed in the following section. 

\section{Distributed Sensitivity Propagation} \label{section:distributed sensitivity propagatiuon}
\subsection{System Sensitivity}\label{subsec:sensitivity}

The closed-loop states ${\bm x}_t^i$ and ${\bm x}_t^l$ are iteratively defined (i.e., propagated) through the system dynamics~\eqref{eq:nominal quadrotor dynamics} and~\eqref{eq:nominal load dynamics}, respectively. Given this, the gradients (sensitivities) $\frac{\partial {\bm x}_{t}^{i}}{\partial \boldsymbol{\theta }^{i}}$ and $\frac{\partial {\bm x}_{t}^{l}}{\partial \boldsymbol{\theta }^{l}}$ can be propagated by taking the partial derivatives w.r.t the corresponding MPC hyperparameters $\boldsymbol{\theta}^i$ and $\boldsymbol{\theta}^l$ on both sides of the respective dynamics equations. 

Let $\mathbf{X}_{i,t}^{i}:= \frac{\partial {\bm x}_{t}^{i}}{\partial \boldsymbol{\theta} ^{i}}\in \mathbb{R}^{13\times m^{i}}$ denote the sensitivity of the $i$-th quadrotor's closed-loop state ${\bm x}_t^i$ w.r.t. its MPC hyperparameters $\boldsymbol{\theta}^i$, $\mathbf{X}_{l,t}^{i}:= \frac{\partial {\bm x}_{t}^{i}}{\partial \boldsymbol{\theta} ^{l}}\in \mathbb{R}^{13\times m^{l}}$ the sensitivity of the $i$-th quadrotor's closed-loop state ${\bm x}_t^i$ w.r.t. the load MPC hyperparameters $\boldsymbol{\theta}^l$, $\mathbf{X}_{l,t}^{l}:= \frac{\partial {\bm x}_{t}^{l}}{\partial \boldsymbol{\theta} ^{l}}\in \mathbb{R}^{13\times m^{l}}$ the sensitivity of the load's closed-loop state ${\bm x}_t^l$ w.r.t. its MPC hyperparameters $\boldsymbol{\theta}^l$, and $\mathbf{X}_{i,t}^{l}:= \frac{\partial {\bm x}_{t}^{l}}{\partial \boldsymbol{\theta} ^{i}}\in \mathbb{R}^{13\times m^{i}}$ the sensitivity of the load's closed-loop state ${\bm x}_t^l$ w.r.t. the $i$-th quadrotor's MPC hyperparameters $\boldsymbol{\theta}^i$. The sensitivity propagation for the multilift system can be defined as follows:
\begin{subequations}
    \begin{align}
    \mathbf{X}_{i,t+1}^{i} &=\mathbf{F}_{t}^{i}\mathbf{X}_{i,t}^{i}+\mathbf{F}_{t}^{il}\mathbf{X}_{i,t}^{l}+\mathbf{G}_{t}^{i}\mathbf{U}_{i,t}^{i}, \ \forall i\in \mathcal{I}_{q} \label{eq:sensitivity quadrotor}\\
    \mathbf{X}_{i,t+1}^{l} &=\mathbf{F}_{t}^{l}\mathbf{X}_{i,t}^{l}+\mathbf{F}_{t}^{li}\mathbf{X}_{i,t}^{i}, \ \forall i\in \mathcal{I}_{q} \label{eq:sensitivity load-quadrotor}\\
    \mathbf{X}_{l,t+1}^{i} &=\mathbf{F}_{t}^{i}\mathbf{X}_{l,t}^{i}+\mathbf{F}_{t}^{il}\mathbf{X}_{l,t}^{l}+\mathbf{G}_{t}^{il}\mathbf{U}_{l,t}^{l}, \ \forall i\in \mathcal{I}_{q}\label{eq:sensitivity quadrotor-load}\\
    \mathbf{X}_{l,t+1}^{l} &=\mathbf{F}_{t}^{l}\mathbf{X}_{l,t}^{l}+\sum_{i=1}^{n}\mathbf{F}_{t}^{li}\mathbf{X}_{l,t}^{i}+\mathbf{G}_{t}^{l}\mathbf{U}_{l,t}^{l}\label{eq:sensitivity load}
    \end{align}
    \label{eq:system sensitivity}%
\end{subequations}
where 
\begin{equation}
    \begin{aligned}
        \mathbf{F}_{t}^{i} &=\frac{\partial \bar{\bm f}_{t}^{i}}{\partial {\bm x}_{t}^{i}}+\frac{\partial \bar{\bm f}_{t}^{i}}{\partial {\bm u}_{0|t}^{\ast,i}}\left ( \frac{{\partial {\bm u}_{0|t}^{\ast,i}}}{\partial {\bm x}_{t}^{i}} +\frac{\partial {\bm u}_{0|t}^{\ast,i}}{\partial {\bm u}_{0|t}^{\ast,l}} \frac{\partial {\bm u}_{0|t}^{\ast,l}}{\partial {\bm x}_{t}^{i}}\right )+\frac{\partial \bar{\bm f}_{t}^{i}}{\partial {\bm u}_{0|t}^{\ast,l}}\frac{\partial {\bm u}_{0|t}^{\ast,l}}{\partial {\bm x}_{t}^{i}},\\
        \mathbf{F}_{t}^{il} &=\frac{\partial \bar{\bm f}_{t}^{i}}{\partial {\bm x}_{t}^{l}}+\frac{\partial \bar{\bm f}_{t}^{i}}{\partial {\bm u}_{0|t}^{\ast,i}}\left ( \frac{{\partial {\bm u}_{0|t}^{\ast,i}}}{\partial {\bm x}_{t}^{l}} +\frac{\partial {\bm u}_{0|t}^{\ast,i}}{\partial {\bm u}_{0|t}^{\ast,l}} \frac{\partial {\bm u}_{0|t}^{\ast,l}}{\partial {\bm x}_{t}^{l}}\right )+\frac{\partial \bar{\bm f}_{t}^{i}}{\partial {\bm u}_{0|t}^{\ast,l}}\frac{\partial {\bm u}_{0|t}^{\ast,l}}{\partial {\bm x}_{t}^{l}},
    \end{aligned}
    \label{eq:coefficient matrices for quadrotor}
\end{equation}
\begin{equation}
    \mathbf{F}_{t}^{l} =\frac{\partial \bar{\bm f}_{t}^{l}}{\partial {\bm x}_{t}^{l}}+\frac{\partial \bar{\bm f}_{t}^{l}}{\partial {\bm u}_{0|t}^{\ast ,l}}\frac{\partial {\bm u}_{0|t}^{\ast ,l}}{\partial {\bm x}_{t}^{l}},\ \mathbf{F}_{t}^{li} =\frac{\partial \bar{\bm f}_{t}^{l}}{\partial {\bm x}_{t}^{i}}+\frac{\partial \bar{\bm f}_{t}^{l}}{\partial {\bm u}_{0|t}^{\ast ,l}}\frac{\partial {\bm u}_{0|t}^{\ast ,l}}{\partial {\bm x}_{t}^{i}},
    \label{eq:coefficient matrices for load}
\end{equation}
\begin{equation}
    \mathbf{G}_{t}^{i} =\frac{\partial \bar{\bm f}_{t}^{i}}{\partial {\bm u}_{0|t}^{\ast ,i}},\ \mathbf{G}_{t}^{il} =\frac{\partial \bar{\bm f}_{t}^{i}}{\partial {\bm u}_{0|t}^{\ast ,i}}\frac{\partial {\bm u}_{0|t}^{\ast ,i}}{\partial {\bm u}_{0|t}^{\ast ,l}} + \frac{\partial \bar{\bm f}_{t}^{i}}{\partial {\bm u}_{0|t}^{\ast ,l}},\ \mathbf{G}_{t}^{l} =\frac{\partial \bar{\bm f}_{t}^{l}}{\partial {\bm u}_{0|t}^{\ast ,l}},
    \label{eq:remaining coefficient matrices}
\end{equation}
$\mathbf{U}_{i,t}^{i}=\frac{\partial {\bm u}_{0|t}^{\ast ,i}}{\partial \boldsymbol{\theta }^{i}}$, and $\mathbf{U}_{l,t}^{l}=\frac{\partial {\bm u}_{0|t}^{\ast ,l}}{\partial \boldsymbol{\theta }^{l}}$. Compared to the sensitivity propagations for single agents~\cite{tao2023difftune,cheng2022difftune,zhang2024inverse}, \eqref{eq:system sensitivity} additionally accounts for the dynamic couplings between each quadrotor and the load, represented by the gradients $\frac{\partial {\bm u}_{0|t}^{\ast,i}}{\partial {\bm u}_{0|t}^{\ast,l}}$, $\frac{\partial {\bm u}_{0|t}^{\ast,i}}{\partial {\bm x}_{t}^{l}}$, $\frac{\partial {\bm u}_{0|t}^{\ast,l}}{\partial {\bm x}_{t}^{i}}$, $\mathbf{X}_{l,t}^{i}$, and $\mathbf{X}_{i,t}^{l}$.

Define ${{\mathbf{\bar X}}_t}:=\left[ {{\mathbf{X}}_{1,t}^1,{\mathbf{X}}_{1,t}^l,{\mathbf{X}}_{l,t}^1, \cdots ,{\mathbf{X}}_{n,t}^n,{\mathbf{X}}_{n,t}^l,{\mathbf{X}}_{l,t}^n,{\mathbf{X}}_{l,t}^l} \right]$ as a new state, ${{\mathbf{\bar U}}_t} := \left[ {{\mathbf{U}}_{1,t}^1, \cdots ,{\mathbf{U}}_{n,t}^n,{\mathbf{U}}_{l,t}^l} \right]$ as a new control. The sensitivity propagation~\eqref{eq:system sensitivity} can be interpreted as the following linear system:
\begin{equation}
{{{\mathbf{\bar X}}}_{t + 1}} = {{{\mathbf{\bar F}}}_t}{{{\mathbf{\bar X}}}_t} + {{{\mathbf{\bar G}}}_t}{{{\mathbf{\bar U}}}_t} \label{eq:linear sensitivity system}
\end{equation}
where
\begin{equation}
\begin{aligned}
{{\mathbf{\bar F}}_t} & = \left[ {\begin{array}{*{20}{c}}
{{\mathbf{F}}_t^1}&{{\mathbf{F}}_t^{1l}}&{\mathbf{0}}&{}&{}&{}&{}&\vline& {\mathbf{0}}\\
{{\mathbf{F}}_t^{l1}}&{{\mathbf{F}}_t^l}&{\mathbf{0}}&{}&{}&{}&{}&\vline& {\mathbf{0}}\\
{\mathbf{0}}&{\mathbf{0}}&{{\mathbf{F}}_t^1}&{}&{}&{}&{}&\vline& {{\mathbf{F}}_t^{1l}}\\
{}&{}&{}& \ddots &{}&{}&{}&\vline&  \vdots \\
{}&{}&{}&{}&{{\mathbf{F}}_t^n}&{{\mathbf{F}}_t^{nl}}&{\mathbf{0}}&\vline& {\mathbf{0}}\\
{}&{}&{}&{}&{{\mathbf{F}}_t^{ln}}&{{\mathbf{F}}_t^l}&{\mathbf{0}}&\vline& {\mathbf{0}}\\
{}&{}&{}&{}&{\mathbf{0}}&{\mathbf{0}}&{{\mathbf{F}}_t^n}&\vline& {{\mathbf{F}}_t^{nl}}\\
\hline
{\mathbf{0}}&{\mathbf{0}}&{{\mathbf{F}}_t^{l1}}& \cdots &{\mathbf{0}}&{\mathbf{0}}&{{\mathbf{F}}_t^{ln}}&\vline& {{\mathbf{F}}_t^l}
\end{array}} \right],\\
{{\mathbf{\bar G}}_t} & = \left[ {\begin{array}{*{20}{c}}
{{\mathbf{G}}_t^1}& \cdots &{\mathbf{0}}&\vline& {\mathbf{0}}\\
{\mathbf{0}}& \cdots &{\mathbf{0}}&\vline& {\mathbf{0}}\\
{\mathbf{0}}& \cdots &{\mathbf{0}}&\vline& {{\mathbf{G}}_t^{1l}}\\
{}& \vdots &{}&\vline&  \vdots \\
{\mathbf{0}}& \cdots &{{\mathbf{G}}_t^n}&\vline& {\mathbf{0}}\\
{\mathbf{0}}& \cdots &{\mathbf{0}}&\vline& {\mathbf{0}}\\
{\mathbf{0}}& \cdots &{\mathbf{0}}&\vline& {{\mathbf{G}}_t^{nl}}\\
\hline
{\mathbf{0}}& \cdots &{\mathbf{0}}&\vline& {{\mathbf{G}}_t^l}
\end{array}} \right],
\end{aligned}
\nonumber
\end{equation}
and $\mathbf{I}$ denotes an identity matrix with an appropriate dimension.

During the interval from $T$ to $T+N_{\mathrm{cl}}$, the new state $\mathbf{\bar X}_t$ is iteratively obtained through the linear system~\eqref{eq:linear sensitivity system}, starting with the zero initial condition $\mathbf{\bar X}_T = \mathbf{0}$. The initial condition is set to zero since the initial state $\mathbf{x}_T$ is independent of the MPC hyperparameters.

\begin{remark}
\label{rm:closed loop state sampled from actual systems}
The system matrices in~\eqref{eq:coefficient matrices for quadrotor}, \eqref{eq:coefficient matrices for load}, and~\eqref{eq:remaining coefficient matrices} are evaluated at the closed-loop states ${\bm x}_{t}^{i}$, $\forall i \in \mathcal{I}_q$, ${\bm x}_t^l$, as well as at the controls ${\bm u}_{0|t}^{\ast,i}$ and ${\bm u}_{0|t}^{\ast,l}$. We allow these closed-loop states to be sampled from the actual system dynamics, where the tension magnitudes are obtained using the hybrid definition in~\eqref{eq:tension magnitude}. This approach permits data collection from a real multilift system for potential online training.    
\end{remark}

In these system matrices, the Jacobians $\frac{\partial \bar{\bm f}_{t}^{i}}{\partial {\bm x}_{t}^{i}}$, $\frac{\partial \bar{\bm f}_{t}^{i}}{\partial {\bm u}_{0|t}^{\ast,i}}$, $\frac{\partial \bar{\bm f}_{t}^{i}}{\partial {\bm u}_{0|t}^{\ast,l}}$, $\frac{\partial \bar{\bm f}_{t}^{i}}{\partial {\bm x}_{t}^{l}}$, $\frac{\partial \bar{\bm f}_{t}^{l}}{\partial {\bm x}_{t}^{l}}$, $\frac{\partial \bar{\bm f}_{t}^{l}}{\partial {\bm u}_{0|t}^{\ast ,l}}$, and $\frac{\partial \bar{\bm f}_{t}^{l}}{\partial {\bm x}_{t}^{i}}$ are obtained directly by differentiating through the system dynamics. These are referred to as dynamics-related gradients. However, the Jacobians $\frac{{\partial {\bm u}_{0|t}^{\ast,i}}}{\partial {\bm x}_{t}^{i}}$, $\frac{\partial {\bm u}_{0|t}^{\ast,i}}{\partial {\bm u}_{0|t}^{\ast,l}}$, $\frac{\partial {\bm u}_{0|t}^{\ast,l}}{\partial {\bm x}_{t}^{i}}$, $\frac{{\partial {\bm u}_{0|t}^{\ast,i}}}{\partial {\bm x}_{t}^{l}}$, and $\frac{\partial {\bm u}_{0|t}^{\ast,l}}{\partial {\bm x}_{t}^{l}}$, as well as the new controls $\mathbf{U}_{i,t}^{i}$ and $\mathbf{U}_{l,t}^{l}$ involves differentiation through the nonlinear MPC problems~\eqref{eq:distributed mpc for quadrotor} and~\eqref{eq:distributed mpc for load}, and thus are classified as MPC-related gradients. We will next demonstrate that these MPC-related gradients can be efficiently calculated by tailoring the Safe-PDP method~\cite{jin2021safe}.

\subsection{Differentiating Nonlinear MPC}\label{subsec:differentiat nmpc}
Safe-PDP~\cite{jin2021safe} efficiently computes the gradient of an MPC's optimal trajectory w.r.t. its hyperparameters while respecting various state and control constraints. This method integrates two techniques. First, it approximates a constrained optimal control problem using an unconstrained counterpart by incorporating constraints into the cost through barrier functions. As discussed in Remark~\ref{rm:soft constraints}, the approximation becomes accurate if a small barrier parameter is used. Second, the gradient of the solution to the resulting unconstrained problem is efficiently computed in a recursive form by implicitly differentiating through Pontryagin's maximum principle (PMP). Safe-PDP was originally developed for open-loop training. To apply it in closed-loop training, that is, to compute the MPC-related gradients, some modifications are needed.

We will use Problem~\eqref{eq:distributed mpc for quadrotor} as an example to demonstrate the derivation process. Since these two MPC problems share the same structure and similar settings (see subsection~\ref{subsec:distributed mpc}), the derivation of the MPC-related gradients for Problem~\ref{eq:distributed mpc for load} can be inferred similarly.

For Problem~\eqref{eq:distributed mpc for quadrotor}, PMP describes a set of first-order optimality conditions which the optimal trajectory $\xi ^{\ast,i}\left ( \boldsymbol{\theta }^{i} \right )=\left \{ X^{\ast,i}\left ( \boldsymbol{\theta }^{i} \right ),  U^{\ast,i}\left ( \boldsymbol{\theta }^{i} \right )\right \}$ should satisfy. To present these conditions, we consider the following Hamiltonian:
\begin{equation}
    H_{k}^{i}= c_{k}^{i}\left ( {\bm x}_{k}^{i},{\bm u}_{k}^{i};{\bm x}_{k}^{j},{\bm x}_{k}^{l},\boldsymbol{\theta }^{i} \right )+{\boldsymbol{\lambda}} _{k+1}^{T}\bar{\bm f}_{k}^{i}\left ( {\bm x}_{k}^{i},{\bm u}_{k}^{i},\Delta t;{\bm x}_{k}^{l},{\bm u}_{k}^{l} \right )
    \label{eq:hamiltonian}
\end{equation}
where $\boldsymbol{\lambda }_{k+1}\in \mathbb{R}^{13}$, $\forall k=1,\cdots,N-1$, denotes the costate variable, which is also known as the Lagrangian multiplier for the constraint of the system model. The running cost $c_k^i$ incorporates the soft constraints using the barrier functions and is defined as
\begin{equation}
\begin{aligned}
   c_{k}^{i} &=\frac{1}{2}\left ( {\bm e}_{x_{k}^{i}}^{T}\mathbf{Q}_{x^{i}}{\bm e}_{x_{k}^{i}} + {\bm e}_{u_{k}^{i}}^{T}\mathbf{Q}_{u^{i}}{\bm e}_{u_{k}^{i}}\right )+\frac{1}{2\gamma } \left ( h_{k}^{i} \right )^{2}-\gamma \ln\left ( -g_{k}^{\mathrm{o},i} \right )\\
   &\ -\gamma \sum_{j\neq i}^{n}\ln\left ( -g_{k}^{j} \right ) -\gamma\sum_{j=1}^{4}\left [\ln\left ( -g_{k,u}^{i,\min} \right ) +\ln\left ( -g_{k,u}^{i,\max}\right )\right ]
\end{aligned}
\label{eq:cost with soft constraints in Hamiltonian}
\end{equation}
where $g_{k,u}^{i,\min}=\left (u_{\min}^{i}  \right )_{j}-\left (u_{k}^{i}  \right )_{j}$, $g_{k,u}^{i,\max}=\left (u_{k}^{i}  \right )_{j}-\left ( u_{\max}^{i} \right )_{j}$, the definitions of $h_{k}^{i}$, $g_{k}^{\mathrm{o},i}$, and $g_{k}^{j}$ are presented below~\eqref{eq:distributed mpc for quadrotor}. Compared with the cost function in~\eqref{eq:distributed mpc for quadrotor}, $c_k^i$ additionally includes the soft constraints for the control input $\bm u_k^i$.

Then, the PMP conditions for the unconstrained approximation of Problem~\eqref{eq:distributed mpc for quadrotor} are determined at $X^{\ast ,i}=\left \{ {\bm x}_{k|t}^{\ast ,i} \right \}_{k=0}^{N}$, $U^{\ast ,i}=\left \{ {\bm u}_{k|t}^{\ast ,i} \right \}_{k=0}^{N-1}$, and $\Lambda ^{\ast ,i}=\left \{ \boldsymbol{\lambda }_{k|t}^{\ast ,i} \right \}_{k=1}^{N}$ (i.e., the optimal costate trajectory). For brevity, $\left ( \boldsymbol{\theta }^{i} \right )$ is omitted after each variable. These conditions are expressed as follows:
\begin{subequations}
    \begin{align}
         \begin{split}
          {\bm x}_{k+1|t}^{\ast ,i} & =\frac{\partial H_{k}^{i}}{\partial \boldsymbol{\lambda }_{k+1|t}^{\ast ,i}}=\bar{\bm f}_{k}^{i}\left ( {\bm x}_{k|t}^{\ast ,i},{\bm u}_{k|t}^{\ast ,i},\Delta t; {\bm x}_{k|t}^{\ast ,l},{\bm u}_{k|t}^{\ast ,l}\right ),\\
          &\quad \forall k=0,\cdots ,N-1,
         \end{split}
         \label{eq:PMP dynamics model constraint} \\
         \begin{split}
          \boldsymbol{\lambda }_{k|t}^{\ast ,i}&=\frac{\partial H_{k}^{i}}{\partial {\bm x}_{k|t}^{\ast,i}}=\frac{\partial c_{k}^{i}}{\partial {\bm x}_{k|t}^{\ast,i}}+\left (\frac{\bar{\bm f}_{k}^{i}}{\partial {\bm x}_{k|t}^{\ast,i}}  \right )^{T}\boldsymbol{\lambda }_{k+1|t}^{\ast ,i},\\
          &\quad \forall k=1,\cdots ,N-1,  
         \end{split}
         \label{eq:PMP costate equation}\\
         \begin{split}
         \mathbf{0}&=\frac{\partial H_{k}^{i}}{\partial {\bm u}_{k|t}^{\ast ,i}}=\frac{\partial c_{k}^{i}}{\partial {\bm u}_{k|t}^{\ast ,i}}+\left (  \frac{\partial \bar{\bm f}_{k}^{i}}{\partial {\bm u}_{k|t}^{\ast ,i}}\right )^{T}\boldsymbol{\lambda }_{k+1|t}^{\ast ,i},\\
         &\quad \forall k=0,\cdots,N-1,
         \end{split}
         \label{eq:PMP input equation}\\
         \boldsymbol{\lambda }_{N|t}^{\ast ,i} & =\frac{\partial c_{N}^{i}}{\partial {\bm x}_{N|t}^{\ast ,i}},\ {\bm x}_{0|t}^{\ast,i}={\bm x}_{t}^{i}
         \label{eq:PMP boundary conditions}
    \end{align}
    \label{eq:PMP conditions}%
\end{subequations}
where $c_N^i$, denoting the terminal cost with the soft constraints, is defined as $c_{N}^{i}=\frac{1}{2}{\bm e}_{x_{N}^{i}}^{T}\mathbf{Q}_{x_{N}^{i}}{\bm e}_{x_{N}^{i}}-\gamma \sum_{j\neq i}^{n}\ln\left ( -g_{N}^{j} \right )+\frac{1}{2\gamma } \left ( h_{N}^{i} \right )^{2}-\gamma \ln\left ( -g_{N}^{\mathrm{o},i} \right )$.  Additionally, ${\bm x}_{k|t}^{\ast,l}$ and ${\bm u}_{k|t}^{\ast,l}$, the optimal state and control solutions to Problem~\eqref{eq:distributed mpc for load} based on the load's feedback state $\bm x_t^l$, are regarded as external signals during the solving of Problem~\eqref{eq:distributed mpc for quadrotor}, as discussed in subsection~\ref{subsec:distributed mpc}.

Recall that our goal is to compute the MPC-related gradients: $\frac{{\partial {\bm u}_{0|t}^{\ast,i}}}{\partial {\bm x}_{t}^{i}}$, $\frac{\partial {\bm u}_{0|t}^{\ast,i}}{\partial {\bm u}_{0|t}^{\ast,l}}$, $\frac{{\partial {\bm u}_{0|t}^{\ast,i}}}{\partial {\bm x}_{0|t}^{\ast,l}}$ (since ${\bm x}_{0|t}^{\ast,l}={\bm x}_t^l$), and $\frac{\partial {\bm u}_{0|t}^{\ast ,i}}{\partial \boldsymbol{\theta }^{i}}$. The PMP conditions~\eqref{eq:PMP conditions} implicitly define the dependence of ${\bm u}_{0|t}^{\ast,i}$ on ${\bm x}_{t}^{i}$, ${\bm u}_{0|t}^{\ast,l}$, ${\bm x}_{0|t}^{\ast,l}$, and $\boldsymbol{\theta }^{i}$. This motivates us to solve for the MPC-related gradients by implicitly differentiating the PMP conditions~\eqref{eq:PMP conditions} on both sides w.r.t. these variables respectively. To unify the presentation, we introduce generalized hyperparameters $\bar {\boldsymbol{\theta}}^i$ that can denote ${\bm x}_{t}^{i}$, ${\bm u}_{0|t}^{\ast,l}$, ${\bm x}_{0|t}^{\ast,l}$, or $\boldsymbol{\theta }^{i}$. This results in the following differential PMP conditions:
\begin{subequations}
    \begin{align}
        \frac{\partial {\bm x}_{k+1|t}^{\ast,i}}{\partial \bar{\boldsymbol{\theta }}^{i}}& ={\mathbf{F}}_{k}\frac{\partial {\bm x}_{k|t}^{\ast,i}}{\partial \bar{\boldsymbol{\theta }}^{i}}+{\mathbf{G}}_{k}\frac{\partial {\bm u}_{k|t}^{\ast,i}}{\partial \bar{\boldsymbol{\theta }}^{i}} + {\mathbf{E}}_{k},\\
        \frac{\partial \boldsymbol{\lambda }_{k|t}^{\ast,i}}{\partial \bar{\boldsymbol{\theta }}^{i}}&={\mathbf{H}}_{k}^{xx}\frac{\partial {\bm x}_{k|t}^{\ast ,i}}{\partial \bar{\boldsymbol{\theta }}^{i}}+{\mathbf{H}}_{k}^{xu}\frac{\partial {\bm u}_{k|t}^{\ast ,i}}{\partial \bar{\boldsymbol{\theta }}^{i}}+ {\mathbf{F}}_{k}^{T}\frac{\partial \boldsymbol{\lambda }_{k+1|t}^{\ast,i}}{\partial \bar{\boldsymbol{\theta }}^{i}} + {\mathbf{H}}_{k}^{x\bar{\theta }},\\
        \mathbf{0}&=\mathbf{H}_{k}^{ux}\frac{\partial {\bm x}_{k|t}^{\ast,i}}{\partial \bar{\boldsymbol{\theta }}^{i}}+\mathbf{H}_{k}^{uu}\frac{\partial {\bm u}_{k|t}^{\ast,i}}{\partial \bar{\boldsymbol{\theta }}^{i}}+\mathbf{G}_{k}^{T}\frac{\partial \boldsymbol{\lambda }_{k+1|t}^{\ast,i}}{\partial \bar{\boldsymbol{\theta }}^{i}} + \mathbf{H}_{k}^{u\bar{\theta }},\\
        \frac{\partial \boldsymbol{\lambda }_{N|t}^{\ast,i}}{\partial \bar{\boldsymbol{\theta}}^{i}}&=\mathbf{H}_{N}^{xx}\frac{\partial {\bm x}_{N|t}^{\ast,i}}{\partial \bar{\boldsymbol{\theta}}^{i}}+\mathbf{H}_{N}^{x\bar{\theta }}, \ \frac{\partial {\bm x}_{0|t}^{\ast,i}}{\partial \bar{\boldsymbol{\theta }}^{i}}=\frac{\partial {\bm x}_{t}^{i}}{\partial \bar{\boldsymbol{\theta }}^{i}}
    \end{align}
    \label{eq:differential PMP}%
\end{subequations}
where the coefficient matrices are defined as follows:
\begin{equation}
    \begin{aligned}
          \mathbf{H}_{k}^{xx}&=\frac{\partial ^{2}H_{k}^{i}}{\partial {\bm x}_{k|t}^{\ast,i}\partial {\bm x}_{k|t}^{\ast,i}}, \mathbf{H}_{k}^{xu}=\frac{\partial ^{2}H_{k}^{i}}{\partial {\bm x}_{k|t}^{\ast,i}\partial {\bm u}_{k|t}^{\ast,i}},\mathbf{F}_{k}  =\frac{\partial \bar{\bm f}_{k}^{i}}{\partial {\bm x}_{k|t}^{\ast,i}},\\
         \mathbf{H}_{k}^{ux}&=\frac{\partial ^{2}H_{k}^{i}}{\partial {\bm u}_{k|t}^{\ast,i}\partial {\bm x}_{k|t}^{\ast,i}}, \mathbf{H}_{k}^{uu}=\frac{\partial ^{2}H_{k}^{i}}{\partial {\bm u}_{k|t}^{\ast,i}\partial {\bm u}_{k|t}^{\ast,i}},\mathbf{G}_{k}  =\frac{\partial \bar{\bm f}_{k}^{i}}{\partial {\bm u}_{k|t}^{\ast,i}},\\
         \mathbf{H}_{N}^{xx}&=\frac{\partial ^{2}c_{N}^{i}}{\partial {\bm x}_{N|t}^{\ast,i}\partial {\bm x}_{N|t}^{\ast,i}}, \mathbf{H}_{k}^{u\bar{\theta }}=\frac{\partial ^{2}H_{k}^{i}}{\partial {\bm u}_{k|t}^{\ast,i}\partial \bar{\boldsymbol{\theta }}^{i}},\mathbf{E}_{k}=\frac{\partial \bar{\bm f}_{k}^{i}}{\partial \bar{\boldsymbol{\theta }}^{i}},\\
         \mathbf{H}_{k}^{x\bar{\theta }}&=\frac{\partial ^{2}H_{k}^{i}}{\partial {\bm x}_{k|t}^{\ast,i}\partial \bar{\boldsymbol{\theta }}^{i}},\mathbf{H}_{N}^{x\bar{\theta }} =\frac{\partial ^{2}c_{N}^{i}}{\partial {\bm x}_{N|t}^{\ast,i}\partial \bar{\boldsymbol{\theta }}^{i}}.
    \end{aligned}
    \label{eq:coefficient matrices for quadrotor pdp}
\end{equation}

\begin{remark}
    \label{rm:centralized gradient}
    There exists an alternative method to compute the MPC-related gradients; specifically, one can differentiate the PMP conditions of the centralized MPC problem~\eqref{eq:centralized mpc} w.r.t. its hyperparameters and solve the resulting differential PMP conditions in a distributed manner. However, this method requires incorporating all the agents' hyperparameters into the centralized hyperparameters, which significantly increases the dimension of the gradient and leads to poor scalability. By comparison, in~\eqref{eq:differential PMP}, we only need to solve for the gradient w.r.t. the individual agent's hyperparameters, substantially improving scalability.
\end{remark}

When $\bar{\boldsymbol{\theta}}^i$ represents different types of hyperparameters, the resulting values of $\frac{\partial {\bm x}_{0|t}^{\ast,i}}{\partial \bar{\boldsymbol{\theta }}^{i}}$, $\mathbf{H}_{k\geq 1}^{x\bar{\theta}}$, $\mathbf{H}_0^{u\bar{\theta}}$, $\mathbf{H}_{k\geq 1}^{u\bar{\theta}}$, $\mathbf{E}_0$, $\mathbf{E}_{k\geq 1}$, and $\mathbf{H}_N^{x\bar{\theta}}$ can vary. The types of $\bar{\boldsymbol{\theta}}^i$ and the respective values of these matrices are summarized in Table~\ref{table:generalized hyperparameters and matrices for the quadrotor}. For the load, the values of the corresponding matrices for $\bar{\boldsymbol{\theta}}^l$ can be inferred similarly and are presented in Appendix-\ref{appendix:matrices for load mpc}.
    \begin{table}[h]
\caption{Coefficient Matrices with $\bar{\boldsymbol{\theta}}^i$ for the $i$-th Quadrotor\label{table:generalized hyperparameters and matrices for the quadrotor}}
\centering
\begin{threeparttable}[t]
\begin{tabular}{ c|c c c c} 
\toprule[1pt]
Matrices  & $\bar{\boldsymbol{\theta}}^i={\boldsymbol{\theta}}^i$ & $\bar{\boldsymbol{\theta}}^i={\bm x}_t^i$ & $\bar{\boldsymbol{\theta}}^i={\bm x}_{0|t}^{\ast,l}$ & $\bar{\boldsymbol{\theta}}^i={\bm u}_{0|t}^{\ast,l}$ \\
\midrule[0.5pt]
$\frac{\partial {\bm x}_{0|t}^{\ast,i}}{\partial \bar{\boldsymbol{\theta }}^{i}}$  & $\mathbf 0$ & $\mathbf I$ & $\mathbf 0$ & $\mathbf 0$ \\
$\mathbf{H}_{k\geq 1}^{x\bar{\theta}}$  & $\neq {\mathbf 0}$ & $\mathbf{0}$ & $\mathbf{0}$ & $\mathbf{0}$  \\
$\mathbf{H}_0^{u\bar{\theta}}$ & $\neq {\mathbf 0}$ & $\mathbf{0}$& $\neq {\mathbf 0}$&$\neq {\mathbf 0}$\\
$\mathbf{H}_{k\geq 1}^{u\bar{\theta}}$ & $\neq {\mathbf 0}$ & $\mathbf{0}$ & $\mathbf{0}$ & $\mathbf{0}$\\
$\mathbf{E}_0$ & $\mathbf{0}$ & $\mathbf{0}$ & $\neq {\mathbf 0}$ & $\neq {\mathbf 0}$\\
$\mathbf{E}_{k\geq 1}$ & $\mathbf{0}$ & $\mathbf{0}$ & $\mathbf{0}$ & $\mathbf{0}$\\
$\mathbf{H}_N^{x\bar{\theta}}$ & $\neq {\mathbf 0}$ & $\mathbf{0}$ & $\mathbf{0}$ & $\mathbf{0}$\\
\bottomrule[0.5pt]
\end{tabular}
\end{threeparttable}
\end{table}

With the above definitions, we can now calculate the MPC-related gradients and present them in a uniform format:
\begin{equation}
    \begin{aligned}
       \frac{\partial {\bm u}_{0|t}^{\ast,i}}{\partial \bar{\boldsymbol{\theta }}^{i}}&=-\left ( \mathbf{H}_{0}^{uu} \right )^{-1}\left\{ {\mathbf{H}_{0}^{ux}\frac{\partial {\bm x}_{0|t}^{\ast ,i}}{\partial \bar{\boldsymbol{\theta }}^{i}}+\mathbf{G}_{0}^{T}\mathbf{W_{1}}+\mathbf{H}_{0}^{u\bar{\theta }}} \right.\\
       &\quad +\mathbf{G}_{0}^{T}\mathbf{P}_{1}\left ( \mathbf{I}+\mathbf{R}_{0}\mathbf{P}_{1} \right )^{-1}\left ( \mathbf{M}_{0}-\mathbf{R}_{0}\mathbf{W}_{1} \right )\\
       &\quad +\left. {\mathbf{G}_{0}^{T}\mathbf{P}_{1}\left ( \mathbf{I}+\mathbf{R}_{0}\mathbf{P}_{1} \right )^{-1}\mathbf{A}_{0}\frac{\partial {\bm x}_{0|t}^{\ast,i}}{\partial \bar{\boldsymbol{\theta }}^{i}}} \right\}
    \end{aligned}
    \label{eq:analytical mpc-related gradient}
\end{equation}
where $\mathbf{A}_{0}=\mathbf{F}_{0}-\mathbf{G}_{0}\left ( \mathbf{H}_{0}^{uu} \right )^{-1}\mathbf{H}_{0}^{ux}$, $\mathbf{R}_{0}=\mathbf{G}_{0}\left ( \mathbf{H}_{0}^{uu} \right )^{-1}\mathbf{G}_{0}^{T}$, and $\mathbf{M}_{0}=\mathbf{E}_{0}-\mathbf{G}_{0}\left ( \mathbf{H}_{0}^{uu} \right )^{-1}\mathbf{H}_{0}^{u\bar{\theta }}$. The matrices $\mathbf{P}_1$ and $\mathbf{W}_1$ are obtained by recursively solving the following equations backward in time ($\forall k=N-1,\cdots,1$), starting with $\mathbf{P}_{N}=\mathbf{H}_{N}^{xx}$ and $\mathbf{W}_{N}=\mathbf{H}_{N}^{x\bar{\theta }}$:
\begin{subequations}
    \begin{align}
        \mathbf{P}_{k}&=\mathbf{Q}_{k}+\mathbf{A}_{k}^{T}\mathbf{P}_{k+1}\left ( \mathbf{I}+\mathbf{R}_{k}\mathbf{P}_{k+1} \right )^{-1}\mathbf{A}_{k},\label{eq:recursion for P}\\
        \begin{split}
            \mathbf{W}_{k}&=\mathbf{A}_{k}^{T} \mathbf{P}_{k+1}\left ( \mathbf{I}+\mathbf{R}_{k}\mathbf{P}_{k+1} \right )^{-1}\left ( \mathbf{M}_{k}-\mathbf{R}_{k}\mathbf{W}_{k+1} \right )\\
            &\quad +\mathbf{A}_{k}^{T}\mathbf{W}_{k+1}+ \mathbf{N}_{k}
        \end{split}
        \label{eq:recursion for W}
    \end{align}
    \label{eq:recursion pmp}%
\end{subequations}
where $\mathbf{Q}_{k}=\mathbf{H}_{k}^{xx}-\mathbf{H}_{k}^{xu}\left ( \mathbf{H}_{k}^{uu} \right )^{-1}\mathbf{H}_{k}^{ux}$, $\mathbf{N}_{k}=\mathbf{H}_{k}^{x\bar{\theta }}-\mathbf{H}_{k}^{xu}\left ( \mathbf{H}_{k}^{uu} \right )^{-1}\mathbf{H}_{k}^{u\bar{\theta }}$, and the definitions of $\mathbf{A}_{k}$, $\mathbf{R}_{k}$, and $\mathbf{M}_{k}$ are consistent with those of $\mathbf{A}_{0}$, $\mathbf{R}_{0}$, and $\mathbf{M}_{0}$, but with their indices updated to $k$.

The theoretical justification for the recursion~\eqref{eq:recursion pmp} is detailed in~\cite{jin2020pontryagin}. Additionally, as the barrier parameter $\gamma$ approaches zero, the gradients of the unconstrained approximation of Problem~\eqref{eq:distributed mpc for quadrotor} can converge toward the gradients of the same problem with arbitrary accuracy. For a detailed discussion of this property, interested readers are referred to~\cite{jin2021safe}.

\subsection{Distributed Sensitivity Propagation Algorithm}\label{subsec:distributed sensitivity algorithm}

\begin{figure}
    \centering
    {\includegraphics[width=0.85\columnwidth]{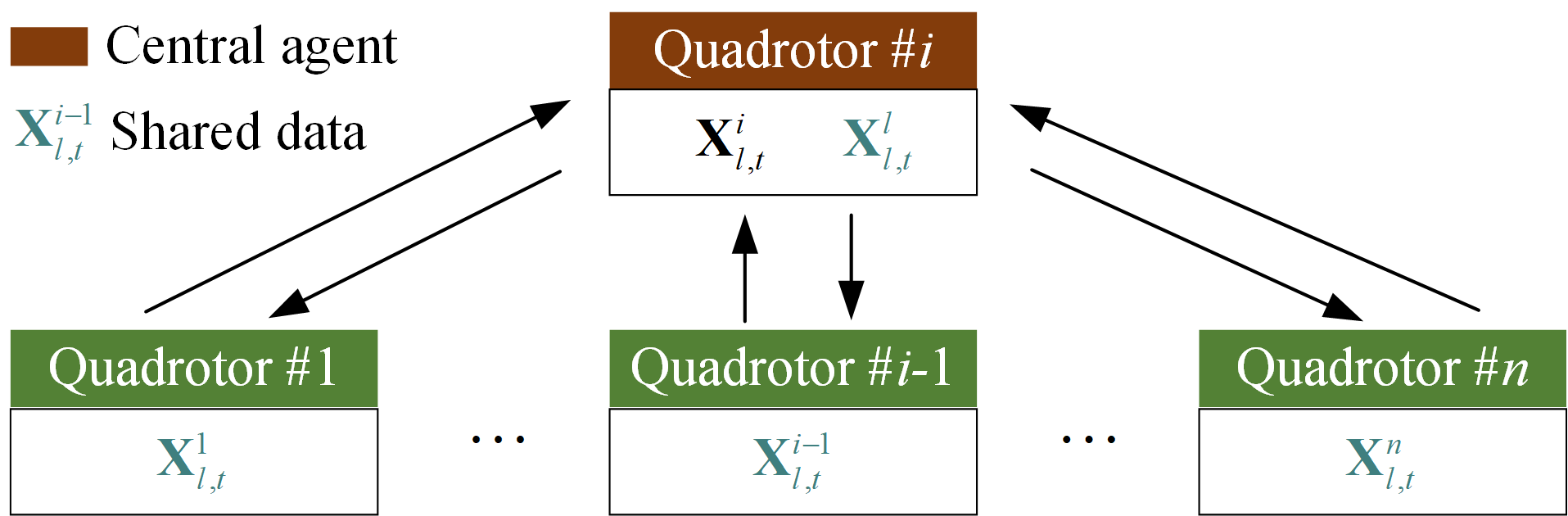}}
    \caption{\footnotesize Diagram of data sharing used in the DSP algorithm. As discussed in Section~\ref{subsec:distributed mpc}, the computation for the load is completed by a randomly selected 'central' agent; therefore, the data are shared between the central quadrotor and the remaining quadrotors.}
    \label{fig:data sharing}
\end{figure}

The linear system~\eqref{eq:linear sensitivity system}, used for the sensitivity propagation, features large yet sparse system matrices $\bar{\mathbf{F}}_t$ and $\bar{\mathbf{G}}_t$. The sparse nature of these matrices, which captures the unique dynamic couplings within the multilift system, facilitates distributed computation. Specifically, the pair of the sensitivities $\mathbf{X}_{i,t}^{i}$ and $\mathbf{X}_{i,t}^{l}$ can be independently propagated through each quadrotor, as $\bar{\mathbf{F}}_t$ lacks coupling terms that would otherwise interconnect this pair with other sensitivities. Although the sensitivities $\mathbf{X}_{l,t}^{i}$ are coupled with the load's sensitivity $\mathbf{X}_{l,t}^{l}$ through the terms $\mathbf{F}_{t}^{il}$ and $\mathbf{F}_{t}^{li}$, we can still compute them in parallel among the quadrotors by sharing data between the load (via a 'central' agent) and the quadrotors at each propagation step, as depicted in Fig.~\ref{fig:data sharing}. 
The complete procedures of distributed sensitivity propagation (DSP) are summarized in Algorithm~\ref{alg: DSP}.
\begin{algorithm}[h]
\caption{Distributed Sensitivity Propagation}
\label{alg: DSP}
\SetKwInput{Input}{Input}
\SetKwInput{Output}{Output}
\Input{The system matrices in~\eqref{eq:coefficient matrices for quadrotor}, \eqref{eq:coefficient matrices for load}, and \eqref{eq:remaining coefficient matrices}, the new controls $\left \{ \bar{\mathbf{U}}_{t}\right \}_{t=T}^{T+N_{\mathrm{cl}}}$, and the initial conditions $\bar{\mathbf{X}}_{T}=\mathbf{0}$.}
\For {$i \leftarrow 1$ \KwTo $n$ (in parallel)}{
\For {$t \leftarrow T$ \KwTo $T+N_{\rm cl}$}{
Receive the load's sensitivity $\mathbf{X}_{l,t}^{l}$ from the central agent;\\
Obtain the sensitivity trajectories $\mathbf{X}_{i,t}^{i}$, $\mathbf{X}_{i,t}^{l}$, and $\mathbf{X}_{l,t}^{i}$ using~\eqref{eq:sensitivity quadrotor},~\eqref{eq:sensitivity load-quadrotor}, and~\eqref{eq:sensitivity quadrotor-load} based on the received $\mathbf{X}_{l,t}^{l}$; \\
Send $\mathbf{X}_{l,t}^{i}$ to the central agent; 
}
}
\For{$t \leftarrow T$ \KwTo $T+N_{\rm cl}$}{
\Comment{In the central agent}
Receive the quadrotors' sensitivities $\mathbf{X}_{l,t}^{i}$, $\forall i\in \mathcal{I}_{q}$;\\
Obtain the load's sensitivity trajectory $\mathbf{X}_{l,t}^{l}$ using \eqref{eq:sensitivity load} based on the received $\mathbf{X}_{l,t}^{i}$;\\
}

\Output{$\bar{\mathbf{X}}_t \left ( \forall t=T,\cdots,T+N_{\mathrm{cl}} \right )$}
\end{algorithm}

\section{Distributed Policy Gradient Algorithm}\label{section:policy gradient}
In this section, we develop a distributed policy gradient RL algorithm to implement Auto-Multilift. This algorithm enables the DNN parameters $\boldsymbol{\varpi}$ to be trained efficiently in a distributed and closed-loop manner.

Trajectory tracking is one of the common practical applications for load delivery using the multilift system. Such trajectories can be effectively planned using off-the-shelf algorithms, such as the minimum snap method~\cite{mellinger2011minimum}, while taking into account the system configuration. To evaluate the tracking performance of the multilift system, we can formulate the individual loss $L^i$ based on the closed-loop tracking errors over the horizon $N_{\mathrm{cl}}$, which takes the following form:
\begin{equation}
    L^{i}=\sum_{t=T}^{T+N_{\mathrm{cl}}}\left \| {\bm x}_{t}^{i}-{\bm x}_{t}^{i,\mathrm{ref}} \right \|_{\mathbf{W}}^{2},\ \forall i\in \mathcal{I}_{A}
    \label{eq:tracking loss}
\end{equation}
where $\bm x_t^{i}$ denotes the closed-loop state sampled from the $i$-th agent's actual dynamics, ${\bm x}_{t}^{i,\mathrm{ref}}$ is the corresponding reference state, and $\mathbf{W}$ is a positive-definite weighting matrix.

\begin{algorithm}[h]
\caption{Distributed Policy Gradient RL}
\label{alg: Distributed policy}
\SetKwInput{Input}{Input}
\SetKwInput{Output}{Output}
\SetKwInput{Forward}{Forward Pass}
\SetKwInput{Backward}{Backward Pass}
\SetKwInput{Initialization}{Initialization}
\SetKw{by}{by}
\Input{The learning rate $\epsilon$ and the pre-planned references ${\bm x}^{i,\mathrm{ref}},\ \forall i\in \mathcal{I}_A$}
\Initialization{$\bm \varpi_0 $}
\While{$L_{\rm mean}$ not converged}{
\For{$t \leftarrow 0$ \KwTo $T_{\mathrm{ep}}$ \by $\Delta \bar{t}$}{
\Forward{}
Obtain the adaptive hyperparameters $\boldsymbol{\theta}_t^{i}$ using~\eqref{eq:adaptive mpc hyperparameters}, $\forall i\in \mathcal{I}_A$;\\
Obtain ${\bm u}_{0|t}^{\ast,i}$, $\forall i\in \mathcal{I}_A$, using Algorithm~\ref{alg: distributed mpc};\\
Compute the individual loss $L_t^i$ using~\eqref{eq:tracking loss}, or~\eqref{eq:obstacle loss}, or a combination of both, $\forall i\in \mathcal{I}_A$;\\
\For {$i \leftarrow 1$ \KwTo $n$ (in parallel)}{
Obtain ${\bm T}_t^{l,i}$ using~\eqref{eq:tension force};\\
Apply ${\bm u}_{0|t}^{\ast,i}$ and $-{\bm T}_t^{l,i}$ to the actual quadrotor model~\eqref{eq:quadrotor model} for updating the closed-loop state ${\bm x}_t^{i}$;\\
}
Apply $\left \{ {\bm T}_{t}^{l,i} \right \}_{i=1}^{n}$ to the actual load model~\eqref{eq:load model} for updating the closed-loop state $\bm{x}_t^l$;\\
\Backward{}
\For {$i \leftarrow 1$ \KwTo $n$ (in parallel)}{Obtain the quadrotor's MPC-related gradients $\frac{\partial {\bm u}_{0|t}^{\ast,i}}{\partial \bar{\boldsymbol{\theta }}^{i}}$ using~\eqref{eq:analytical mpc-related gradient}; \Comment{run in each agent and sent to the central agent}}
Obtain the load's MPC-related gradients $\frac{\partial {\bm u}_{0|t}^{\ast,l}}{\partial \bar{\boldsymbol{\theta }}^{l}}$ using the gradient solver similar to~\eqref{eq:analytical mpc-related gradient}; \Comment{run in the central agent}

\uIf{$t\geq N_{\mathrm{cl}}$}{
Obtain the starting step by $T\leftarrow t-N_{\mathrm{cl}}$;\\
Obtain the system matrices using~\eqref{eq:coefficient matrices for quadrotor},~\eqref{eq:coefficient matrices for load}, and~\eqref{eq:remaining coefficient matrices};\\
Obtain the sensitivities $\bar{\mathbf{X}}_t$ via Algorithm~\ref{alg: DSP};\\
Obtain $\frac{dL}{d\boldsymbol{\varpi }^{i}}$ using~\eqref{eq:chain rule} and \eqref{eq:chain rule for the load}, $\forall i\in \mathcal{I}_A$;\\
Update $\boldsymbol{\varpi}_t$ using gradient-based optimization;
}
\Else{Maintain $\boldsymbol{\varpi}_t$;}
}
Calculate $L_{\mathrm{mean}}=\frac{\Delta \bar{t}}{T_{\mathrm{ep}}}\sum_{i=1}^{n+1}\sum_{k=0}^{\frac{T_{\mathrm{ep}}}{\Delta \bar{t}}}L_{k}^{i}$ for the next episode \Comment{ one training episode}
}
\end{algorithm}

Obstacle avoidance is always required during flights in diverse environments. Although obstacle avoidance constraints can be directly included in the MPC formulation, as demonstrated in~\eqref{eq:centralized mpc}, complex constraints typically make it numerically difficult to solve the resulting MPC problem. By comparison, it is more favorable to plan adaptive references that actively respond to obstacles, thereby facilitating the solution of the MPC problem. To learn such references for the multilift system, the individual loss $L^{i}$ can be chosen to penalize the relative distance between the agent and an obstacle, as follows:
\begin{equation}
    L^{i}=\alpha\sum_{t=T}^{T+N_{\mathrm{cl}}} \mathrm{exp}\left ( -\eta \left \| {\bm p}_{t}^{i}-{\bm p}^{\mathrm{obs}} \right \|_{2} \right ),\ \forall i\in \mathcal{I}_{A}
    \label{eq:obstacle loss}
\end{equation}
where $\alpha \in \mathbb{R}_{+}$ and $\eta  \in \mathbb{R}_{+}$ are positive coefficients, $\bm p_t^{i}$ and $\bm p^{\mathrm{obs}}$ denote the positions of the agent and the obstacle, respectively, in the world frame $\mathcal{I}$. Probably more useful in practice, combining these two kinds of loss,~\eqref{eq:tracking loss} and \eqref{eq:obstacle loss}, into one could achieve more feasibility and flexibility in training.

We learn $\boldsymbol{\varpi}$ through gradient descent. We first obtain the MPC-related gradients using~\eqref{eq:analytical mpc-related gradient}, then compute the system sensitivities using Algorithm~\ref{alg: DSP}, and finally apply the chain rule~\eqref{eq:chain rule} to update $\boldsymbol{\varpi}$ in parallel among the quadrotors. We summarize the training procedures in Algorithm~\ref{alg: Distributed policy}, where $L_{\mathrm{mean}}$ is the mean value of the sum of the individual losses over one training episode with the duration $T_{\mathrm{ep}}$, $\Delta \bar{t}$ is the time step used to discretize the actual system dynamics, and $L_k^{n+1}$ denotes the load's loss.

\section{Experiments} \label{section: simulation}
We validate the effectiveness of Auto-Multilift via extensive simulations aimed at learning different MPC hyperparameters across various flight scenarios. In particular, we will show the following advantages of Auto-Multilift. First, Auto-Multilift is able to learn adaptive MPC weightings directly from trajectory tracking errors. Additionally, it significantly improves training stability and tracking performance over a state-of-the-art open-loop learning method. Finally, beyond its improved training ability to learn adaptive MPC weightings, our method can effectively learn an adaptive tension reference, enabling the multilift system to reconfigure itself when traversing through obstacles.

To enhance the fidelity of the simulation, we consider several realistic aspects. First, in the spring-damper tension model~\eqref{eq:tension magnitude}, we select the stiffness and the damping ratio to be $K=4000 \ {\rm{N/m}}$ and $c_t=0.01\ {\rm{s}}$, respectively. The extremely high stiffness is used to reflect the potential for sudden tension changes, a common phenomenon in the cable-suspended load transportation, as will be observed in the subsequent simulations. Second, the quadrotor's control ${\bm u}^{i}=\left [ f^{i},\boldsymbol{\tau }^{i} \right ]$ is generated by the angular speed of the propeller, which cannot change immediately. Given this delay, we model the dynamics of the quadrotor control using the following first-order system:
\begin{equation}
    \frac{d}{dt}{\bm u}^{i}=\frac{1}{\tau _{\Omega }}\left ( {\bm u}_{\mathrm{cmd}}^{i}-{\bm u}^{i} \right )
    \label{eq:quadrotor control dynamics}
\end{equation}
where ${\bm u}_{\mathrm{cmd}}^{i}$ is the commanded control input generated by solving the quadrotor MPC problem~\eqref{eq:distributed mpc for quadrotor} and $\tau _{\Omega }$ is the motor time constant. Throughout all the simulation experiments, the time constant is chosen as $\tau _{\Omega }=33\ {\rm {ms}}$, as used in~\cite{bauersfeld2021neurobem}. Finally, the actual models~\eqref{eq:quadrotor model} and~\eqref{eq:load model} are discretized using the 4th order Runge-Kutta method with a time step $\Delta \bar t= 0.005\ {\rm s}$, whereas the control models~\eqref{eq:nominal quadrotor dynamics} and~\eqref{eq:nominal load dynamics}, used in MPCs, are discretized using the same method but with a larger time step of $\Delta t= 0.02\ {\rm s}$. This larger time step is employed because the control system typically updates at a frequency that is less than that of the physical systems, accommodating computational limitations or strategic design choices in the control algorithm.

We utilize the following parameterization to constrain the adaptive MPC hyperparameters $\boldsymbol{\theta }^{i}\left ( \boldsymbol{\varpi} ^{i} \right )$ within a specific range:
\begin{equation}
    \boldsymbol{\theta }^{i}=\boldsymbol{\theta }_{\min}^{i}+\left ( \boldsymbol{\theta }_{\max}^{i}-\boldsymbol{\theta }_{\min}^{i} \right )\boldsymbol{\Theta }^{i},\ \forall i\in \mathcal{I}_{A}
    \label{eq:parameterization}
\end{equation}
where $\boldsymbol{\theta }_{\min}^{i}$ and $\boldsymbol{\theta }_{\max}^{i}$ denote the lower and upper bounds, respectively, while $\boldsymbol{\Theta }^{i}$ represents the normalized hyperparameters that range from $0$ to $1$. This parameterization is essential for different types of $\boldsymbol{\theta }^{i}\left ( \boldsymbol{\varpi} ^{i} \right )$. For instance, when $\boldsymbol{\theta }^{i}\left ( \boldsymbol{\varpi} ^{i} \right )$ serves as the adaptive MPC weightings in its cost function, a positive lower bound ensures the positive definiteness of these weighting matrices, provided that they are diagonal. The upper bound prevents these weightings from growing to infinity. In addition, when $\boldsymbol{\theta }^{i}\left ( \boldsymbol{\varpi} ^{i} \right )$ serves as the adaptive MPC references, such as tension references, the specified range accommodates the physical limitations of the system, making the references dynamically feasible. 

\begin{figure}
    \centering
    {\includegraphics[width=0.8\columnwidth]{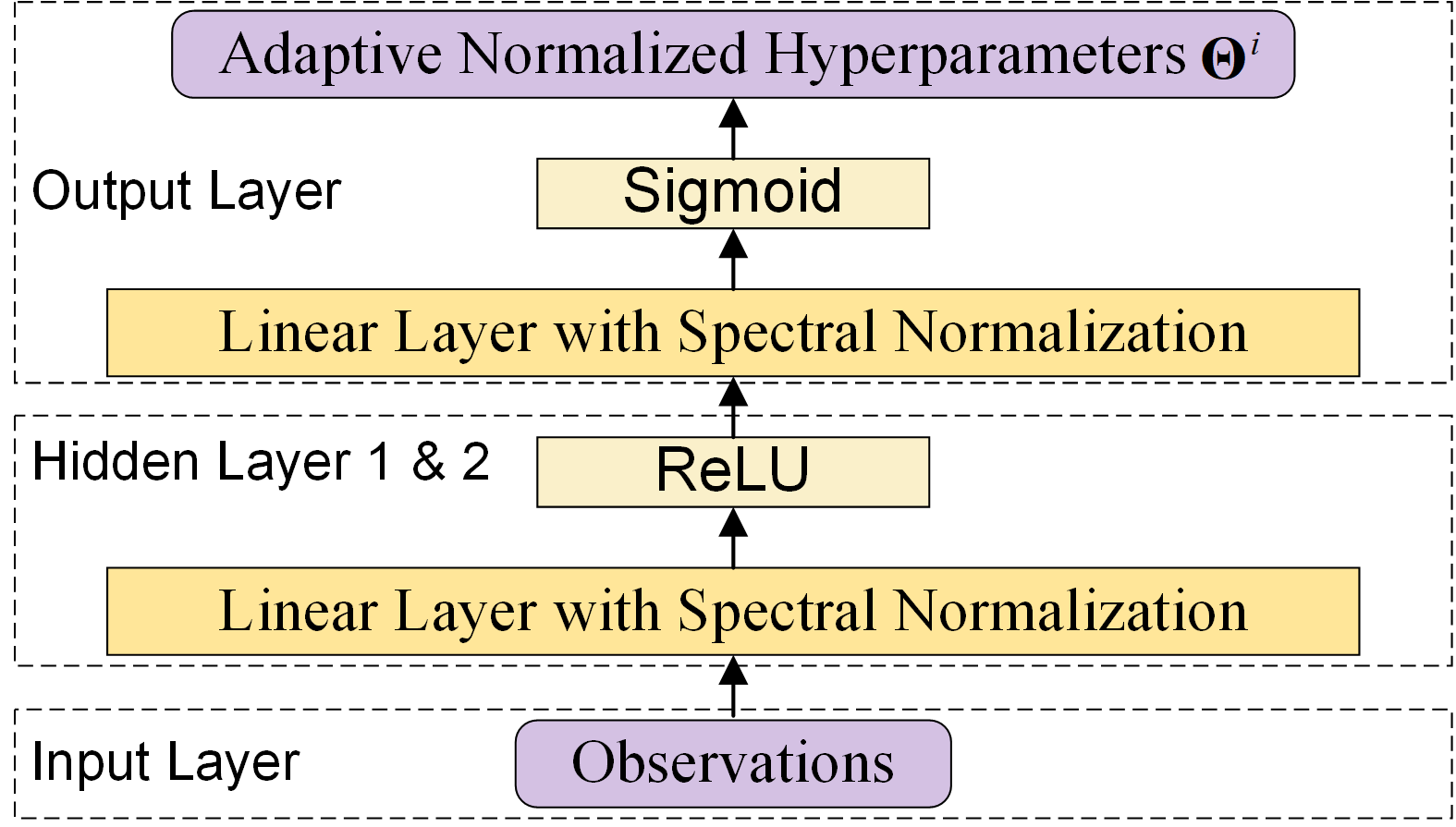}}
    \caption{\footnotesize Architecture of the neural network for producing the adaptive MPC normalized hyperparameters $\boldsymbol{\Theta}^i$ online. As the network's input, the observations can include the system tracking errors or obstacle information, depending on the applications.}
    \label{fig:network structure}
\end{figure}

A multilayer perception (MLP) network is adopted to generate the adaptive normalized hyperparameters $\boldsymbol{\Theta }^{i}$ online. The network's output layer uses a Sigmoid activation function to ensure outputs between $0$ and $1$. Fig.~\ref{fig:network structure} illustrates the complete architecture of the network, which includes two hidden layers using the rectified linear unit (ReLU) activation function. The linear layer employs spectral normalization, a technique that enhances the network's robustness and generalizability by constraining the Lipschitz constant of the layer~\cite{bartlett2017spectrally}.

To reflect the above parameterization, the chain rules \eqref{eq:chain rule} and \eqref{eq:chain rule for the load} for the gradient computation are therefore modified as follows:
\begin{subequations}
    \begin{align}
        \frac{dL}{d\boldsymbol{\varpi }^{i}} &= \left (\frac{\partial L}{\partial X_{\mathrm cl}^{i}}\frac{\partial X_{\mathrm cl}^{i}}{\partial \boldsymbol{\theta }^{i}} + \frac{\partial L}{\partial X_{\mathrm cl}^{l}}\frac{\partial X_{\mathrm cl}^{l}}{\partial \boldsymbol{\theta }^{i}}\right )\frac{\partial \boldsymbol{\theta }^{i}}{\partial \boldsymbol{\Theta }^{i}}\frac{\partial \boldsymbol{\Theta }^{i}}{\partial \boldsymbol{\varpi }^{i}},\ \forall i\in \mathcal{I}_{q},\label{eq:modified chain rule for quadrotor}\\
        \frac{dL}{d\boldsymbol{\varpi }^{l}} &= \left (\frac{\partial L}{\partial X_{\mathrm cl}^{l}}\frac{\partial X_{\mathrm cl}^{l}}{\partial \boldsymbol{\theta }^{l}} + \sum_{i=1}^{n}{\frac{\partial L}{\partial X_{\mathrm cl}^{i}}\frac{\partial X_{\mathrm cl}^{i}}{\partial \boldsymbol{\theta }^{l}}}\right )\frac{\partial \boldsymbol{\theta }^{l}}{\partial \boldsymbol{\Theta }^{l}}\frac{\partial \boldsymbol{\Theta }^{l}}{\partial \boldsymbol{\varpi }^{l}}.\label{eq:modified chain rule for load}
    \end{align}
    \label{eq:modified chain rule}%
\end{subequations}
We implement our method in Python and solve the distributed MPC problems described in Algorithm~\ref{alg: distributed mpc} using the ACADOS toolkit~\cite{Verschueren2021}. The MLP, illustrated in Fig.~\ref{fig:network structure}, is built using PyTorch~\cite{paszke2019pytorch} and trained using \texttt{Adam}~\cite{kingma2015adam}. During the implementation, we customize the loss function $L^i$, which can be~\eqref{eq:tracking loss} or~\eqref{eq:obstacle loss}, or a combination of both, to align with the typical training procedure in PyTorch. Specifically, let us take the customized loss function for training the quadrotor network as an example, which is defined as $L_{{\mathrm{pytorch}}}^i = {\left. {\frac{{d{L}}}{{d{{\boldsymbol{\Theta }}^i}}}} \right|_{{\boldsymbol{\Theta }}_t^i}}{{\boldsymbol{\Theta }}^i},\ \forall i\in \mathcal{I}_{q}$. Here ${\left. {\frac{{d{L}}}{{d{{{\boldsymbol{\Theta} }}^i}}}} \right|_{{{\boldsymbol{\Theta} }}_t^i}} = \left ({\left. {\frac{{\partial {L}}}{{\partial X_{{\mathrm{cl}}}^i}}} \right|_{X_{{\mathrm{cl}},t}^i}}{\left. {\frac{{\partial X_{{\mathrm{cl}}}^i}}{{\partial {\boldsymbol{\theta} ^i}}}} \right|_{\boldsymbol{\theta} _t^i}} + {\left. {\frac{{\partial {L}}}{{\partial X_{{\mathrm{cl}}}^l}}} \right|_{X_{{\mathrm{cl}},t}^l}}{\left. {\frac{{\partial X_{{\mathrm{cl}}}^l}}{{\partial {\boldsymbol{\theta} ^i}}}} \right|_{\boldsymbol{\theta} _t^i}} \right){\left. {\frac{{\partial {\boldsymbol{\theta} ^i}}}{{\partial {\boldsymbol{\Theta} ^i}}}} \right|_{\boldsymbol{\Theta} _t^i}}$ represents the gradient of $L$ w.r.t. $\boldsymbol{\Theta}^i$ evaluated at $\boldsymbol{\Theta}_t^i$. This ensures that $\frac{dL_{\mathrm{pytorch}}^{i}}{d\boldsymbol{\varpi }^{i}}=\frac{dL}{d\boldsymbol{\varpi }^{i}}$. The customized loss function for training the load network can be defined accordingly.

In the subsequent simulations, we focus on learning adaptive MPC weightings and references, assuming that the multi-lift system model can be obtained through system identification, such as the method proposed in~\cite{geng2021estimation}.

\subsection{Distributed Learning of Adaptive Weightings}\label{subsec:learning weightings}

We design a trajectory tracking flight scenario to demonstrate the first and second advantages of our method. We consider a challenging multilift system with a non-uniform load mass distribution, resulting in a biased load CoM coordinate within its body frame. In this simulation, the load CoM coordinate is set to ${\bm r}^{g}=\left [ 0.1,0.1,-0.1 \right ]\ {\rm m}$. Under these conditions, balancing the load during flight requires uneven and dynamic tension allocation among the quadrotors, making manual selection of the MPC weightings extremely difficult. Assuming no obstacles during flight, we focus on learning adaptive MPC weightings that enable the multilift system to track the reference trajectory while stabilizing the load's attitude. To this end, the individual loss function $L^i$ is defined using~\eqref{eq:tracking loss}. We plan circular reference trajectories for each agent using the minimum snap algorithm~\cite{mellinger2011minimum}, based on a fixed system configuration\footnote{The configuration refers to the tilt angle of each cable relative to the vertical direction. The tension references are computed based on the load's reference trajectory and then evenly distributed among the quadrotors.}. To reduce the network size, we constrain the MPC weighting matrices to be diagonal, resulting in a total of $28$ weightings for each quadrotor and $24+n$ weightings for the load (where $n$ is the number of quadrotors). We set the number of neurons in the hidden layer to $30$ and take the tracking errors as input. Finally, the lower and upper bounds for the adaptive MPC weightings are set to $0.01$ and $100$, respectively.

\subsubsection{Ablation Study}

\begin{figure}[h]
    \centering
    {\includegraphics[width=0.9\columnwidth]{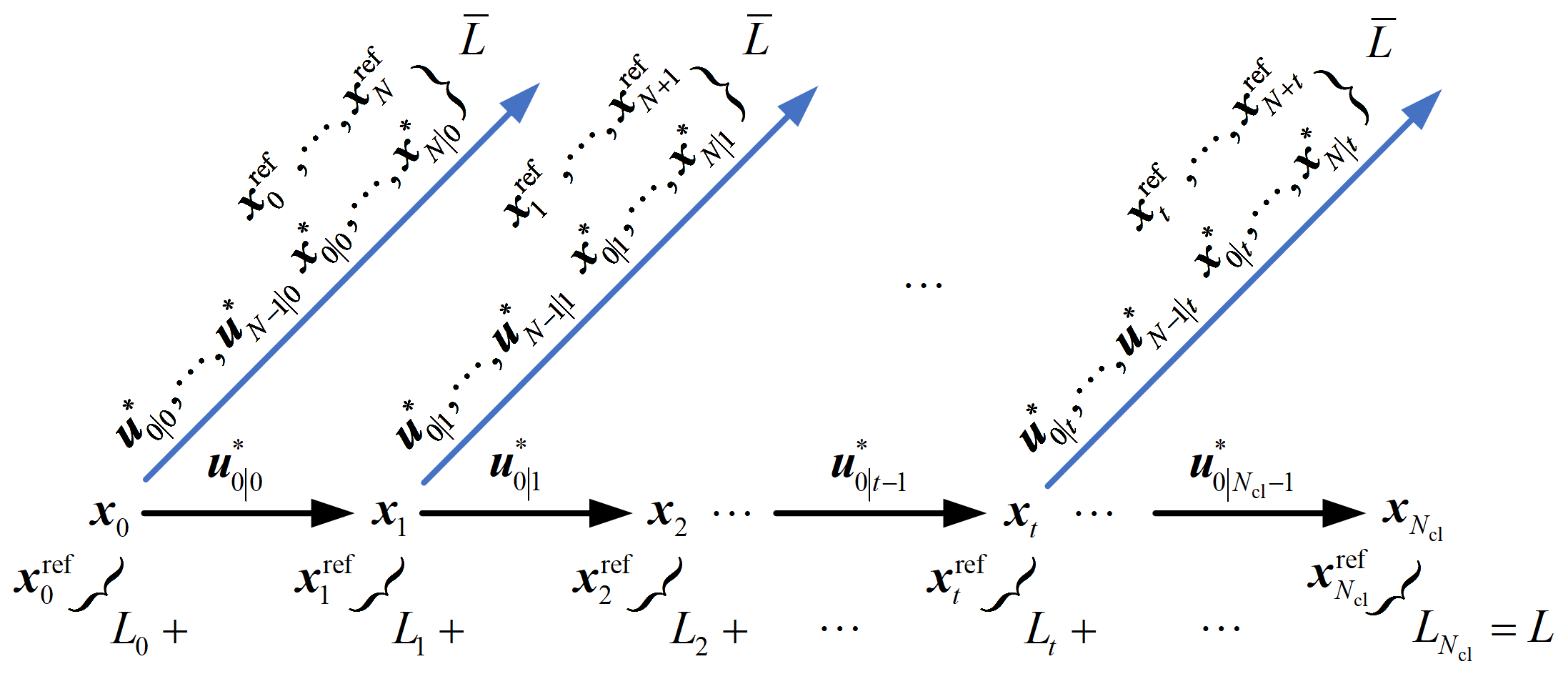}}
    \caption{Comparison of the training data between our method and the open-loop Safe-PDP method. The blue arrows indicate the MPC open-loop prediction trajectories at each time step, which are compared with the reference trajectories to form the open-loop losses (denoted by $\bar L$) that are minimized by the Safe-PDP method. The black arrows represent the transfers of the closed-loop states under the MPC policies.}
    \label{fig:data flow}
\end{figure}

We then validate the learning capability of our method using a multilift system with six quadrotors transporting a $10 \ {\mathrm{kg}}$ load. In this case, we need to train a total of seven networks (six for the quadrotors and one for the load). To highlight the important role that the DSP plays in our method, we compare it with the state-of-the-art open-loop (OL) training method, Safe-PDP~\cite{jin2021safe}. The open-loop method relies solely on the MPC open-loop prediction trajectories, completely removing the DSP algorithm from the training process, as detailed in \eqref{eq:open loop tuning problem}. In contrast, the training of our method uses the closed-loop state trajectories. This difference in training data is illustrated in Fig.~\ref{fig:data flow}.
Additionally, compared to closed-loop MPC training methods for single agents, such as those presented in~\cite{tao2023difftune,zhang2024inverse}, our DSP algorithm accounts for the dynamic couplings between each quadrotor and the load, as discussed in \eqref{subsec:sensitivity}. To demonstrate the effects of these couplings on learning performance, we remove them from the DSP and refer to the remaining method as the 'closed-loop without couplings' (CLWC) method. \autoref{auto:table:ablation study} summaries and compares the components of the three methods used in the ablation study.
\begin{table}[h]
\caption{Methods for Ablation Study  \label{auto:table:ablation study}}
\centering
\begin{threeparttable}[t]
\begin{tabular}{ c|c |c |c } 
\toprule[1pt]
\multirow{2}{*}{Method}  & \multirow{2}{*}{Safe-PDP} & \multicolumn{2}{c}{DSP Algorithm}  \\
\cline{3-4}
   & & Sensitivity Prop. & Couplings \\
\midrule[0.5pt]
 Our method  & \checkmark & \checkmark & \checkmark \\
 CLWC method & \checkmark & \checkmark &  \\
OL method & \checkmark &  & \\
\bottomrule[0.5pt]
\end{tabular}
\begin{tablenotes}[flushleft]
      \footnotesize
      \item The couplings refer to $\frac{\partial {\bm u}_{0|t}^{\ast,i}}{\partial {\bm u}_{0|t}^{\ast,l}}$, $\frac{\partial {\bm u}_{0|t}^{\ast,i}}{\partial {\bm x}_{t}^{l}}$, $\frac{\partial {\bm u}_{0|t}^{\ast ,l}}{\partial {\bm x}_{t}^{i}}$, $\mathbf{X}_{l,t}^{i}$, and $\mathbf{X}_{i,t}^{l}$.
\end{tablenotes}
\end{threeparttable}
\end{table}

We set $N=10$ as the MPC horizon for open-loop prediction and $N_{\mathrm{cl}}=20$ as the loss horizon for closed-loop training. The distributed MPC problems \eqref{eq:distributed mpc for quadrotor} and \eqref{eq:distributed mpc for load} are solved using Algorithm~\ref{alg: distributed mpc} in a receding horizon manner. One training episode involves training these networks for each agent in parallel over their $15$-s-long circular reference trajectories once (i.e., $T_{\mathrm{ep}}=15\ {\mathrm{s}}$ in Algorithm~\ref{alg: Distributed policy}). When strictly aligning with the open-loop training method in~\cite{jin2021safe}, one can only implement Algorithm~\ref{alg: distributed mpc} once per episode. In that case, an extremely long MPC horizon is required to cover the whole trajectory, which is computationally expensive and often infeasible. Given this limitation, in the open-loop training for the ablation study, we continue to implement Algorithm~\ref{alg: distributed mpc} in a receding horizon manner, as done in the closed-loop training, and solve Problem \eqref{eq:open loop tuning problem} using the Safe-PDP method with $N_{\mathrm{ol}}=10$. Note that in Problem \eqref{eq:open loop tuning problem}, the individual losses are based on their respective open-loop prediction trajectories from the MPCs (See Fig.~\ref{fig:data flow}).

\begin{figure}[h]
    \centering
    {\includegraphics[width=0.75\columnwidth]{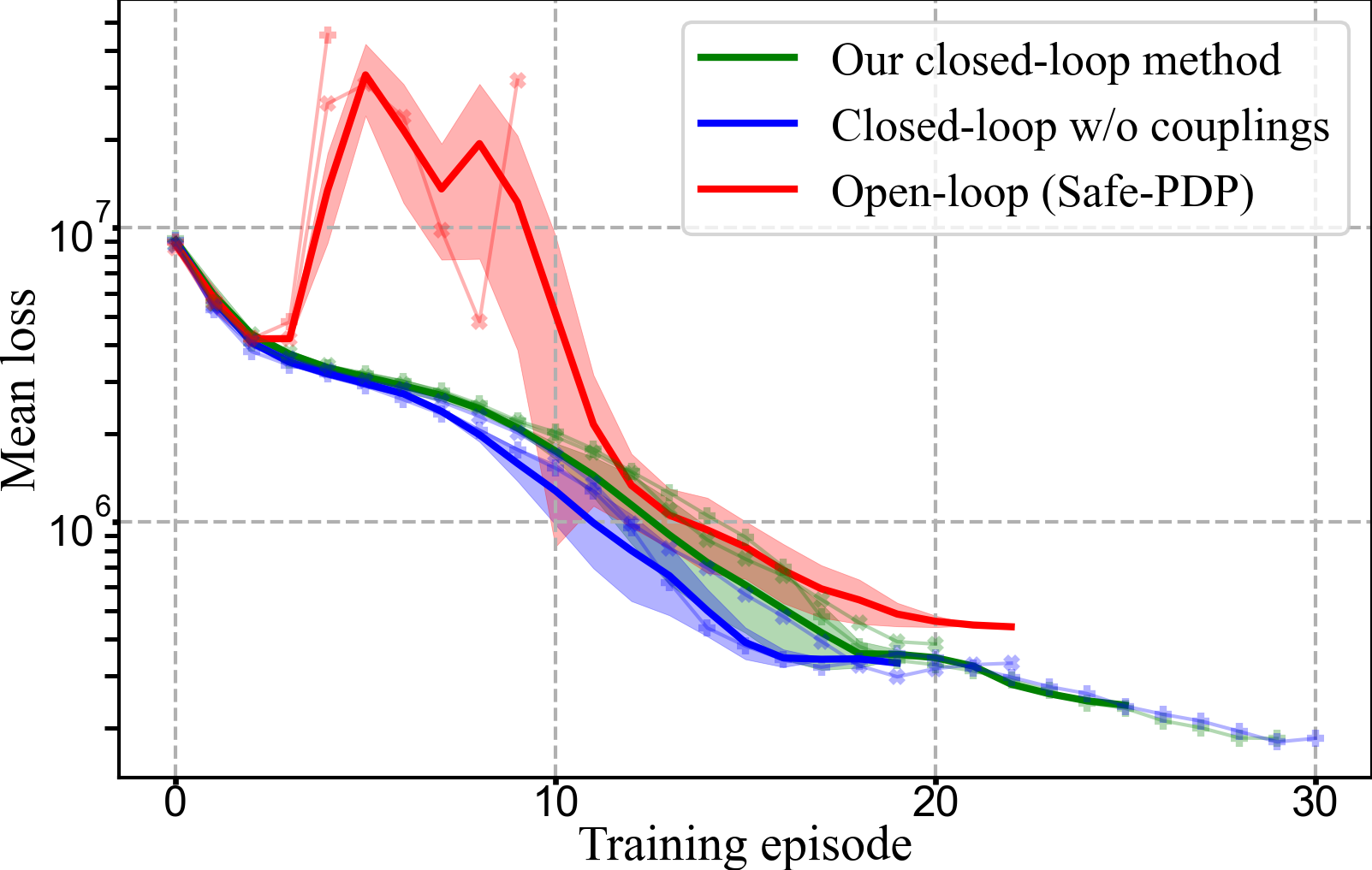}}
    \caption{Comparison of the mean loss among the three methods across 5 groups. These methods are implemented using the same stopping criterion. Specifically, the training will stop when $\left | L_{\mathrm{mean}}^{k}-L_{\mathrm{mean}}^{k-1} \right |\leq \frac{1}{1000}L_{\mathrm{mean}}^{0},\ \forall k\geq 1$, where $L_{\mathrm{mean}}^{k}$ denotes the mean loss at iteration $k$. The mean value (solid line) and the standard deviation (shaded area) of each method are computed using data from the first three groups. The symbols '$+$' and '$\times$' denote the mean losses for the 4th and 5th groups, respectively.}
    \label{fig:meanloss_comparison}
\end{figure}

\begin{figure*}[t]
\centering
\begin{subfigure}[b]{0.325\textwidth}
\centering
\includegraphics[width=0.85\textwidth]{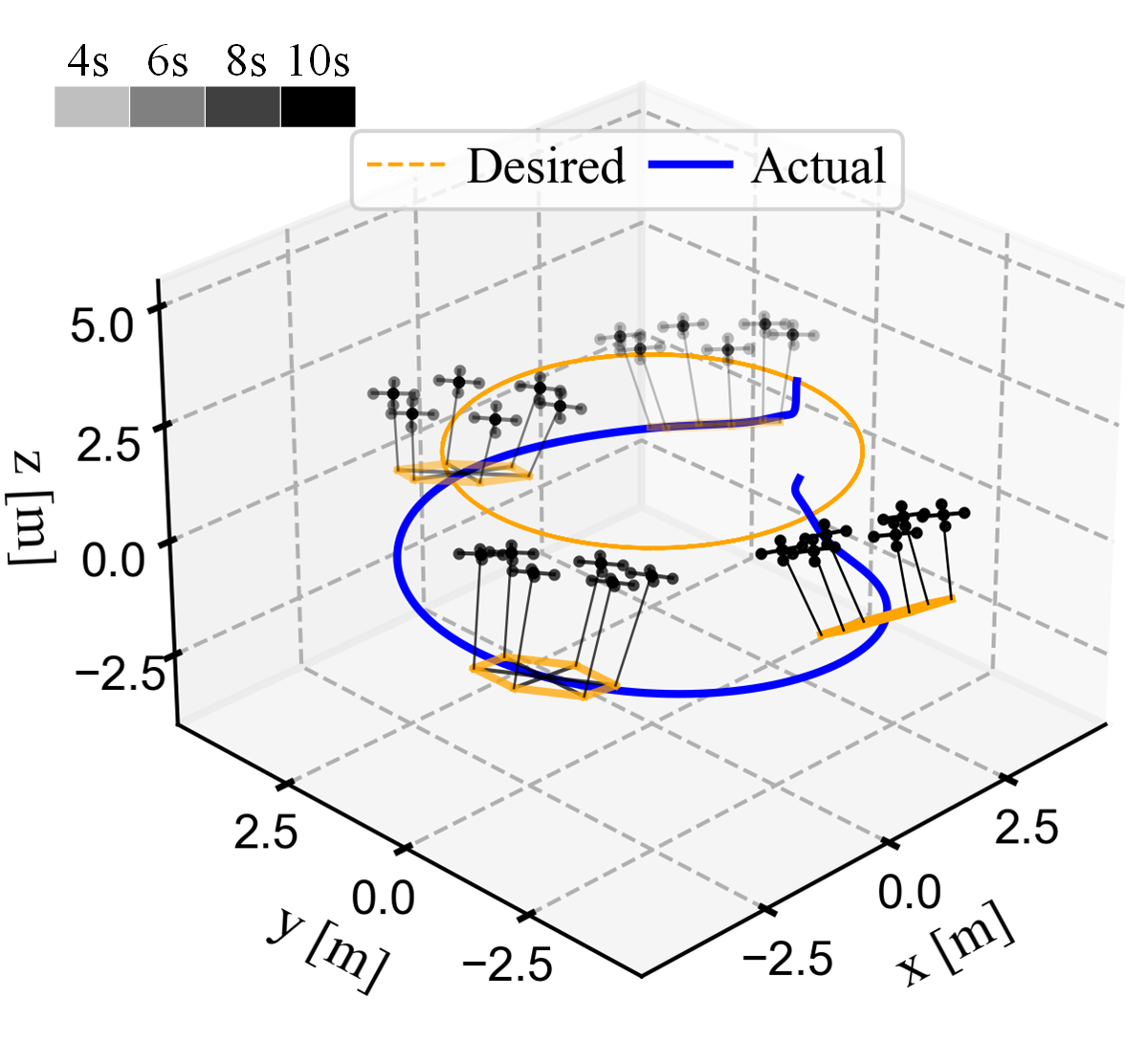}
\caption{Episode $0$ of the OL method.}
\label{fig:6 quad episode 0_ol}
\end{subfigure}
\hfill
\begin{subfigure}[b]{0.325\textwidth}
\centering
\includegraphics[width=0.85\textwidth]{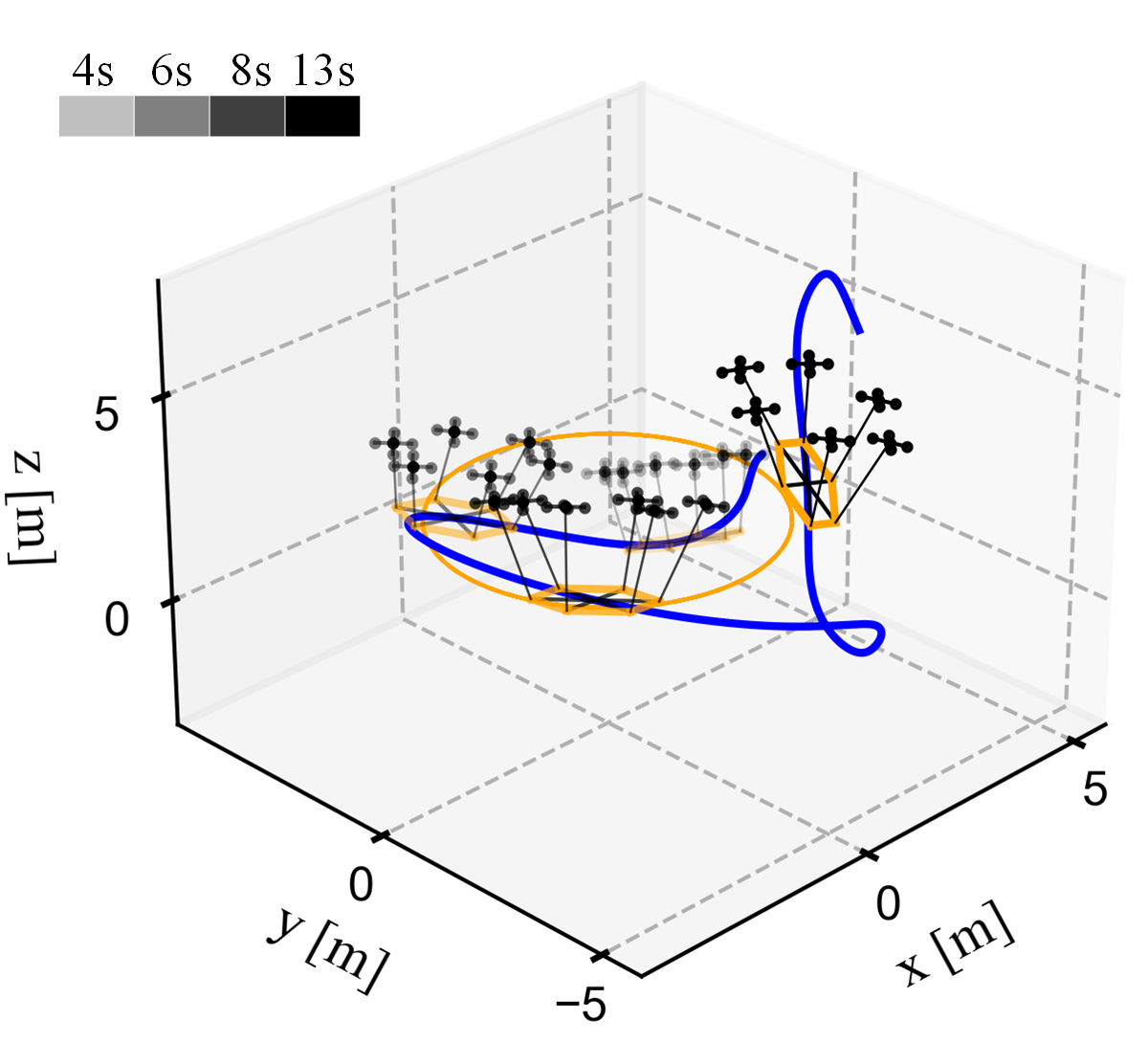}
\caption{Episode $4$ of the OL method.}
\label{fig:6 quad episode 4_ol}
\end{subfigure}
\hfill
\begin{subfigure}[b]{0.325\textwidth}
\centering
\includegraphics[width=0.85\textwidth]{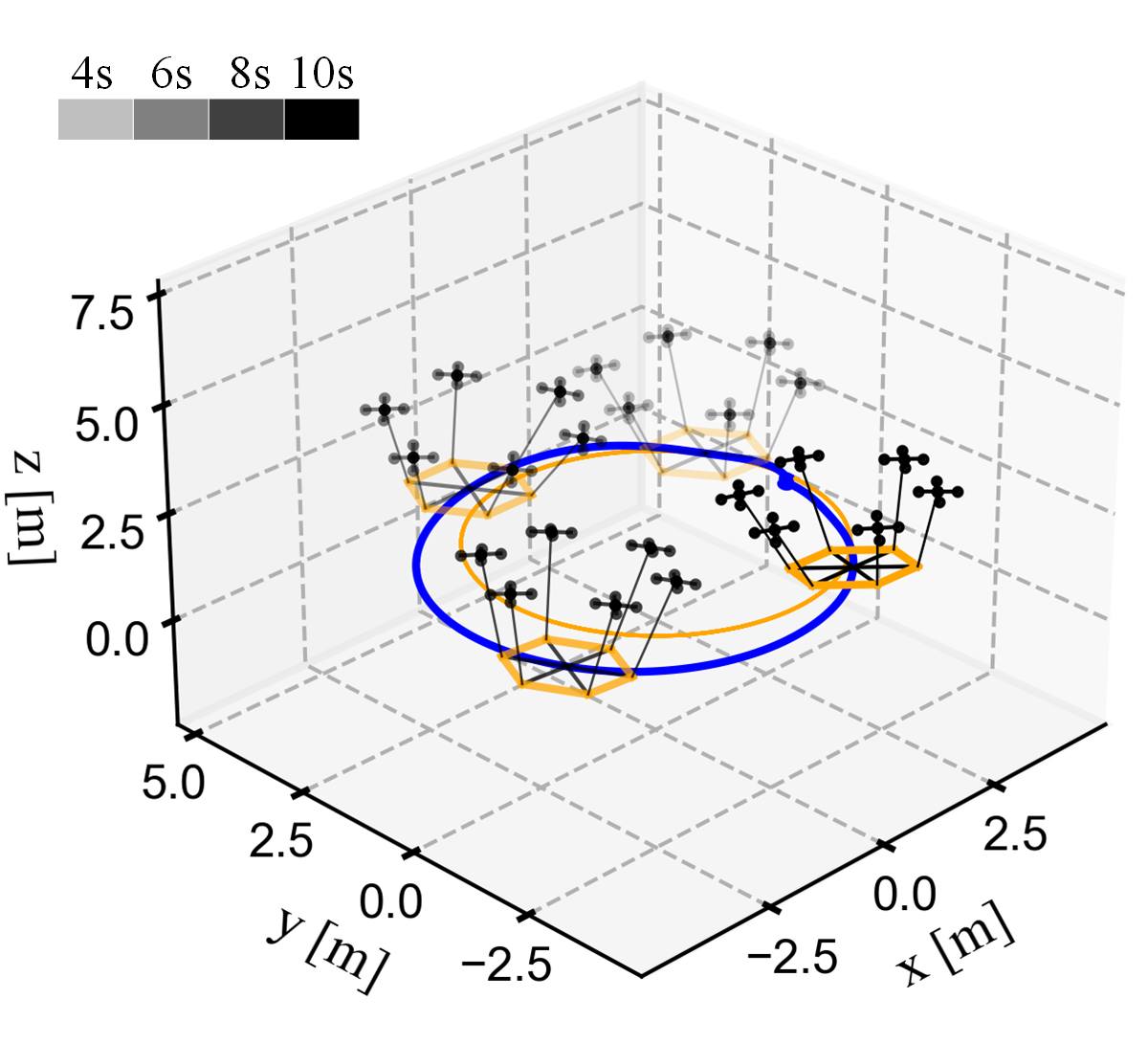}
\caption{Episode $18$ of the OL method.}
\label{fig:6 quad episode 18_ol}
\end{subfigure}
\vskip\baselineskip
\centering
\begin{subfigure}[b]{0.325\textwidth}
\centering
\includegraphics[width=0.85\textwidth]{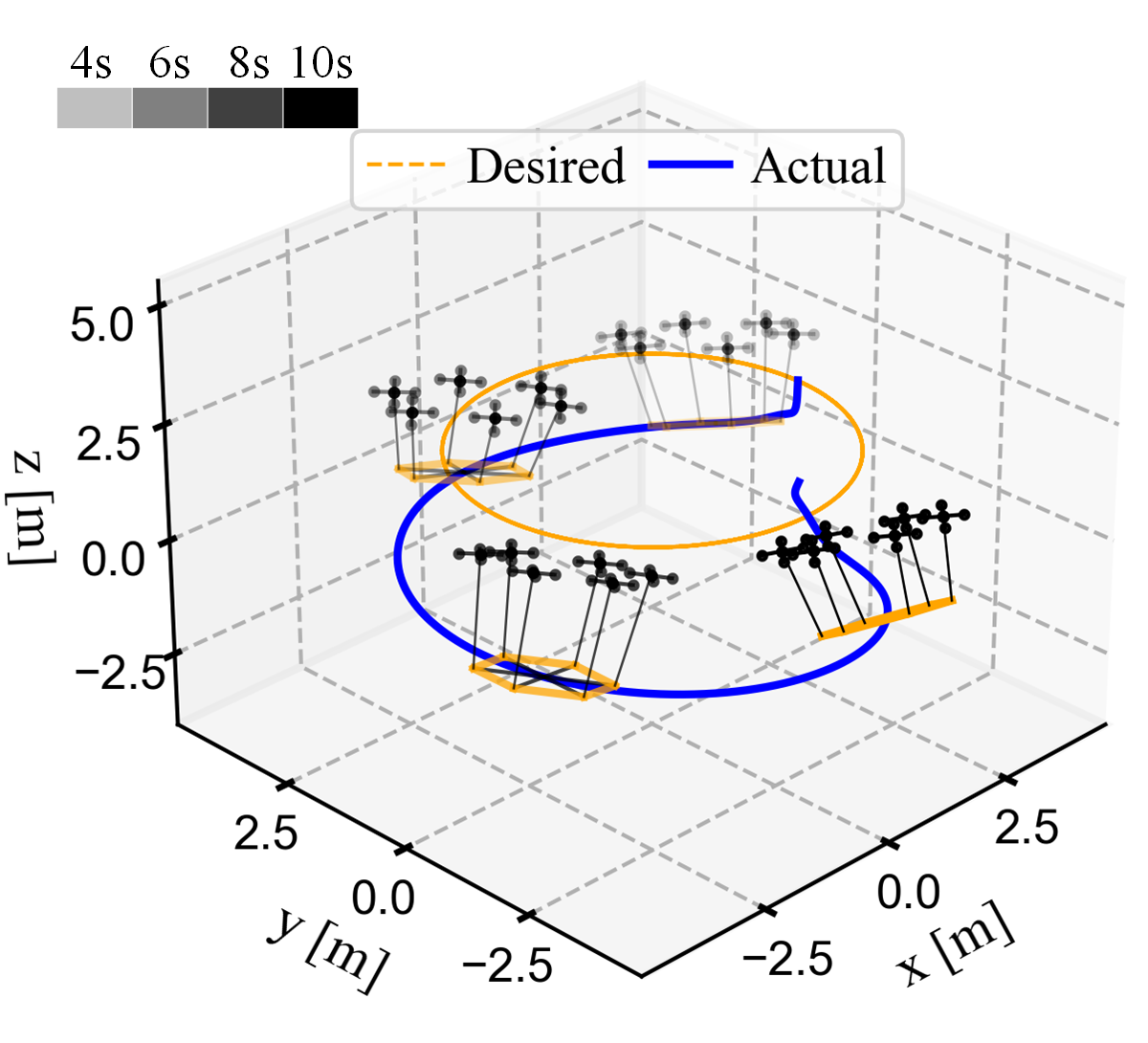}
\caption{Episode $0$ of our method.}
\label{fig:6 quad episode 0_our}
\end{subfigure}
\hfill
\begin{subfigure}[b]{0.325\textwidth}
\centering
\includegraphics[width=0.85\textwidth]{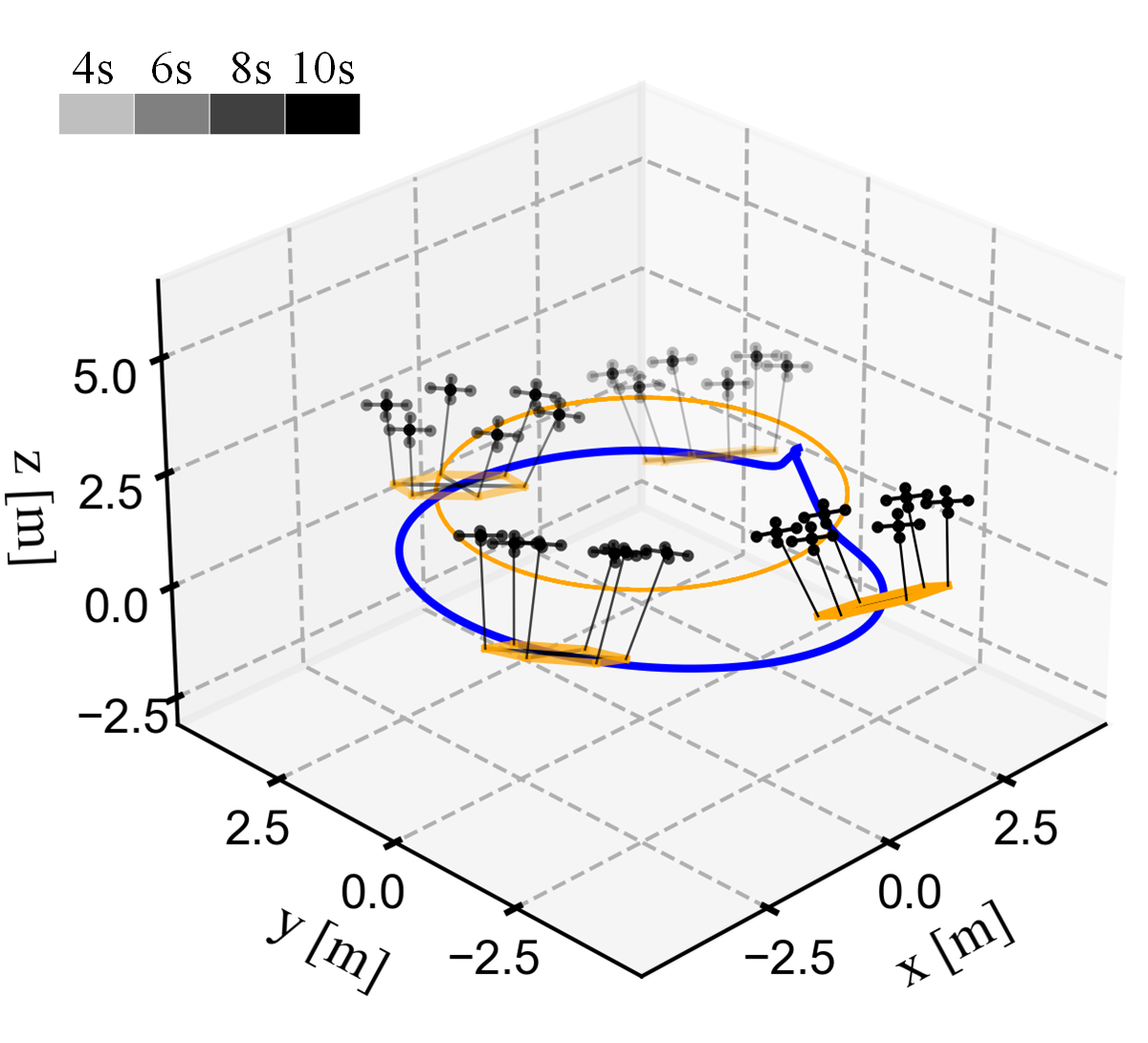}
\caption{Episode $4$ of our method.}
\label{fig:6 quad episode 4_our}
\end{subfigure}
\hfill
\begin{subfigure}[b]{0.325\textwidth}
\centering
\includegraphics[width=0.85\textwidth]{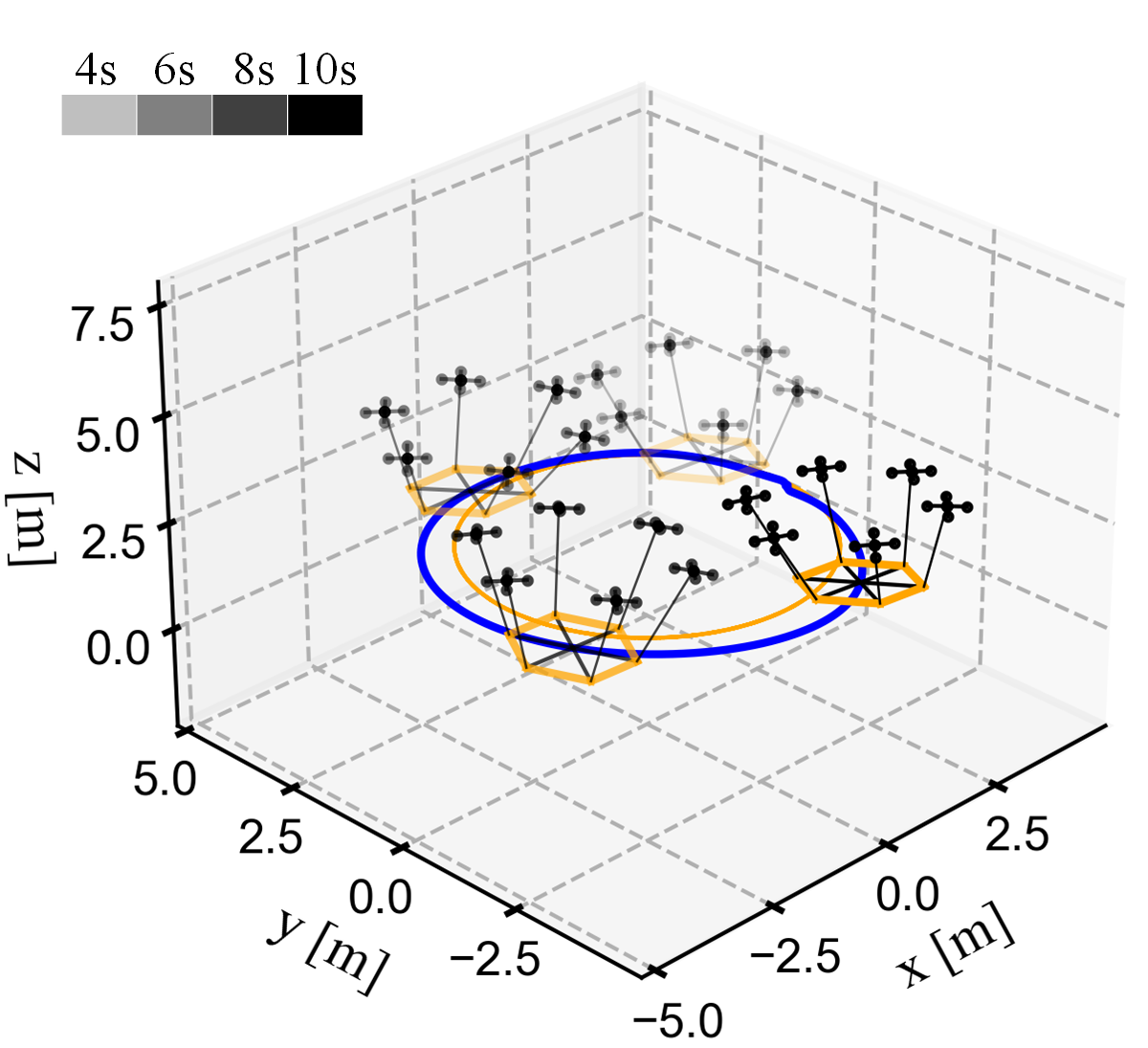}
\caption{Episode $25$ of our method.}
\label{fig:6 quad episode 25_our}
\end{subfigure}
\caption{3D comparisons of the learning process between the open-loop (OL) method and our method using data from the 3rd group. }
\label{fig: training process with 6 quadrotors}
\end{figure*}

We randomly initialize five groups of DNNs using the network model defined in Fig.~\ref{fig:network structure}. The DNNs in each group are trained separately using the three different methods. For each method, all the DNNs in a group are retrained from their respective initial values, which remain the same across all the methods. Fig.~\ref{fig:meanloss_comparison} compares the mean loss during training among the three methods. For fair comparisons, all the mean losses shown in the figure are computed using the same loss function \eqref{eq:tracking loss}, based on the closed-loop system states. Despite the closed-loop implementation of the MPCs, the mean loss of the OL method exhibits significant oscillations, indicating a severely unstable training process. This is visualized in Fig.~\ref{fig:6 quad episode 4_ol}, where, as time increases, the load's trajectory deviates significantly from the desired trajectory, the load becomes more tilted, and the formation of the quadrotors gradually collapses. In some extreme cases (4th and 5th groups), the training of the OL method even halts unexpectedly before reaching the stopping criterion, as the system states become highly unstable and the MPC solver fails to produce valid solutions (i.e., 'NaN' encountered). In sharp contrast, both the closed-loop methods (i.e., ours and the CLWC method) stabilize the training process, even in these two extreme cases. The stable learning process of our method is visualized using data from the 3rd group (as examples) in Figures \ref{fig:6 quad episode 0_our}, \ref{fig:6 quad episode 4_our}, and \ref{fig:6 quad episode 25_our}, where our method improves tracking performance over training episodes while gradually stabilizing the load's attitude in the desired horizontal direction. A comparison of the two closed-loop methods in the first three groups shows that our method further improves learning performance by stabilizing the mean loss at a lower value, demonstrating the benefit of including the couplings in the sensitivity propagation. However, this improvement is less significant than the increase in learning stability achieved by switching from the open-loop to the closed-loop method. In the extreme cases (the 4th and 5th groups), although the CLWC method and our method achieve comparable mean loss values, considering the couplings remains necessary, as doing so is mathematically sound for multilift systems.

\begin{figure}[h]
    \centering
    \begin{subfigure}[b]{0.49\textwidth}
    \centering
    \includegraphics[width=0.75\textwidth]{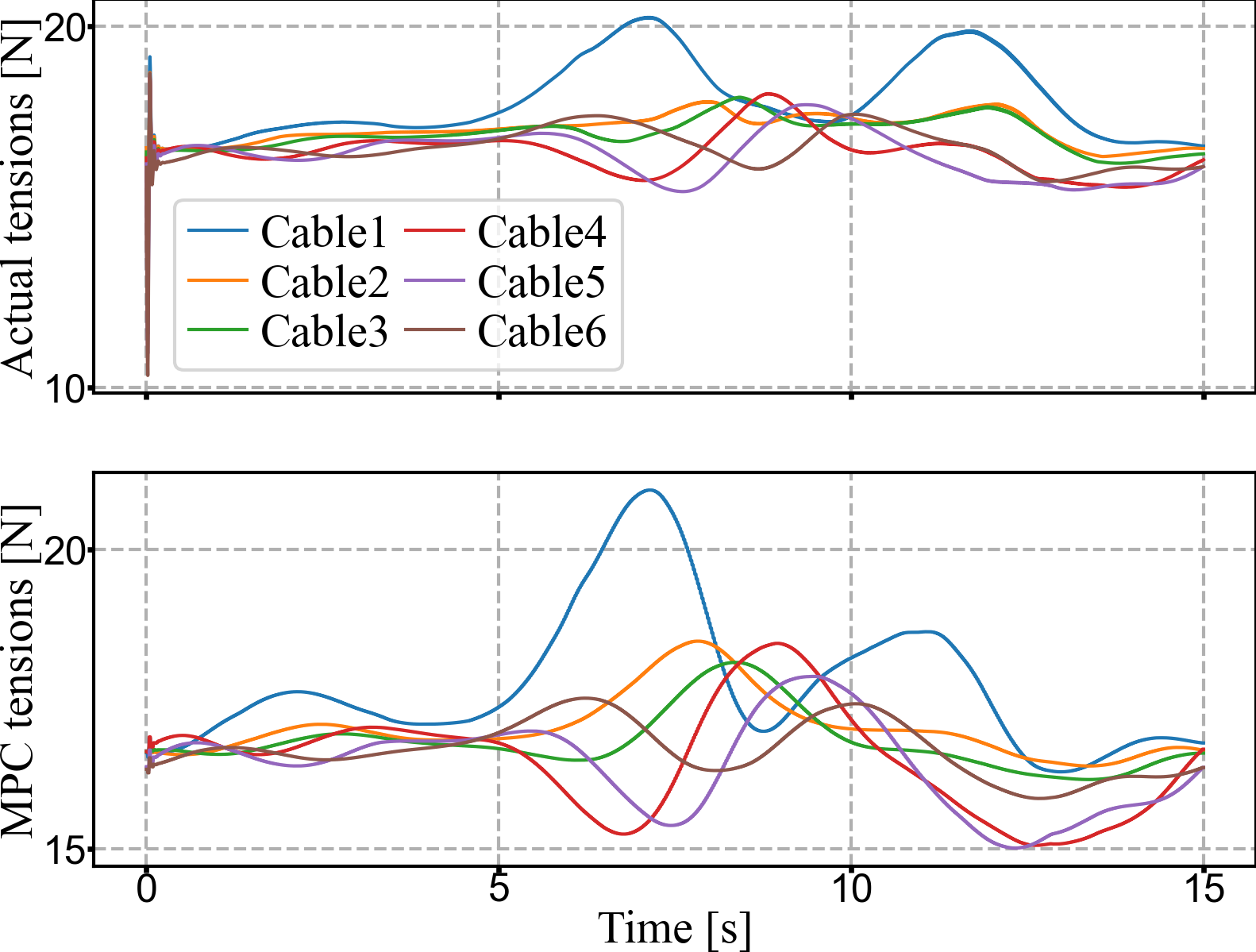}
    \caption{Cable tensions of our method in episode 4.}
    \label{fig:tension cl epi4}
    \end{subfigure}
    \hfill
    \begin{subfigure}[b]{0.49\textwidth}
    \centering
    \includegraphics[width=0.75\textwidth]{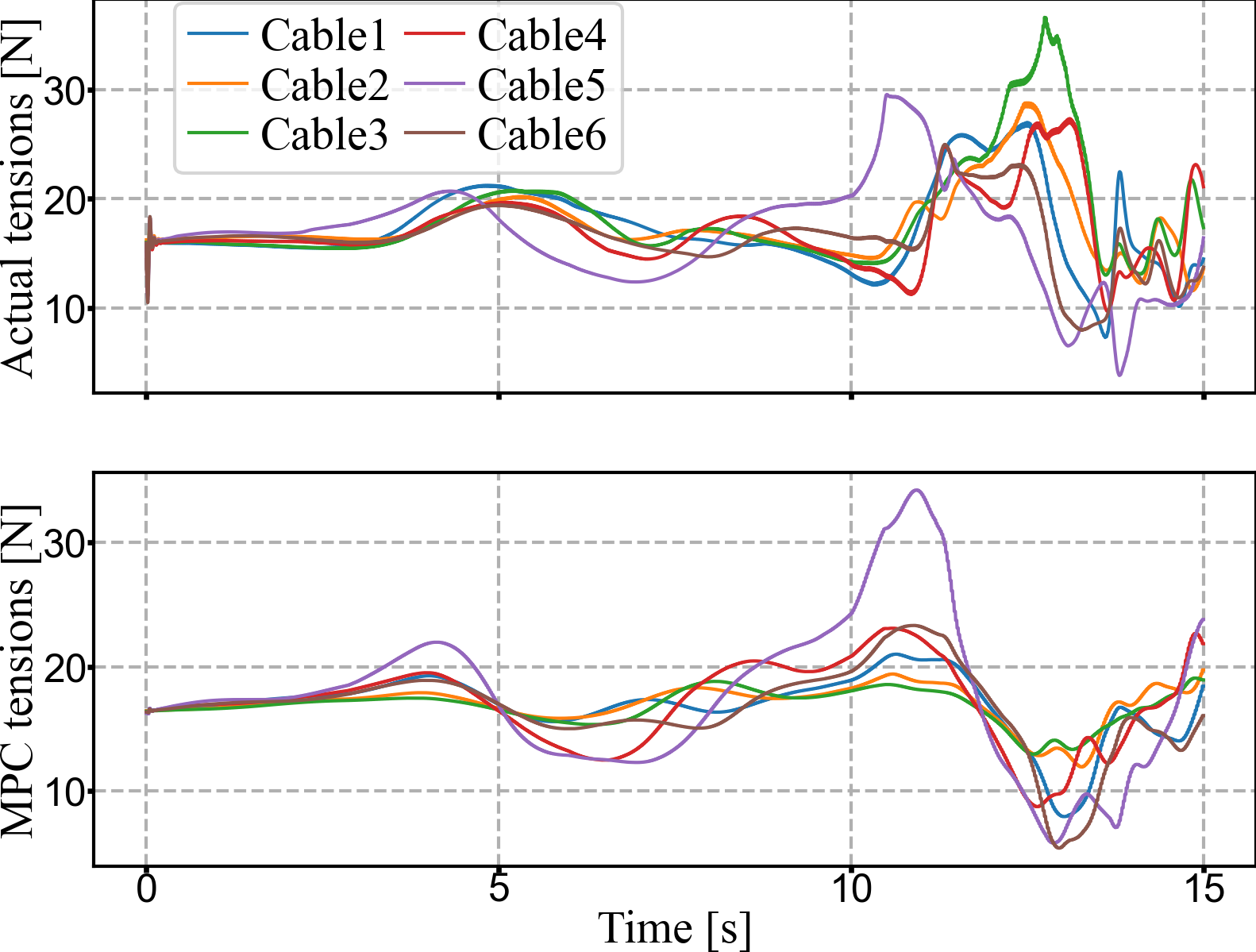}
    \caption{Cable tensions of the OL method in episode 4.}
    \label{fig:tension ol epi4}
    \end{subfigure}
    \caption{Comparisons of the actual and the MPC-computed tension magnitudes in episode 4 between our method and the OL method using data from the 3rd group.}
    \label{fig:tension_6quad_training}
\end{figure}

Our method's advantages over the OL method stem from two main factors. First, we train directly on the closed-loop system states from the actual system dynamics, whereas the OL method relies on the open-loop predicted states from the control system models. This distinction is critical since it addresses the discrepancies between the actual\footnote{The actual tension refers to the tension magnitude computed using the hybrid model \eqref{eq:tension magnitude}.} and the MPC-computed tensions. As shown in Fig.~\ref{fig:tension_6quad_training}, the MPC-computed tensions are smoother than the actual tensions, especially at the beginning. Although both methods use $\mathcal{L}_1$ adaptive control~\cite{wu20221} to robustify the MPCs against the tension differences, training on the closed-loop data provides a more robust and effective means to improve tracking performance. A notable example is in episode 4, where the OL method begins to exhibit instability. Fig.~\ref{fig:tension cl epi4} shows that our method effectively computes appropriate gradients from the closed-loop states to minimize the tension discrepancies, thereby stabilizing the training process. In contrast, the inappropriate gradients in the OL method, calculated using the open-loop prediction trajectories, amplify the tension discrepancies, particularly after $10\ {\mathrm s}$, as shown in Fig.~\ref{fig:tension ol epi4}. This destabilizes each quadrotor and further leads to the collapse of the entire multilift system due to the dynamic couplings. Second, our method permits a longer loss horizon than the MPC horizon. This results in gradients that contain richer information for training compared to open-loop methods like Safe-PDP.

\subsubsection{Evaluation on an Unseen Agile Trajectory}

\begin{figure}[h]
\centering
\begin{subfigure}[b]{0.49\textwidth}
\centering
\includegraphics[width=0.725\textwidth]{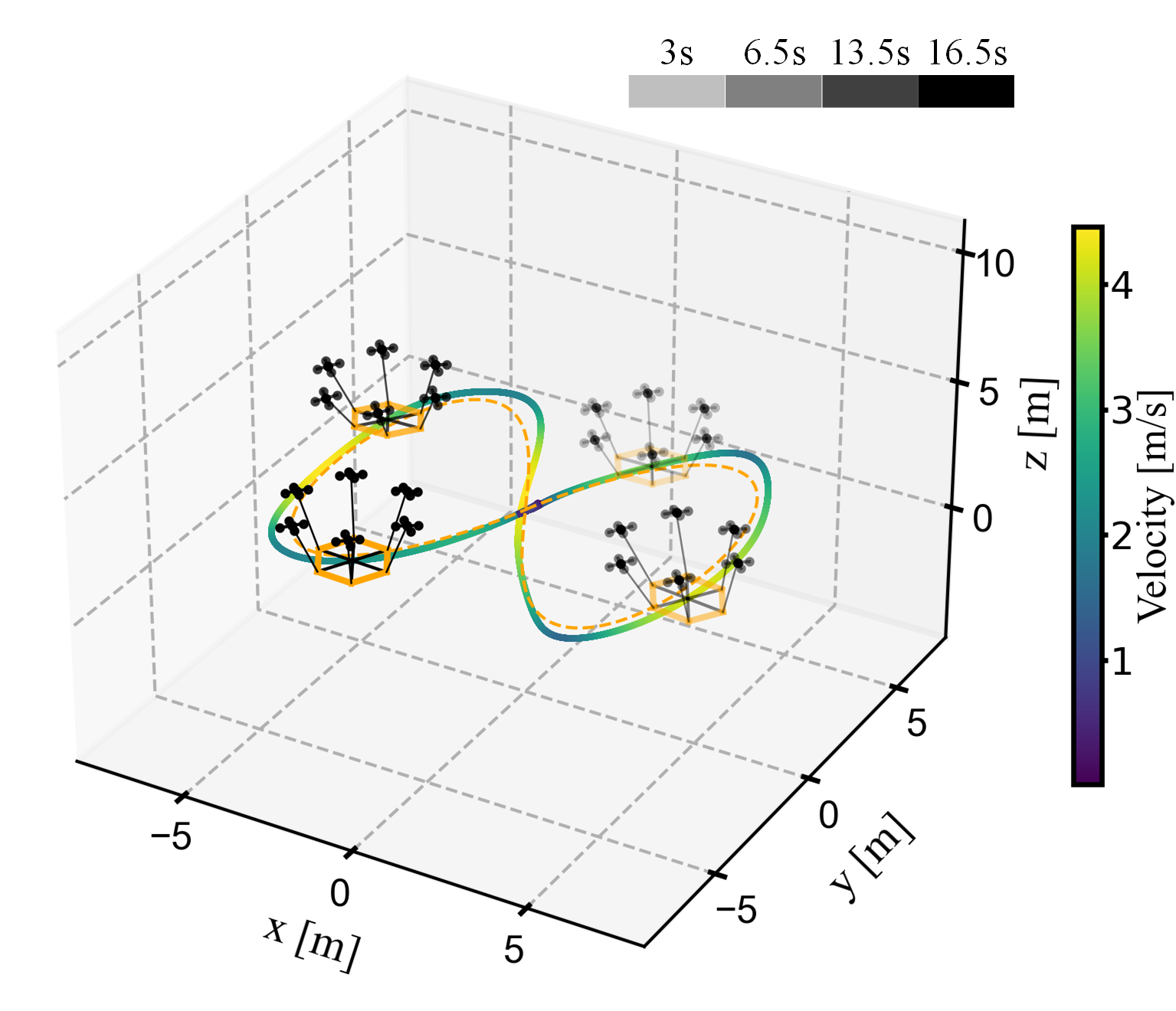}
\caption{Evaluation of our method.}
\label{fig:6 quad fig8 evaluation cl}
\end{subfigure}
\hfill
\begin{subfigure}[b]{0.49\textwidth}
\centering
\includegraphics[width=0.7\textwidth]{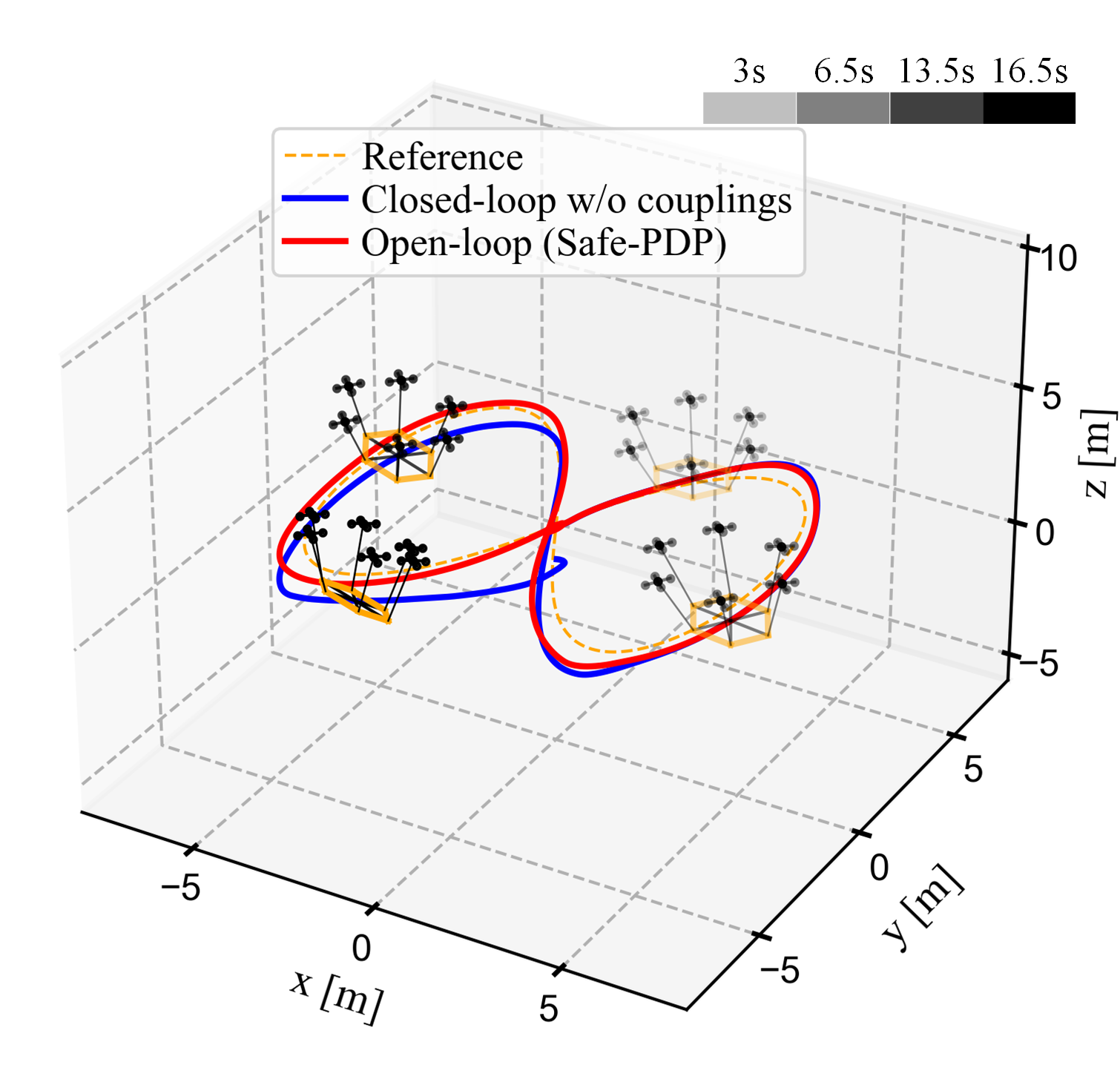}
\caption{ Evaluations of the OL and the CLWC methods.}
\label{fig:6 quad fig8 evaluation ol clwc}
\end{subfigure}
\caption{\footnotesize 3D Illustrations of the evaluation results on an unseen circular trajectory. Compared with the slower trajectory used in training, we reduce the load’s mass to $70 \%$ of
its original value. This reduction ensures sufficient thrust margins to accommodate the larger accelerations required by the more agile test trajectory, addressing stability concerns.}
\label{fig: evaluation of 6 quadrotors}
\end{figure}

In evaluation, we compare the load's tracking performance across the three methods on a previously unseen Figure-8 trajectory, which is more agile than the circle trajectory used in training. Specifically, we evaluate the three methods from the 3rd group, as they perform the best among the first three groups\footnote{For fair comparisons, we do not evaluate the three methods from the 4th and 5th groups, as these two groups represent the extreme cases where the training of the OL method halts unexpectedly due to the numerical issues caused by the highly unstable system states.}. Fig.~\ref{fig: evaluation of 6 quadrotors} illustrates the comparisons using 3D plots. Our method achieves smaller tracking errors and brings the load's attitude closer to horizontal than the other two methods, demonstrating better generalizability. We quantify the tracking performance using root-mean-square errors (RMSEs) and summarize the comparisons in Table~\ref{table:rmse for three quadrotors}. Notably, our method produces significantly smaller RMSEs in all directions compared to the other two methods, improving the accuracy of trajectory tracking and load attitude stabilization by up to $93\%$ and $81\%$, respectively. The only exception is in the $x$ direction, where the RMSE of our method is comparable to that of the OL method.

\begin{table}[h]
\caption{Comparisons of Load Tracking Errors (RMSE)  \label{table:rmse for three quadrotors}}
\centering
\begin{threeparttable}[t]
\begin{tabular}{ c|c c c| c c c } 
\toprule[1pt]
\multirow{2}{*}{Method}  & ${p_x}$ & ${p_y}$ & ${p_z}$ & Roll & Pitch & Yaw\\
       & $\left[ {\mathrm m} \right]$ & $\left[ {\mathrm m} \right]$ & $\left[ {\mathrm m} \right]$ & $\left[ {\mathrm {deg}} \right]$ & $\left[ {\mathrm {deg}} \right]$ & $\left[ {\mathrm {deg}} \right]$\\
\midrule[0.5pt]
Our method  & ${\mathbf {0.16}}$ &  ${\mathbf {0.38}}$ & ${\mathbf{0.05}}$ & ${\mathbf{6.18}}$ & $\mathbf {3.12}$ & $\mathbf{0.61}$\\
CLWC method & $0.20$ & $0.44$ & $0.76$ & $15.65$ & $16.34$ & $2.69$\\
OL method & $0.17$ & $0.41$ & $0.16$& $11.46$ & $9.58$ & $2.97$\\
\bottomrule[0.5pt]
\end{tabular}
\end{threeparttable}
\end{table}

\subsection{Distributed Learning of Adaptive References}\label{subsec:learning references}

\begin{figure*}[t]
	\centering
	{\includegraphics[width=0.8\textwidth]{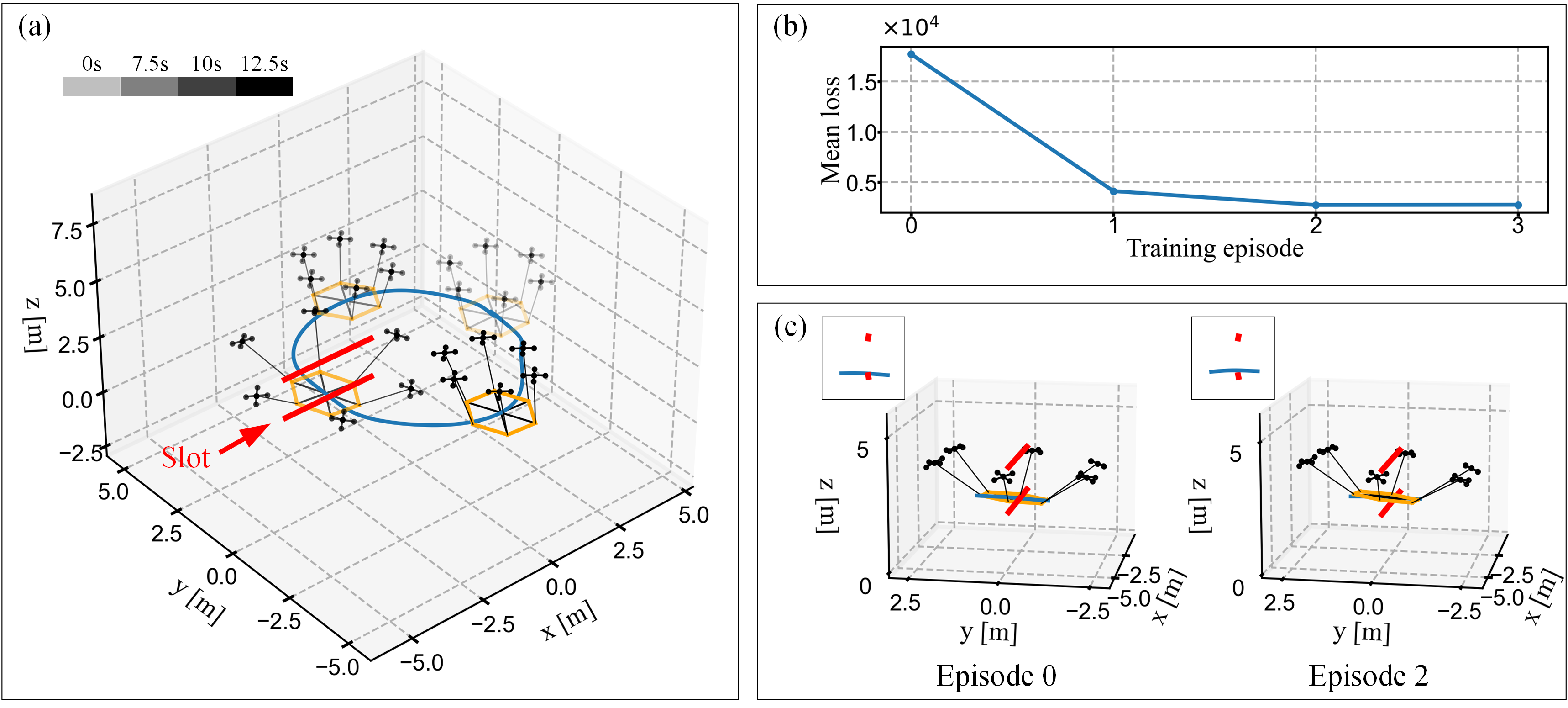}}
	\caption{Illustrations of the process of learning the tension reference using our method. (a) A 3D plot showing the dynamic change of the system configuration in episode 0. (b) The changing pattern of the mean loss during training. (c) 3D plots showing the learning process of the multilift system as it passes through the slot, with the angle of view aligned with the direction of the red arrow in (a).}
\label{fig:training tension reference}	
\end{figure*}

\begin{figure*}[t]
\centering
\begin{subfigure}[b]{0.49\textwidth}
\centering
\includegraphics[width=0.7\textwidth]{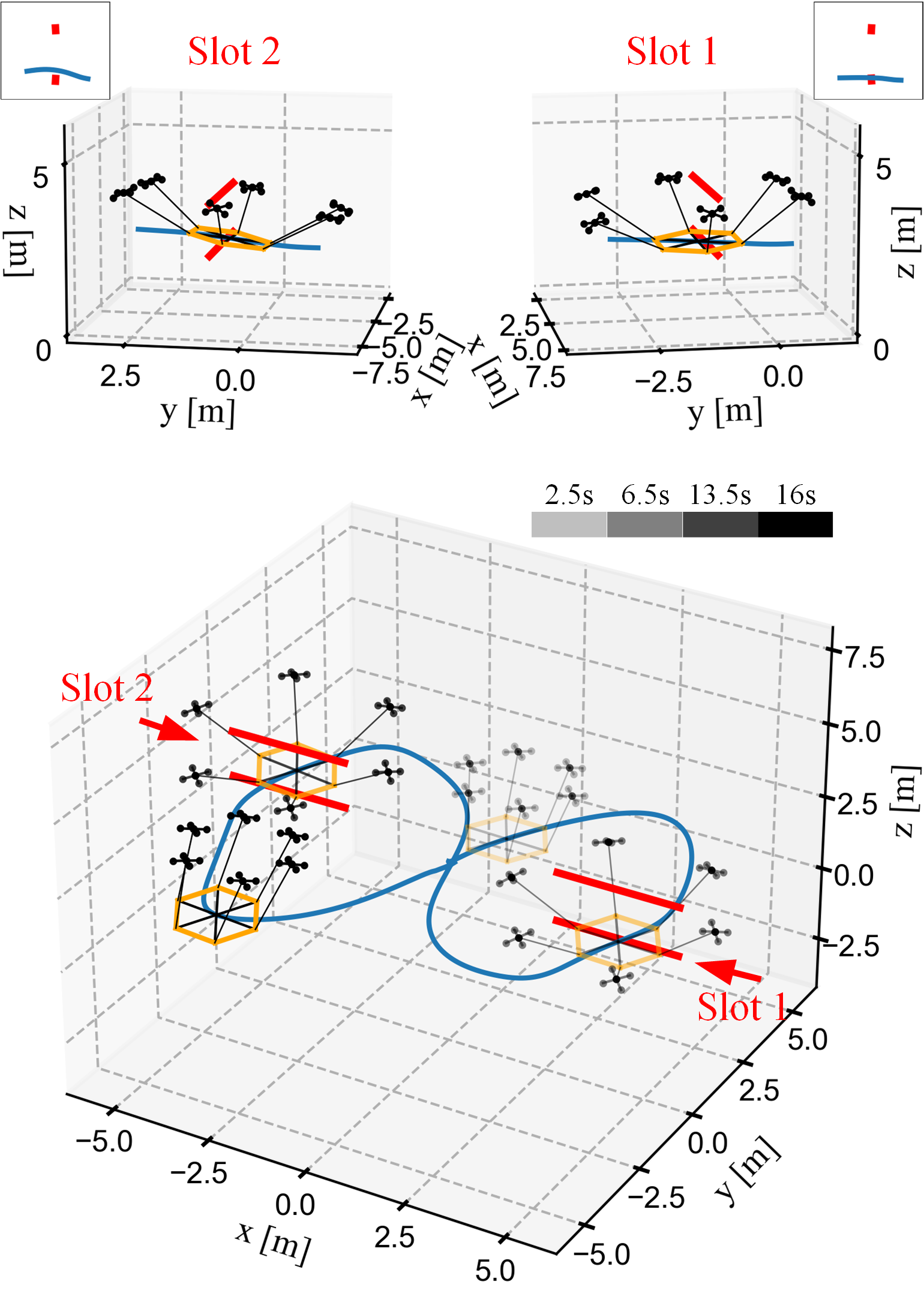}
\caption{Passing through the slots with $\Delta T^{\mathrm{ref}}$ enabled.}
\label{fig:evaluation of tension reference with compensation}
\end{subfigure}
\hfill
\begin{subfigure}[b]{0.49\textwidth}
\centering
\includegraphics[width=0.7\textwidth]{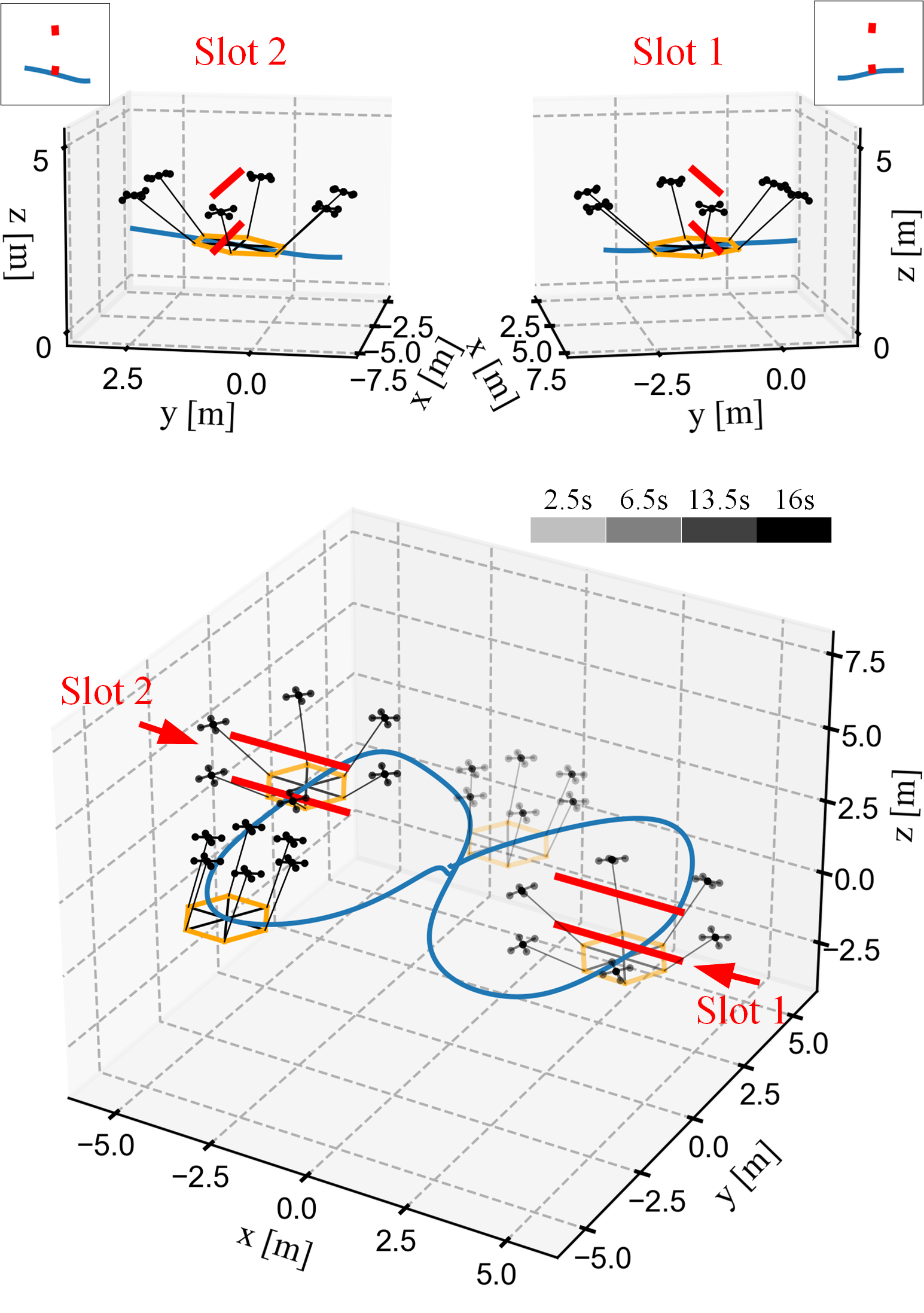}
\caption{Colliding with the slots with $\Delta T^{\mathrm{ref}}$ disabled.}
\label{fig:evaluation of tension reference without compensation}
\end{subfigure}
\caption{3D comparison of the large multilift system passing through the slots, with and without the tension reference compensation.}
\label{fig: evaluation of tension reference}
\end{figure*}

To demonstrate the third advantage of our method, we design an obstacle avoidance flight scenario. Specifically, the same multilift system, equipped with six quadrotors and trained using our method in subsection~\ref{subsec:learning weightings}, is required to pass through a narrow slot. As shown in Fig.~\ref{fig:training tension reference}(a), the slot is on a circular trajectory and has its height shorter than the cable length of $2\ {\mathrm m}$. Successfully passing through confined spaces like this slot is a crucial capability for multilift systems, especially in applications like rescue operations. In this scenario, the quadrotors must spread apart to pass through and then gather together afterward. 

Dynamically changing the configuration poses significant challenges to the design of reference trajectories for multilift systems. To simplify the problem, we consider a dynamic configuration based on the following adaptive cable tilt angle:
\begin{equation}
    \beta_t =\beta_{\min}+\left ( \beta _{\max}- \beta_{\min}\right )\mathrm{exp}\left ( -\eta _{\mathrm{cf}}\left \| {\bm p}_{t}^{l}-{\bm p}^{\mathrm{s}} \right \| _{2}^{4}\right )
    \label{eq:adaptive tilt angle}
\end{equation}
where $\beta_{\min}$ and $\beta_{\max}$ represent the lower and upper bounds, respectively, $\beta_{\max}$ is determined using the geometric relationship between the cable length and the slot height, ${\bm p}^{\mathrm{s}}$ denotes the slot's position in the world frame, and $\eta_{\mathrm{cf}}\in \mathbb{R}_+$ is a positive coefficient. This tilt angle applies to all the cables. Compared with $\beta_t$, designing the tension reference is more challenging. Note that the tension references from the static configuration in subsection~\ref{subsec:learning weightings} are not sufficient to ensure collision avoidance while passing through the slots. This is because the tension reference must be adjusted dynamically when the configuration changes, guiding the load's MPC to generate appropriate tensions that maintain the load's height above the lower boundary of the slot. To achieve this, we aim to learn such an adaptive tension reference using Algorithm~\ref{alg: Distributed policy}. The adaptive tension reference acts as a tension compensation $\Delta T^{\mathrm{ref}}\in \mathbb{R}$ added to the tension reference employed in subsection~\ref{subsec:learning weightings}, and applies to all the cables. It is modeled by an $8$th network for the large multilift system, which takes as inputs the height and vertical velocity tracking errors, along with the dynamic tilt angle $\beta_t$. We define the training loss by combining~\eqref{eq:tracking loss} and~\eqref{eq:obstacle loss} as follows:
\begin{equation}
    L^{l}=\alpha \sum_{t=T}^{T+N_{\mathrm{cl}}}\left [ \left ( 1-\alpha _{t}^{\mathrm{s}} \right )\left ( z_{t}^{l}-z_{t}^{l,\mathrm{ref}} \right )+ \alpha _{t}^{\mathrm{s}}\mathrm{exp}\left ( -\eta \left ( z_{t}^{l}-z^{\mathrm{s}} \right ) \right )\right ]
    \label{eq:loss for learning tension reference}
\end{equation}
where $\alpha _{t}^{\mathrm{s}}=\mathrm{exp}\left ( -\eta ^{\mathrm{s}}\left \| {\bm p}_{t}^{l}-{\bm p}^{\mathrm{s}} \right \|_{2}^{4} \right )$, $\eta^{\mathrm{s}}\in \mathbb{R}_+$ is a positive coefficient, $z_t^l$ is the load's actual height, $z_t^{l,\mathrm{ref}}$ is the reference height from the reference circular trajectory, and $z^{\mathrm{s}}$ denotes the height of the slot's lower boundary.

Fig.~\ref{fig:training tension reference}(b) shows the stable mean loss decreasing over the learning episodes. Fig.~\ref{fig:training tension reference}(c) visualizes the learning process via 3D plots, focusing on the traversing stage. During training, our method gradually increases the load's height when passing through the slot and finally lifts it above the slot's lower boundary in episode 2, achieving collision-free passage. 

\begin{figure}[h]
\centering
\begin{subfigure}[b]{0.49\textwidth}
\centering
\includegraphics[width=0.75\textwidth]{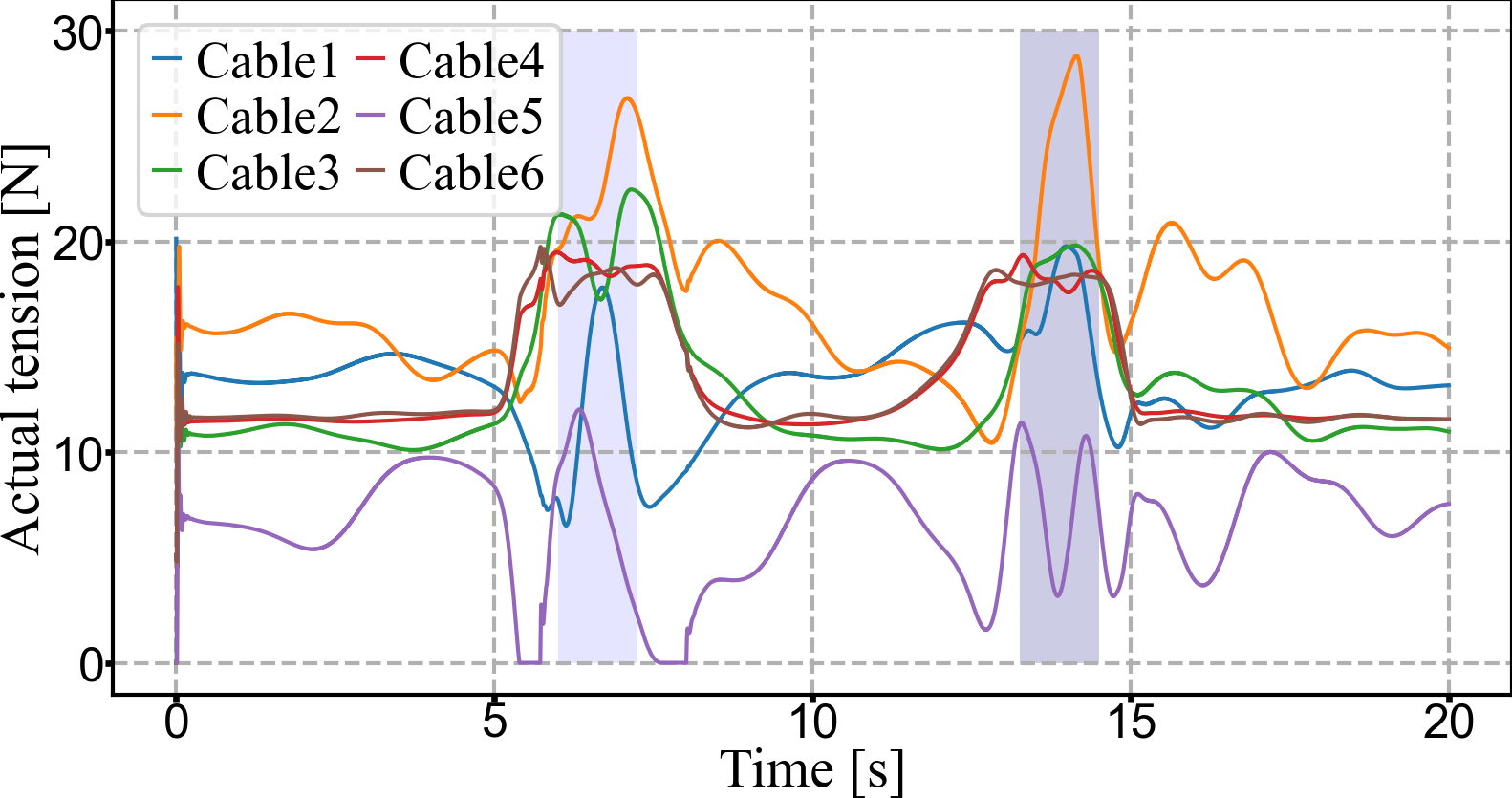}
\caption{Actual tensions with $\Delta T^{\mathrm{ref}}$ enabled}
\label{auto:fig:actual tensions with compensation}
\end{subfigure}
\hfill
\begin{subfigure}[b]{0.49\textwidth}
\centering
\includegraphics[width=0.75\textwidth]{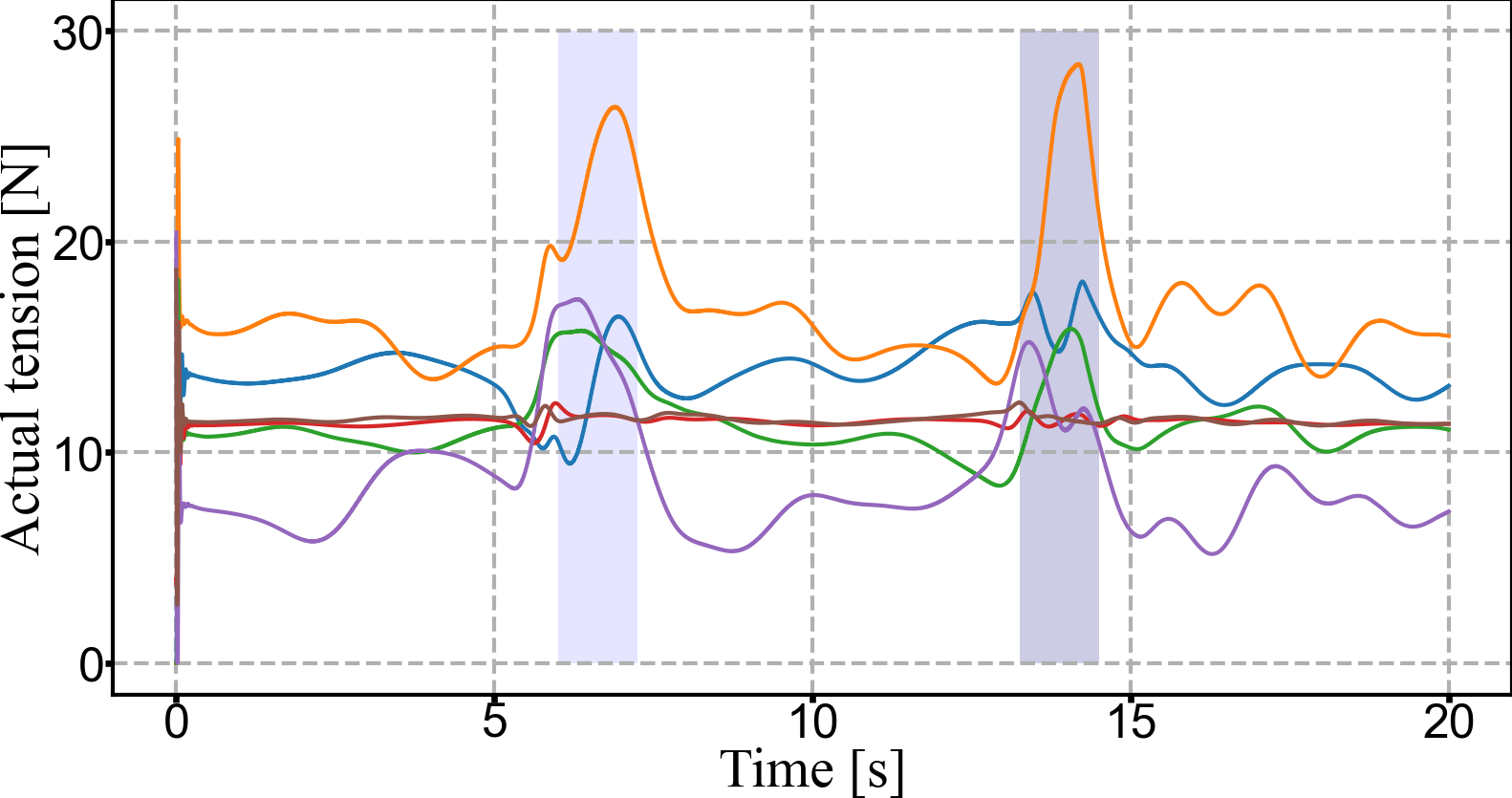}
\caption{Actual tensions with $\Delta T^{\mathrm{ref}}$ disabled.}
\label{auto:fig:actual tensions without compensation}
\end{subfigure}
\caption{Comparison of the actual cable tensions, with and without the tension compensation $\Delta T^{\mathrm{ref}}$. The left and right shadow blocks represent the first and second passing stages, respectively.}
\label{fig: actual tension comparison with and without compensation}
\end{figure}

In evaluation, we test our method on a previously unseen Figure-8 trajectory and add a second slot to the trajectory. Additionally, we slightly lower the heights of both slots compared to the slot height used in training. This requires larger $\beta_t$ values to pass through the slots compared to the training phase, posing a greater challenge to the tension compensation $\Delta T^{\mathrm{ref}}$. Fig.~\ref{fig: evaluation of tension reference} compares the collision avoidance performance of the multilift system when passing through the slots, with and without $\Delta T^{\mathrm{ref}}$. A comparison of the zoom-in plots in Figures~\ref{fig:evaluation of tension reference with compensation} and~\ref{fig:evaluation of tension reference without compensation} shows that the tension compensation is crucial for avoiding collisions with the lower boundaries. This is because, with the compensation, the actual tensions in these six cables increase appropriately during the two passing stages, compared to when the compensation is not applied, as illustrated in Fig.~\ref{fig: actual tension comparison with and without compensation}.

\section{Discussion}\label{section: discussion}
Auto-Multilift features a distributed and closed-loop learning framework, which efficiently trains adaptive higher-level policies (i.e., the hyperparameters modeled by DNNs) via RL for the distributed MPCs of the multilift system. Our method exhibits versatility in learning various MPC hyperparameters. In addition, it outperforms the state-of-the-art open-loop training method Safe-PDP in terms of learning stability and trajectory tracking performance.

On the other hand, Auto-Multilift can be further improved in the following aspects. 
The computation time for the distributed MPCs (implemented via Algorithm~\ref{alg: distributed mpc}) is relatively long due to the high nonlinearity of the multilift system dynamics. To reduce this time for real-world flight, optimized numerical solvers and code structures would be needed. Another limitation is that the stability of nonlinear distributed MPCs and neural network-based controllers remains an open problem. Although a theoretical analysis of the stability may prove difficult, one could explore and extend the boundaries of this stability using meta-learning with curriculum training strategies.

\section{Conclusion} \label{section: conclusion}
This paper proposed a novel learning and control framework, Auto-Multilift, for cooperative cable-suspended load transportation with quadrotors. Auto-Multilift can automatically tune various adaptive MPC hyperparameters, which are modeled by DNNs and difficult to tune manually, via RL in a distributed and closed-loop manner. Our critical insight is that the unique dynamic couplings within the multilift system can be fully exploited to develop the distributed sensitivity propagation algorithm, which is the core of our method that efficiently computes the sensitivities of the closed-loop states w.r.t. the MPC hyperparameters. We have demonstrated through extensive simulations that our method offers good scalability to large multilift systems, and it improves both learning stability and tracking performance over the state-of-the-art open-loop training method. Our future work involves further enhancing the training method of Auto-Multilift via meta-learning and conducting real-world flight experiments to validate its online learning performance in real-time.


\appendix  

%
\subsection{Coefficient Matrices for the Load's MPC Problem}\label{appendix:matrices for load mpc}
We introduce $\bar{\boldsymbol{\theta}}^l$ as the generalized hyperparameters for the load's MPC problem~\eqref{eq:distributed mpc for load}, which can denote ${\bm x}_t^l$, ${\bm x}_{0|t}^{\ast,i} \left ( \forall i\in \mathcal{I}_{q} \right )$, or $\boldsymbol{\theta}^l$. The corresponding matrices are defined as follows:
\begin{equation}
    \begin{aligned}
        \mathbf{H}_{k}^{u\bar{\theta }}&=\frac{\partial ^{2}H_{k}^{l}}{\partial {\bm u}_{k|t}^{\ast,l}\partial \bar{\boldsymbol{\theta }}^{l}}, \mathbf{E}_{k}=\frac{\partial \bar{\bm f}_{k}^{l}}{\partial \bar{\boldsymbol{\theta }}^{l}}, \frac{\partial {\bm x}_{0|t}^{\ast,l}}{\partial \bar{\boldsymbol{\theta }}^{l}}=\frac{\partial {\bm x}_{t}^{l}}{\partial \bar{\boldsymbol{\theta }}^{l}},\\
        \mathbf{H}_{k}^{x\bar{\theta }}&=\frac{\partial ^{2}H_{k}^{l}}{\partial {\bm x}_{k|t}^{\ast,l}\partial \bar{\boldsymbol{\theta }}^{l}},\mathbf{H}_{N}^{x\bar{\theta }} =\frac{\partial ^{2}c_{N}^{l}}{\partial {\bm x}_{N|t}^{\ast,l}\partial \bar{\boldsymbol{\theta }}^{l}}
    \end{aligned}
    \label{eq:matrices for the load pdp}
\end{equation}
where $H_k^l$ is the Hamiltonian for the unconstrained approximation of Problem~\eqref{eq:distributed mpc for load}, defined as $H_{k}^{l}=c_{k}^{l}\left ( {\bm x}_{k}^{l},{\bm u}_{k}^{l};{\bm x}_{k}^{i},\boldsymbol{\theta }^{l} \right )+\boldsymbol{\lambda }_{k+1}^{T}\bar{f}_{k}^{l}\left ( {\bm x}_{k}^{l},{\bm u}_{k}^{l},\Delta t;{\bm x}_{k}^{i} \right )$, similar to~\eqref{eq:cost with soft constraints in Hamiltonian}, $c_k^l$ can be obtained by adding the barrier functions for the tension constraint~\eqref{eq:tension magnitude constraint} to the running cost in~\eqref{eq:cost of load distributed mpc}. The types of $\bar{\boldsymbol{\theta}}^l$ and the respective values of the matrices~\eqref{eq:matrices for the load pdp} are summarized in Table~\ref{table:generalized hyperparameters and matrices for the load}.

\begin{table}[h]
\caption{Coefficient Matrices with $\bar{\boldsymbol{\theta}}^l$ for the Load\label{table:generalized hyperparameters and matrices for the load}}
\centering
\begin{threeparttable}[t]
\begin{tabular}{ c|c c c} 
\toprule[1pt]
Matrices  & $\bar{\boldsymbol{\theta}}^l={\boldsymbol{\theta}}^l$ & $\bar{\boldsymbol{\theta}}^l={\bm x}_t^l$ & $\bar{\boldsymbol{\theta}}^i={\bm x}_{0|t}^{\ast,i}$  \\
\midrule[0.5pt]
$\frac{\partial {\bm x}_{0|t}^{\ast,l}}{\partial \bar{\boldsymbol{\theta }}^{l}}$  & $\mathbf 0$ & $\mathbf I$ & $\mathbf 0$ \\
$\mathbf{H}_{k\geq 1}^{x,\bar{\theta}}$  & $\neq {\mathbf 0}$ & $\mathbf{0}$ & $\mathbf{0}$  \\
$\mathbf{H}_0^{u\bar{\theta}}$ & $\neq {\mathbf 0}$ & $\mathbf{0}$& $\neq {\mathbf 0}$\\
$\mathbf{H}_{k\geq 1}^{u\bar{\theta}}$ & $\neq {\mathbf 0}$ & $\mathbf{0}$ & $\mathbf{0}$\\
$\mathbf{E}_0$ & $\mathbf{0}$ & $\mathbf{0}$ & $\neq {\mathbf 0}$ \\
$\mathbf{E}_{k\geq 1}$ & $\mathbf{0}$ & $\mathbf{0}$ & $\mathbf{0}$ \\
$\mathbf{H}_N^{x,\bar{\theta}}$ & $\neq {\mathbf 0}$ & $\mathbf{0}$ & $\mathbf{0}$ \\
\bottomrule[0.5pt]
\end{tabular}
\end{threeparttable}
\end{table}

\bibliographystyle{IEEEtran}
\bibliography{reference}

\end{document}